\documentclass{article}

\usepackage[final]{neurips_2025}

\usepackage[utf8]{inputenc} % allow utf-8 input
\usepackage[T1]{fontenc}    % use 8-bit T1 fonts
\usepackage{hyperref}       % hyperlinks
\usepackage{url}            % simple URL typesetting
\usepackage{booktabs}       % professional-quality tables
\usepackage{amsfonts}       % blackboard math symbols
\usepackage{nicefrac}       % compact symbols for 1/2, etc.
\usepackage{microtype}      % microtypography
\usepackage{xcolor}         % colors
\usepackage{pifont}
\usepackage{amsmath}
\usepackage{algorithm}
\usepackage{algpseudocode}
\usepackage{algorithmicx}
\usepackage{algpseudocode}
\usepackage{subcaption}
\usepackage{graphicx}
\usepackage{amssymb}
\usepackage{multirow}
\usepackage{makecell}
\usepackage{wrapfig}
\usepackage[export]{adjustbox}
\usepackage{enumitem}
\usepackage{bm}

\newcommand{\metric}{target efficiency}     % 同上

\title{MoESD: Unveil Speculative Decoding's Potential for Accelerating Sparse MoE}
\author{%
Zongle Huang$^1$ \quad Lei Zhu$^2$ \quad Zongyuan Zhan$^2$ \quad Ting Hu$^2$ \quad Weikai Mao$^2$ \quad Xianzhi Yu$^2$\\
\textbf{Yongpan Liu}$^{1,3\text{\dag}}$ \quad \textbf{Tianyu Zhang}$^{2\text{\dag}}$\thanks{\dag ~Corresponding Author.} \\
$^1$Tsinghua University \quad$^2$Huawei Noah’s Ark Lab \quad$^3$BNRist\\
{\small \texttt{\{huangzl23\}@mails.tsinghua.edu.cn} \quad \texttt{\{ypliu\}@tsinghua.edu.cn}} \\
{\small\texttt{\{zhulei168,zhanzongyuan,huting35,maoweikai,yuxianzhi,zhangtianyu59\}@huawei.com}}
}

\begin{document}

\maketitle

\begin{abstract}
  % Large Language Models (LLMs) have achieved remarkable success across many applications, with Mixture of Experts (MoE) models demonstrating great potential. Compared to traditional dense models, MoEs achieve better performance with less computation. Speculative decoding is a widely used technique to accelerate LLM inference without accuracy loss, but it has been considered efficient only for dense models. In \sysname, we first demonstrate that, under medium batch sizes, speculative decoding can accelerate MoE models, being surprisingly more effective than for dense models. More interestingly, as MoE becomes sparser, which is the developing trend of MoE, such acceleration becomes more obvious. To fairly assess various target models' intrinsic potential for speedup, we propose a new metric named \metric, which isolates the impact of algorithms and helps researchers focus on the system bottleneck. \sysname{} is based on a comprehensive theoretical analysis, from which we further develop a modeling framework that can consistently predict the speedup trend. For scenarios like private serving, \sysname{} discloses a new dimension for sparse MoE acceleration, while existing solutions struggle to accelerate them effectively. Experiments on A800 show a 1.8× performance improvement for QwenMoE at medium batch sizes, while for dense Qwen models, speculative decoding has no acceleration. \sysname{} is also expected to have \textcolor{red}{xxx} speedup in private serving.

  Large Language Models (LLMs) have achieved remarkable success across many applications, with Mixture of Experts (MoE) models demonstrating great potential. Compared to traditional dense models, MoEs achieve better performance with less computation. Speculative decoding (SD) is a widely used technique to accelerate LLM inference without accuracy loss, but it has been considered efficient only for dense models. In this work, we first demonstrate that, under medium batch sizes, MoE surprisingly benefits more from SD than dense models. Furthermore, as MoE becomes sparser -- the prevailing trend in MoE designs -- the batch size range where SD acceleration is expected to be effective becomes broader. To quantitatively understand tradeoffs involved in SD, we develop a reliable modeling based on theoretical analyses. While current SD research primarily focuses on improving acceptance rates of algorithms, changes in workload and model architecture can still lead to degraded SD acceleration even with high acceptance rates. To address this limitation, we introduce a new metric \textit{\metric} that characterizes these effects, thus helping researchers identify system bottlenecks and understand SD acceleration more comprehensively. For scenarios like private serving, this work unveils a new perspective to speed up MoE inference, where existing solutions struggle. Experiments on different GPUs show up to 2.29x speedup for Qwen2-57B-A14B at medium batch sizes and validate our theoretical predictions.
\end{abstract}

\section{Introduction}

Recent years have witnessed remarkable success in Large Language Models~(LLMs), with Mixture of Experts~(MoE) architectures showing tremendous potential. Unlike dense models use a single feed-forward network~(FFN) to process all inputs, MoE models replace the FFN with multiple specialized "expert" networks plus a router that selectively activates only a few experts for each input token. Such sparsity in structure enables MoEs with more parameters to achieve higher computational efficiency, and multiple state-of-the-art LLMs, such as DeepseekV3~\cite{liu2024deepseek} and Qwen2.5-Max~\cite{qwen25}, are all MoEs. MoE model architectures are evolving toward larger scales with increased sparsity~\cite{dai2024deepseekmoe, liu2024deepseekV2,liu2024deepseek} and more balanced workload distribution among experts~\cite{shazeer2017outrageously,lepikhin2020gshard}.
% MoE正在发展变为更大，更稀疏，且专家之间负载更均衡。
% MoEs have shown the following developing trends. (1) The total volume keeps growing, usually exceeding that of contemporary dense models: (2) MoEs are becoming sparser, with the ratio of activated experts to total experts decreasing: from 6 out of 64 experts in Deepseek V1~\cite{dai2024deepseekmoe}, to 6 out of 160 in V2~\cite{liu2024deepseekV2}, and 8 out of 256 in V3~\cite{liu2024deepseek}. (3) MoE experts are trained to have more balanced loads to avoid routing collapse~\cite{shazeer2017outrageously} and inefficiency in expert parallelism~\cite{lepikhin2020gshard}.

Speculative decoding~(SD) is a lossless technique to accelerate LLM inference, but conventional wisdom suggests that its efficacy diminishes when applied to MoEs. In SD, a smaller draft model is introduced to rapidly generate multiple candidate tokens, while the larger target model verifies these predictions in parallel, preserving only correctly speculated tokens. For dense models' inference, the time taken to generate a single token and verify multiple ones is roughly the same, as both tasks require the full set of parameters to be loaded once. Therefore, SD gains acceleration through fewer forward rounds of the target model and shorter decoding time of the draft model. However, this acceleration has been demonstrated to diminish in MoEs~\cite{li2024eagle,moesd-inefficiency}, as the multiple draft tokens in verification activate more experts than a single token, 
% thus incurring larger bulks of memory access, and making the verification time much longer than a standard decoding step.
% 
leading to larger memory access and significantly longer verification time compared to a standard decoding step.

In this work, we challenge the conventional belief and demonstrate that, under a moderate batch size, SD can be more effective for MoEs than for dense models. Our key insight is that when the batch size is moderate such that all experts are already activated in a single decoding step, verifying multiple draft tokens will not incur additional expert parameter loading costs. Furthermore, as the MoE becomes sparser, each expert processes fewer tokens per parameter loading, 
% leaving arithmetic units more underutilized and thus creating greater acceleration opportunities for SD.
% 
leading to lower utilization of arithmetic units and thereby creating greater acceleration opportunities for SD.

% 想不出好名字，是不是干脆就不要了？
% \sysname{} is guided by comprehensive theoretical analyses, based on which we develop a modeling method for MoE execution that has been validated and shows speedup trends consistent with real-GPU experiments. 
% The insight above is supported by comprehensive theoretical analyses, based on which we develop a modeling method of SD acceleration for MoE that aligns well with real-GPU experiments. 
% Existing works use acceptance rate to assess the effectiveness of SD, however, it cannot reflect 

% The insight above is supported by comprehensive theoretical analyses, based on which we build a model of SD speedup for MoE. Our modeling is validated to be reliable through comparison with experimental results. Furthermore, it provides an approach for analyzing the execution time of different components, making the end-to-end SD acceleration results more transparent and explainable.
% Existing works typically use acceptance rate to evaluate the efficacy of SD~\cite{chen2023accelerating, leviathan2023fast, li2024eagle, cai2401medusa}. However, it fails to reflect the significant impacts of workload and target model architecture on final speedup. Therefore, we introduce a new metric \textit{\metric{}} to quantify the SD acceleration preference for these factors.
% This metric isolates the extrinsic factors of speculating algorithm selection, thus enabling researchers to focus on the intrinsic system bottlenecks caused by the target model's computational and memory access requirements.
% 

The insight above is supported by comprehensive theoretical analyses, through which we identify a new metric \textit{\metric{}} to quantify how systemic factors (such as workload and target model architecture) affect SD speedup. In contrast to existing SD works that use acceptance rate~\cite{chen2023accelerating, leviathan2023fast, li2024eagle, cai2401medusa}, an algorithmic metric to evaluate how accurately the draft model speculates the target model, our proposed \metric{} isolates extrinsic factors like algorithm selection and focuses on intrinsic system bottlenecks caused by the target model's computational and memory access requirements. As demonstrated in the following sections, even with similar acceptance rates, systemic factors can greatly impact SD effectiveness, making our metric \metric{} necessary for a comprehensive understanding of SD acceleration.

% The insight above is supported by comprehensive theoretical analyses, through which we identify a new metric \textit{\metric{}} to quantify how \textit{systemic} factors (such as workload and target model architecture) affect the speedup by SD. 
% In contrast to the acceptance rate, which is an \textit{algorithmic} metric to evaluate SD in existing works~\cite{chen2023accelerating, leviathan2023fast, li2024eagle, cai2401medusa}, the proposed \metric{} isolates the extrinsic factors of speculating algorithm selection, enabling researchers to focus on the intrinsic system bottlenecks caused by the target model's computational and memory access requirements.
% Existing SD works use \textit{acceptance rate}~\cite{chen2023accelerating, leviathan2023fast, li2024eagle, cai2401medusa}, an \textit{algorithmic} metric to evaluate how accurately the draft model speculates the target model. However, this metric only reflects the impact of extrinsic factors like algorithm selection, while overlooking intrinsic system bottlenecks caused by the target model's computational and memory access requirements. As demonstrated in the following sections, even with similar acceptance rates, systemic factors can greatly impact the effectiveness of SD, making our metric \metric{} necessary.
% 
As a further step, we build a quantitative modeling of SD speedup for MoE based on these theoretical analyses. The consistent matching between our modeling and experiment results confirms the reliability of our analyses. Additionally, the modeling itself provides an approach for analyzing the execution time of different components, making the end-to-end SD acceleration results more transparent and explainable.

% intrinsic bottlenecks (不要加systemic了) by model's 计算和访存的需求？
% We also introduce a new metric, \metric{}, which assesses the potential speedup enabled by SD given the speculation accuracy of the draft model. This metric isolates the \textit{extrinsic} impact of \textbf{algorithmic} optimizations in the draft model, enabling researchers to focus on the \textit{intrinsic} \textbf{system} bottlenecks caused by the target model architecture.

% We remark that \sysname{} unveils a new approach for MoE acceleration, addressing the efficiency gap in moderate batch scenarios. While MoEs excel in extreme cases --- small batches activate only partial parameters, and large batches fully utilize GPU FLOPs --- they struggle with moderate batches, which are common in private serving. In these scenarios, fully-activated experts bring huge memory loads while requests are insufficient to saturate computing resources. The system becomes even more memory-bound when parameters have to be offloaded to CPU memory due to GPU capacity constraint. SD can address this challenge and achieve a significant speedup, while existing methods are less effective for larger and sparser MoE with more balanced experts.
% 

Our work offers a new perspective for lossless MoE acceleration, particularly well-suited for private serving scenarios~\cite{privatellm1,privatellm2,privatellm3}. Private serving has gained popularity among enterprises seeking to safeguard data and model security, with typical applications such as in-house chatbots. These environments typically process moderate batches containing tens of requests. Additionally, our findings can be applied to latency-critical scenarios where large batch sizes are infeasible, or memory-restricted environments where MoEs exceed GPU capacity.
% --- a common scenario given their large parameter counts --- 
% SD delivers significant acceleration. 
% By reducing expensive parameter loadings from CPU memory, it alleviates memory constraints and enhances overall GPU utilization.

In Summary, the main contributions of our work are:
\begin{itemize}[leftmargin=0.8cm]
    \item We refine the conventional belief that speculative decoding cannot effectively accelerate MoEs, demonstrating that under moderate batch sizes, SD is actually more effective in a wider range of batch sizes for sparser MoEs than dense models.
    
    % \item Based on theoretical analysis, we developed a modeling method predicting SD's speedup trends, which consistently align with real-GPU measurements. We further introduce a new systemic metric that reveals speedup opportunities inherent in the model architecture, independent of speculating algorithms.

    \item Based on theoretical analysis, we developed a reliable modeling for SD speedup, thus making the acceleration process transparent and explainable. Existing metrics only assess algorithmic optimization efficiency and cannot fully explain SD speedup, so we introduce a new systemic metric \textit{\metric{}} that reveals speedup opportunities inherent in the target model.
    % , independent of speculating algorithms.
    
    % \item Our findings highlight that SD offers a promising solution to address the efficiency gap in MoE inference for moderate batch scenarios such as private serving. SD is estimated to achieve \textcolor{red}{xx} speedup, while other existing optimizations become ineffective when most experts are activated.
    
    \item Our findings can be applied to accelerate scenarios like private serving. Experiments on various GPUs with the Qwen2-57B-14A-Instruct model demonstrate that SD achieves the highest speedup at the moderate batch size, reaching 2.29x. These experiments also validate our theoretical prediction that SD is more favorable for sparser MoEs.
    % SD shows greater acceleration potential when applied to sparser MoEs.
    
    % These experiments also verify our proposed modeling framework
    % delivers up to 1.8× speedup without specific algorithmic optimizations. In contrast, when using the same batch size for dense models, SD shows no acceleration. These experiments also validate our proposed modeling framework.
\end{itemize}

% 太罗嗦，可以考虑删除
\section{Related Work}

\paragraph{MoE acceleration.}
MoE has emerged as a promising LLM architecture, and many techniques optimize its inference. Model compression methods, including pruning~\cite{xie2024moepruner,lee2024stun}, quantization~\cite{frantar2023qmoe,imani2024mixture}, distillation~\cite{salinas2022knowledge,shu2024llava}, and decomposition~\cite{yang2024moeI,li2023merge}, have been applied to MoEs and achieved great acceleration. They sacrifice model quality for speedup, as in dense models. 
When MoEs are too large to fit in GPU memories and offloading becomes a necessity, several system-level approaches have emerged to optimize inference latency through improved scheduling techniques. Expert prefetching~\cite{xue2024moe,zhong2024adapmoe} predicts and pre-loads experts for upcoming layers based on previous activation patterns, thus overlapping expert loading with current layer computation. Expert caching~\cite{he2024expertflow,tang2024hobbit} caches most frequently activated experts in GPU memory, leveraging expert locality to reduce expensive offloading. Compared to them, our work unveils a new perspective for MoE acceleration that is lossless and doesn't depend on expert imbalance.
% Heterogeneous computing~\cite{ktransformers} leverages CPU resources to perform partial computations, thus alleviating burdens on GPUs.

% When MoEs are too large to fit in the GPU memories and offloading becomes a necessity, 有一些system-level的工作通过改善调度优化这种情况下的推理性能。
% Beyond model-level optimization, system-level works also accelerate MoEs. MoE models offer higher computational efficiency, allowing them to be trained into larger scales than their dense counterparts. Currently, most LLMs exceeding 500B parameters are MoEs~\cite{liu2024deepseek,LLAMA4}. Consequently, GPU memories usually cannot hold the complete model, requiring parameter offloading to slower CPU memory, creating system bottlenecks. Existing works alleviate this problem through expert prefetching or caching~\cite{xue2024moe,zhong2024adapmoe,he2024expertflow,tang2024hobbit}. The former predicts and pre-loads experts for upcoming layers based on previous activation patterns, thus overlapping expert loading with current layer computation. The latter caches most frequently activated experts in GPU memory, leveraging expert locality to reduce expensive offloading. However, as MoEs become sparser and are trained with more balanced experts, the effectiveness of these methods diminishes.

\paragraph{Speculative Decoding.}
Speculative decoding (SD), initially proposed by \cite{leviathan2023fast} and \cite{chen2023accelerating}, has emerged as a widely adopted technique for accelerating LLM inference without sacrificing generation quality. Basic SD employs a smaller model to rapidly generate draft tokens, which are then verified in parallel by the target model that needs to be accelerated. Afterwards, more algorithms are developed to lift the acceptance rate of draft tokens. \cite{miao2024specinfer,cai2401medusa,he2023rest, svirschevski2024specexec,li2024eagle, li2024eagle2, li2025eagle3} adopt tree-structured generation patterns rather than chains to explore a broader range of potential completions. \cite{liu2024deepseek,cai2401medusa,li2024eagle,li2024eagle2,li2025eagle3} propose to replace draft models with specifically trained speculative heads integrated in the target model. 
% This architectural modification enables more effective utilization of the target model's intermediate representations while eliminating the computational overhead associated with model-switching operations. These innovations have substantially improved speculation accuracy and achieved significant speedup factors.

Despite advances in SD algorithms, it has long been considered ineffective for \textit{large batches}~\cite{batchsd2,batchsd3,miao2024specinfer} or \textit{MoE}~\cite{li2024eagle,moesd-inefficiency}, since the verification time in these cases significantly increases. Until recently, MagicDec~\cite{sadhukhan2024magicdec} first challenged that in long-sequence regimes, SD can effectively accelerate \textit{large batches}, primarily due to the significantly increased KV cache altering the computation-to-memory access ratio of the model. However, SD research for \textit{MoE} remains unexplored. In response, our work fills this gap, unveiling that under certain conditions, SD can effectively accelerate MoE models. 

% Since our focus is not on algorithmic optimization, this paper employs the most fundamental chain-like speculation pattern.

% \sysname{} challenges the traditional wisdom on SD, but for MoEs and identifies specific regimes where SD is effective.
% For small batch sizes, LLM decoding is memory-bound, so verifying multiple tokens can utilize the spared computational resources and yield similar latency as generating one. However, as batch sizes increase, the verification time grows, since the system becomes more compute-bound.

% MagicDec~\cite{sadhukhan2024magicdec} first challenges that SD can accelerate large batches when sequences are long, as loading substantial KV caches returns the system to a memory-bound state. In MoE, while verification time also increases, this occurs not because of compute-boundness but rather because verifying multiple tokens activates additional experts, requiring more parameter loads. Like MagicDec, \sysname{} challenges the traditional wisdom on SD, but for MoEs and identifies specific regimes where SD is effective.

\section{Theoretical Analysis}
\label{chap:theo}
% TODO: needs to be updated when the following subsections are finished.
In this section, we present the theoretical analyses supporting our conclusion that SD can be more effective for MoE than dense models at moderate batch sizes. We begin by formalizing general SD speedup and introducing our new metric \textit{\metric} (Sec.~\ref{chap:general_SD_form}). Then, we focus on MoEs, analyzing how workload and MoE sparsity collectively affect the number of activated experts and SD speedup (Sec.~\ref{chap:MoeAnlysis}). Based on these analyses, we develop a performance model that aligns with GPU results (Sec.~\ref{chap:modeling}). We further discuss the practical value of our theoretical findings (Sec.~\ref{chap:application}).
 
\paragraph{Preliminaries.} LLM inference time is collaboratively determined by computation and memory access. When an operator is processed on a GPU, memory access and computation operations are pipelined and overlapped, causing the more time-consuming operation to become the bottleneck and determine the overall processing time, as depicted by the roofline model~\cite{llmviewer,roofline}. The roofline model ridge point (RP) of hardware and the arithmetic intensity (AI) of software are defined as Eq.~\ref{eq:AIRP}.
When AI < RP, the system is \textit{memory-bound}, and adding more computation will not significantly increase processing time. When AI > RP, the system is \textit{compute-bound}, and increases in computation will directly reflect in processing time. In this paper, when we describe a system as "more memory-bound", we mean $\frac{\textbf{AI}}{\textbf{RP}}$ is smaller.
\begin{equation}
\begin{small}
    \textbf{RP} = \frac{\text{peak computation power \textit{(unit: Flops)}}}{\text{peak memory bandwidth \textit{(unit: bytes/second)}}} \quad \textbf{AI} = \frac{\text{computation operation \textit{(unit: times)}}}{\text{memory access volume \textit{(unit: bytes)}}}
    \label{eq:AIRP}
\end{small}
\end{equation}
% For hardware, we can compute its roofline model ridge point (RP), defined as the peak FLOPS divided by peak memory bandwidth (FLOPS/bytes/second); while for the workload, we can compute its arithmetic intensity (AI), defined as computation operations divided by memory access volume. 
\subsection{Formulation of Speculative Decoding Speedup and Target Efficiency}
\label{chap:general_SD_form}

We first formalize the processing time of speculative decoding, denoted as $T_{SD}$. To generate a sequence of length $S$, SD goes through $R$ rounds, each containing three stages: \ding{172} the draft model proposes $\gamma$ tokens as specified by the speculation strategy; \ding{173} the target model verifies these tokens; \ding{174} rejection sampling~\cite{chen2023accelerating} discards incorrectly predicted tokens based on logits from target and draft models.
We use $T_T(b,s)$ and $T_D(b,s)$ to represent the time for once forwarding of the target and draft model, respectively, where $b$ and $s$ are the formal arguments for batch size and the number of tokens to process.\footnote{Since we work with typical sequence lengths and moderate batch sizes, the impact of KV-cache on performance is limited, allowing us to omit the already generated sequence length from our analysis. For cases where KV-cache becomes the dominant factor, see \cite{sadhukhan2024magicdec}.} Therefore, the time for processing a batch containing $B$ requests is given by:
\begin{equation}
    T_{SD} = R \times (T_{propose}+T_{verify}+T_{reject}) = R \times \Big( \gamma \cdot T_D(B,1) + T_T(B,\gamma) + T_{reject}\Big)\\
    \label{eq:SDtime}
\end{equation}
Then the speedup of SD to normal auto-regression decoding $T_{AR}$ is given by:
\begin{align}
    Speedup = \frac{T_{AR}}{T_{SD}} & = \frac{S \cdot T_{T}(B,1)}{R \cdot \Big(\gamma \cdot T_D(B,1) + T_T(B,\gamma) + T_{reject}\Big)} \\
    &= \frac{S}{R} \cdot \frac{1}{\gamma \cdot \frac{T_D(B,1)}{T_T(B,1)}+\frac{T_T(B,\gamma)}{T_T(B,1)}+\frac{T_{reject}}{T_T(B,1)}}
    \label{eq:speedup}
\end{align}
$\frac{S}{R}$ represents the average length of accepted tokens per SD round, which can be further expressed as $\sigma \times (\gamma+1)$. Here, $\sigma$ is the ratio of actually generated tokens to the theoretical maximum if all draft tokens were accepted. We note that $\sigma$ differs from the acceptance rate $\alpha$ commonly referenced in previous works~\cite{leviathan2023fast,chen2023accelerating,li2024eagle}, which represents the probability of the target model accepting a new draft token given the prefix. $\sigma$ can be computed from $\alpha$ as shown in Eq.~\ref{eq:alpha2gamma}. 
The numerator follows from~\cite{leviathan2023fast}, and the denominator accounts for all $\gamma$ draft tokens accepted, plus a bonus token generated during the forward verification pass.
\begin{equation}
    \sigma = \frac{\textit{\text{expected generated tokens}}}{\textit{\text{maximal possible accepted tokens}}} = \frac{\frac{1-\alpha^{\gamma+1}}{1-\alpha}}{\gamma+1}
    \label{eq:alpha2gamma}
\end{equation}
The denominator of Eq.~\ref{eq:speedup} consists of three terms. $\frac{T_D(B,1)}{T_T(B,1)}$ is the ratio of draft-model forward time over target-model forward time, reflecting the relative volume of draft and target models. It is also kept small (usually less than 1/10~\cite{miao2024specinfer,li2024eagle,leviathan2023fast}) to ensure the speculation is efficient. $\frac{T_{reject}}{T_T(B,1)}$ is even smaller, since $T_{reject}$ only involves sampling rather than model inference. $\frac{T_T(B,\gamma)}{T_T(B,1)}$, which is the ratio of multi-token forward time over single-token forward time, has the biggest value among these three items and significantly affects the final speedup. As indicated by Eq.~\ref{eq:speedup}, its increase causes speedup reduction. 
% SD's ineffectiveness for large batches and MoE both result from the increase in $\frac{T_T(B,\gamma)}{T_T(B,1)}$, driven by two factors
Two different factors drive its increase, explaining SD's ineffectiveness under (1) \textit{large batches} for both dense models and MoE, and (2) \textit{MoE} with small batches, respectively:
\begin{enumerate}[label=(\arabic*)]
    \item The compute-boundness. The model's $\frac{T_T(B,\gamma)}{T_T(B,1)}$ approaches 1 when more memory-bound (smaller batch size $B$) but increases to $\gamma$ when more compute-bound (larger batch size $B$).
    \item The extra memory loads. For small $B$s, $T_T(B,\gamma)$ is notably greater than ${T_T(B,1)}$ as more experts are activated and need to be loaded. Since the system is still memory-bound at small $B$s, memory load profoundly determines the processing time.
\end{enumerate}

% Given that underlying factors can be reflected through this ratio, 

Therefore, we define \textit{\textbf{\metric{}}} as $\frac{T_T(B,1)}{T_T(B,\gamma)}$, which helps understand the systemic causes of SD acceleration degradation as listed above. Our experiments shown in Fig.~\ref{fig:basic} demonstrate that \metric{} consistently reflects the trend of SD speedup variations. Despite the importance of this value, previous works have rarely noticed it, primarily due to differences in research focus. Previous SD research mainly addresses the question by lifting the acceptance rate: 
\begin{quote}
\textit{Given the target model, which \textbf{draft model or algorithm} achieve better speedups?}
\end{quote}
In contrast, our work focuses on the following question by examining \metric: 
\begin{quote}
    \textit{Under the same level of algorithmic optimization, which types of \textbf{target models or workloads} are more favorable for SD?}
\end{quote}

We believe \metric{} help researchers understand SD more comprehensively. Even when target-draft pairs have the same acceptance rate $\alpha$, changes in the target model's architecture and the workload can significantly affect overall speedup. By introducing \metric, we can decouple algorithmic optimization from systemic optimization, thus helping to identify the systemic acceleration bottlenecks and assess potential speedup.

\subsection{Moderate Batch Size Enables Speculative Decoding Speedup for MoE}
\label{chap:MoeAnlysis}

Although SD is ineffective for MoE under small batches, we demonstrate in this subsection that at moderate batch sizes—an overlooked regime in previous studies—SD speedup increases and benefits more from MoE with higher sparsity. 
Essentially, when the batch size falls within ranges where all experts are activated but remain far from being assigned adequate workloads, FFNs become memory-bound, presenting an opportunity to leverage computational power almost for free through SD. 
To demonstrate this, we first formalize the expected number of activated experts, and then show MoE FFNs become more memory-bound as the model becomes sparser.

\begin{figure}
    \begin{small}
  \centering
  \begin{subfigure}[b]{0.328\textwidth}
      \centering
      \includegraphics[width=\textwidth]{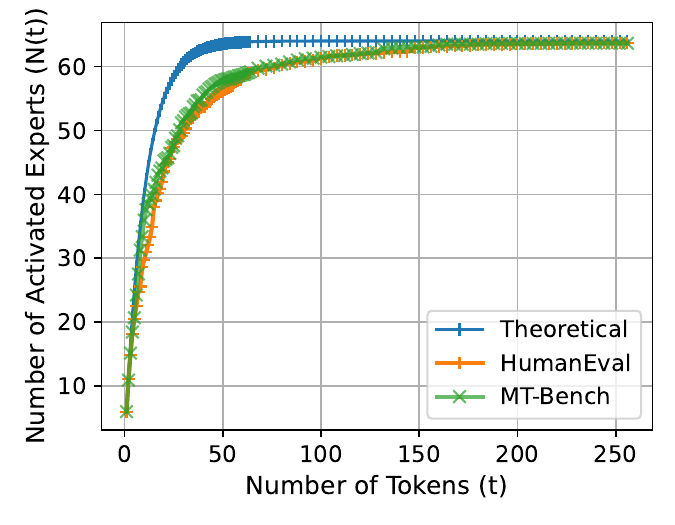}
      \vspace{-20pt}
      \caption{}
      \label{fig:vsdeepseek}
  \end{subfigure}
  \begin{subfigure}[b]{0.328\textwidth}
      \centering
      \includegraphics[width=\textwidth]{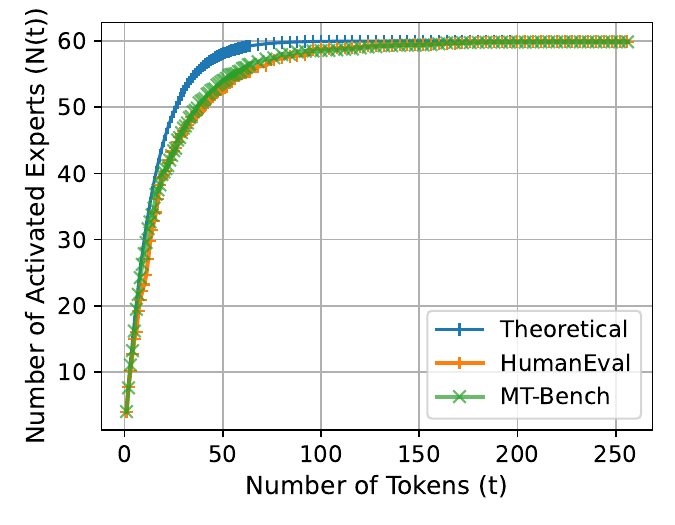}
      \vspace{-20pt}
      \caption{}
      \label{fig:vsqwen}
  \end{subfigure}
  \begin{subfigure}[b]{0.328\textwidth}
      \centering
      \includegraphics[width=\textwidth]{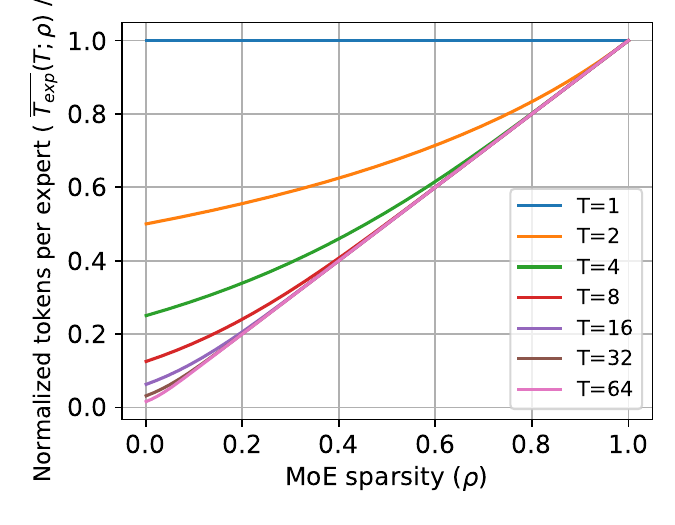}
      \vspace{-20pt}
      \caption{}
      \label{fig:Texp}
  \end{subfigure}
  \end{small}
  \vspace{-15pt}
  \caption{Activation status and workload of experts. (a) and (b): Comparison between theoretical and actual number of activated experts $N(t)$ on different datasets. (a) is for Deepseek-V2-Lite-Chat ($\rho=6/62$) and (b) is for Qwen1.5-MoE-Chat ($\rho=4/60$). (c): Normalized number of tokens to process per expert ($\overline{T_{exp}}$) versus MoE sparsity ($\rho$) for given input token count $T$. 
  % $\overline{T_{exp}}$ decreases as $\rho$ becomes smaller when $T>1$.
  % (b) and (c) shows the theoretical analysis aligns well with actual results, regardless of dataset or model.
  }
  \label{fig:simple_plot}
\end{figure}

We use the Bernoulli random variable $X$ to indicate the activation for experts: $X_i=1$ for expert $i$ being activated, $0$ otherwise. For simplicity, we assume $X$s are i.i.d. Then, the expected number of activated experts $N$ can be expressed as Eq.~\ref{eq:expectation}, where $E$ denotes the total expert count and $Pr(X_i)$ represents the probability that the $i^{th}$ expert is activated.
\begin{equation}
    N = \sum_i \mathbb{E}[X_i]=\sum_i Pr(X_i)=E \cdot Pr(X)
    \label{eq:expectation}
\end{equation}
Given $t$ tokens passed through the MoE gate, then $Pr(X)$ is expressed as Eq.~\ref{eq:PrX}. $K$ denotes the number of activated experts per token, which is an architectural hyperparameter for MoE:
\begin{equation}
    Pr(X)=1-Pr(\text{ None of the $t$ tokens activates the expert })=1-(\frac{E-K}{E})^t
    \label{eq:PrX}
\end{equation}
Therefore, the overall expression of $N(t)$ is given by Eq.~\ref{eq:N_full}. 
Our derivation assumes uniformly activated experts, which is reasonable for well-trained MoE models. Load imbalance among experts can lead to routing collapse~\cite{shazeer2017outrageously} and decrease computational efficiency in expert-parallel deployment~\cite{lepikhin2020gshard}, so state-of-the-art MoE models are typically trained with methods like incorporating auxiliary loss~\cite{fedus2022loadbalance1, lepikhin2020gshard} to ensure that experts have balanced loads. The experiment results also verified our theoretical derivation of $N(t)$, as shown in Fig.~\ref{fig:vsdeepseek} and \ref{fig:vsqwen}.
\begin{equation}
    N(t) = E \cdot \Big(1-(\frac{E-K}{E})^t\Big)
    \label{eq:N_full}
\end{equation}
We then solve how many tokens can lead to full activation. Since $N(t)$ asymptotically approaches $E$ when $t$ tends to infinity, and in practice $N(t)$ should be a finite integer, we deem $N(t)>\tau E$ as almost full activation, where $\tau$ is usually a large ratio such as 0.95. We further express $K=\rho E$, where $\rho$ is the sparsity of MoE, then the token threshold $T_{thres}$ can be solve by:
\begin{equation}
    N(T_{thres}) = E \cdot \Big(1-(1-\rho)^{T_{thres}}\Big) \ge \tau E ~~ \Rightarrow ~~ T_{thres} =\lceil \log_{(1-\rho)}(1-\tau) \rceil
    \label{eq:NTthres}
\end{equation}
Therefore, when $B$ exceeds $T_{thres}$, the number of activated experts saturates, causing the $B\gamma$ tokens in verification to incur only marginally larger memory access. Having addressed the second factor (namely, extra memory loads) for {\small $\frac{T_T(B,\gamma)}{T_T(B,1)}$}'s increase analyzed in Sec.~\ref{chap:general_SD_form}, we now turn to the potential limitations caused by the first factor of compute-boundness. If such $B$s make the system compute-bound, SD would also fail to accelerate MoE effectively. Our answer to this concern is: Sparser MoEs \textbf{\textit{delay}} the transition from memory-bound to compute-bound when input tokens count increases.

We have obtained that given $t$ tokens, $N(t)$ experts are activated. Since each token activates $K$ experts, the number of tokens each expert needs to process on average $\overline{T_{exp}}$ can be computed as:
\begin{equation}
    \overline{T_{exp}}(t;\rho) =\frac{t\cdot K}{N} = \frac{t\cdot(\rho E)}{E\cdot \Big(1-(1-\rho)^{t}\Big)}=\frac{\rho t}{1-(1-\rho)^t}
    \label{eq:Texp}
\end{equation}
As proven in Appendix and shown in Figure~\ref{fig:Texp}, given $t=T>1$, $\overline{T_{exp}}(T;\rho)$ decreases with $\rho$, indicating that as MoE becomes sparser, each expert processes \textit{fewer tokens} per parameter loading. Consequently, the system running sparser MoEs is more \textit{memory-bound}, leading to lower utilization of arithmetic units. The verification stage can therefore leverage these spare resources without notably increasing processing time. In contrast, dense models are extreme cases with $\rho=1$, where the FFN consistently approaches the maximal possible arithmetic intensity of $T$, and the system transitions rapidly to the compute-bound regime as $T$ increases.

% Intuitively, as $\rho$ decreases from $1$, the denominator of $\overline{T_{exp}}$ saturates, while the numerator decreases \textit{linearly}. Therefore, the decrease of the numerator dominates the overall trend. This conclusion tells that, as MoEs become \textit{sparser} (namely, smaller $\rho$), each expert processes \textit{fewer tokens} per parameter loading. 

% Especially, when pushing the sparsity $\rho$ to its limit value $1$, which corresponds to the dense model, $\overline{T_{exp}}=T$. In this case, each weight in FFN achieves the \textit{maximal possible reuse}, which is certainly more compute-bound than any MoE.

We should note that our conclusion is based on a relatively large MoE FFN portion in the whole model, which holds true for current MoE models whose most parameters are experts. In a hypothetical extreme case where Attention dominates and the MoE FFN is negligible, MoE's sparsity would have only a limited impact on overall system performance as indicated by Amdahl's Law.

% The experiment results further confirm our theoretical analysis. Figure~\ref{fig:vsqwen} and \ref{fig:vsdeepseek} compare the theoretical and real $N(t)$ across different benchmarks and MoEs, showing consistent trends.

\subsection{A Modeling Method for Speculative Decoding Speedup}
\label{chap:modeling}

\begin{algorithm}[t]
  \begin{small}
      \caption{The Modeling of SD Speedup and Corresponding Fitting Method}
      \label{alg:main}
      \begin{algorithmic}[1]
          \State \textbf{Measurement Input}: A total of $m$ measurements denoted as $\mathbf{M}$. Each $\mathbf{M}_i, i=1,2,...,m$ contains the attributes including batch size $B$, draft length $\gamma$, number of activated experts per token $K$, total number of experts $E$, the ratio of accepted token counts to the maximal possible accepted tokens $\sigma$, \textit{Speedup} for the actual speedup achieved.
          \State \textbf{Output}: The optimal fitting parameter \textit{params*}.
          % \State \textbf{Workload Input}: batch size $B$, draft length $\gamma$, number of activated experts per token $K$, total number of experts $E$, estimated ratio of accepted token counts to the maximal possible accepted tokens $\sigma$.
          % \State \textbf{Workload Output}: the SD speedup under the given workload input parameters, denoted as \textit{Speedup}.
          % batch size $\mathbf{B}$, draft length $\mathbf{\gamma}$, speculation accuracy in a round $\mathbf{\sigma}$
          % \State \textbf{Hyperparameters}: \texttt{RP}, \texttt{slope}; \texttt{attn\_bias}, \texttt{attn\_scaler}; \texttt{moe\_expert\_scaler}, \texttt{moe\_bias}, \texttt{moe\_scaler}; \texttt{draft\_bias}, \texttt{draft\_scaler}; \texttt{reject\_bias}, \texttt{reject\_slope}.
          \vspace{5pt}
            \State \textit{\textbf{def ComputeSpeedup}}(\textit{params}, $B$, $\gamma$, $K$, $E$, $\sigma$): \Comment{\textcolor{purple}{Compute the SD Speedup}}
                % \State \hspace{\algorithmicindent} {\normalsize \textbf{Function}~} \textit{GetActivatedExpertNum}($t$, $E$, $K$):
                %     \State \hspace{\algorithmicindent}\hspace{\algorithmicindent} return $E \cdot (1-((E-K)/E)^t)$
                % \State \hspace{\algorithmicindent} $N =$ \textit{GetActivatedExpertNum}($bs$, $E$, $K$)
                \State \hspace{\algorithmicindent} \textit{bias}, $k_1$, $k_2$, $k_3$, \textit{draft\_bias}, \textit{draft\_k}, \textit{reject\_bias}, \textit{reject\_k}, $\lambda$, \textit{s} $=$ \textit{params}
                \Comment{\textcolor{purple}{Unpack parameters}}
                
                \State \hspace{\algorithmicindent} $N_{ar} = E \cdot (1-((E-K)/E)^{B}),~~T_{ar} = B\cdot K / N_{ar}$ \Comment{\textcolor{purple}{Compute AR forward time}}
                
                \State \hspace{\algorithmicindent} \textit{ar\_time} $=$ \textit{bias} $+k_1 \cdot G(B;\lambda RP, s) + k_2 \cdot N_{ar} + k_3\cdot G(T_{ar};\lambda RP, s)$
                
                \State \hspace{\algorithmicindent} $N_{sd} = E \cdot (1-((E-K)/E)^{B\gamma}),~~T_{sd} = B\cdot \gamma\cdot K / N_{sd}$ 
                \Comment{\textcolor{purple}{Compute SD forward time}}

                \State \hspace{\algorithmicindent} \textit{verify\_time} $=$ \textit{bias} $+k_1 \cdot G(B\gamma;\lambda RP, s) + k_2 \cdot N_{sd} + k_3\cdot G(T_{sd};\lambda RP, s)$

                \State \hspace{\algorithmicindent} \textit{draft\_time} $=$ \textit{draft\_bias} + \textit{draft\_k} $\cdot~G(B;\lambda RP,s)$
                \Comment{\textcolor{purple}{Compute draft model forward time}}

                \State \hspace{\algorithmicindent} \textit{reject\_time} $=$ \textit{reject\_bias} + \textit{reject\_k} $\cdot B$
                \Comment{\textcolor{purple}{Compute rejection sampling time}}

                \State \hspace{\algorithmicindent} \textit{Speedup} $= \sigma \cdot (\gamma+1) \cdot \frac{\textit{\text{ar\_time}}}{\textit{\text{draft\_time}}+\textit{\text{ar\_time}}+\textit{\text{verify\_time}}+\text{reject\_time}}$
                \Comment{\textcolor{purple}{Compute the speedup as formalized in Eq.~\ref{eq:speedup}}}
                
                \State \hspace{\algorithmicindent} return \textit{Speedup}

            \vspace{5pt}
            
            % \State \Comment{\textcolor{purple}{Find the optimal \textit{params*} by fitting the model to the measured inputs using the least squares criterion.\qquad}}
            % \vspace{-2pt}
            \State \textit{params*} $= \underset{\textit{\text{params}}}{\operatorname{argmin}} \frac{1}{2}\displaystyle\sum_{i=1}^{m}\Big($\textit{\textbf{ComputeSpeedup}}(\textit{params}, $\mathbf{M}_i.B,~\mathbf{M}_i.\gamma,~\mathbf{M}_i.K,~\mathbf{M}_i.E,~\mathbf{M}_i.\sigma$)$-\mathbf{M}_i.$\textit{Speedup}$\Big)^2$ \Comment{\textcolor{purple}{Decide the optimal \textit{params*} by fitting the model to the measured inputs using the least squares criterion.}}

            % \State \Comment{\textcolor{purple}{Compute the final \textit{Speedup} use optimal \textit{params*}.\qquad\qquad\qquad\qquad\qquad\qquad\qquad\qquad\qquad\qquad\qquad\quad}}
            % \vspace{4pt}
            % \State \textit{Speedup} $=$ \textit{\textbf{ComputeSpeedup}}(\textit{params*}, $B, \gamma, K,E,\sigma$) \Comment{\textcolor{purple}{Compute the final \textit{Speedup} use \textit{params*}.}}

            % \State return \textit{Speedup}

      \end{algorithmic}
  \end{small}
  \end{algorithm}

% \begin{figure}
%     \begin{small}
%   \centering
%   \begin{subfigure}[b]{0.245\textwidth}
%       \centering
%       \includegraphics[width=\textwidth]{project/pic/attn_modeling_log.pdf}
%       % \caption{The number of }
%       \caption{}
%       \label{fig:logattn}
%   \end{subfigure}
%   \begin{subfigure}[b]{0.245\textwidth}
%       \centering
%       \includegraphics[width=\textwidth]{project/pic/attn_modeling.pdf}
%       % \caption{Qwen1.5-MoE-Chat}
%       \caption{}
%       \label{fig:attn}
%   \end{subfigure}
%   \begin{subfigure}[b]{0.245\textwidth}
%       \centering
%       \includegraphics[width=\textwidth]{project/pic/expnum1_moe_modeling.pdf}
%       % \caption{Deepseek-V2-Lite-Chat}
%       \caption{}
%       \label{fig:moeexp1}
%   \end{subfigure}
%   \begin{subfigure}[b]{0.245\textwidth}
%       \centering
%       \includegraphics[width=\textwidth]{project/pic/expnum8_moe_modeling.pdf}
%       % \caption{Deepseek-V2-Lite-Chat}
%       \caption{}
%       \label{fig:moeexp8}
%   \end{subfigure}
%   \end{small}
%   \caption{Profiling results of $T_{Attn}(b,s)$, $T_{FFN}(b,s)$ versus our modeling. (a) and (b) exhibit $T_{Attn}(b,s)$, with (a) adopting a log x-axis to zoom regions with small input token counts. (c) exhibits $T_{FFN}(b,s)$ when model sparsity $\rho=1/64$, and (d) for $T_{FFN}(b,s)$ with $\rho=8/64$.}
%   \label{fig:profiling}
% \end{figure}

Given the numerous factors affecting final speedup, quantitatively understanding each factor's impact is challenging. Therefore, we developed a modeling method that makes SD speedup results more \textit{explainable} and \textit{transparent}. As demonstrated by Eq.~\ref{eq:speedup}, the core of modeling SD speedup lies in characterizing the model's forward pass time. Based on theoretical analysis in previous sections, we identified three primary factors affecting forward execution: (1) the roofline model effect, (2) the number of active experts, and (3) expert load. Since GPU execution is dynamic in practice, and not all operators are optimized to their theoretical limits, we introduced several parameters for relaxation. The values of these parameters are then automatically determined by fitting GPU measurements.
% to adjust the intensity of these factors. 
% \raym{for relaxation}.
% 
% We determined values of these parameters by fitting them to real measurements. 
% 
% \raym{The values of these parameters are then automatically determined by fitting measurements from profiling.}
% 
These factors and their impacts on execution time are examined as follows.

\vspace{-5pt}
\paragraph{(1) The roofline model effect.} It manifests as execution time increases with token counts $t$, with a growth \textit{rate} that starts slow, then accelerates, and finally stabilizes. The underlying reasons are as follows. When $t$ is small, parameter loading time exceeds computation time, creating a memory bottleneck. Therefore, given the parameter volume, the memory access time is stable (memory-bound regime). As $t$ increases, computation time exceeds parameter loading time and becomes the bottleneck. With fixed arithmetic units in the hardware, computation time scales linearly with computational load (compute-bound regime). To characterize this trend, we design $G(t; \lambda RP, s)$ 
%
% \rayq{$s$ is not defined?}
% 
as shown in Eq.~\ref{eq:gt}, where $\lambda RP$ represents the transition point between regimes, and $s$ controls the increasing rate of execution time. Here, $RP$ follows Eq.~\ref{eq:AIRP}, and $\lambda$ is a constant less than 1 that accounts for practical limitations in memory bandwidth utilization. $G(t)$ exhibits a gradually increasing slope before the transition point, then shifts to a linear function afterwards, maintaining first-order gradient continuity at the transition.
\begin{equation}
\small
    G(t;\lambda RP, s)=
    \begin{cases}
    s^{t}, & t\leq \lambda RP \\
    s^{\lambda RP} + \Big(\frac{\textbf{d}(s^t)}{\textbf{d}t}|_{t=\lambda RP}\Big)(t-\lambda RP) = s^{\lambda RP}\Big(1+ln(s)\cdot (t-\lambda RP)\Big), &t>\lambda RP
    \end{cases}
    \label{eq:gt}
\end{equation}
% For dense parts in models, the roofline model effect predominates. It manifests as execution time increasing with token count $t$, with a growing rate initially slow and gradually stabilizing at a higher rate. The underlying mechanism is: when $t$ is small, computation time is shorter than parameter loading time, making memory access the bottleneck. Since parameter volume remains constant in dense models, processing time stays relatively unchanged (memory-bound regime). As $t$ increases, computation time eventually exceeds parameter loading time and becomes the bottleneck. With fixed arithmetic units in the hardware, computation time scales linearly with computational load (compute-bound regime). We design $G(t; \lambda RP, s)$ as shown in Eq.~\ref{eq:gt} to characterize this growth trend, using $\lambda RP$ as the transition point between memory-bound and compute-bound regimes. Here, $RP$ follows preliminary and $\lambda$ is a constant less than 1, considering GPUs often cannot utilize peak memory bandwidth. Before the transition point, $G(t)$ follows an exponential curve with gradually increasing slope; After the transition point,  $G(t)$ follows a linear function, maintaining first-order gradient continuity at the transition.
\vspace{-15pt}
\paragraph{(2) The number of activated experts.} When it increases, the memory access volume increases, thus adding to the final processing time. We use the derived Eq.~\ref{eq:N_full} of $N$ to characterize how workload and model architecture affects the number of activated experts.
\vspace{-10pt}
\paragraph{(3) Expert load.} This refers to the fact that after token distribution through the MoE gating, each expert processes only a subset of tokens $\overline{T_{exp}}(t;\rho)$ rather than the entire input token count $t$. Therefore, we should use $G(\overline{T_{exp}})$ rather than $G(t)$ when applying the roofline model to MoE experts. This corroborates our theoretical conclusion that sparser MoEs \textit{delay} the transition from memory-bound to compute-bound when input tokens count increases. 

For the MoE target model, factors (1), (2), and (3) are all involved. We combine them in a first-order style and introduce parameters \textit{bias}, $k_1$, $k_2$, and $k_3$ to adjust for non-ideal factors in actual GPU execution, with the full expression shown in lines 6 and 8 of Alg.~\ref{alg:main}. These parameters have clear practical meanings: \textit{bias} represents the time required to load fixed parameters; $k_2\cdot N$ represents the time needed to load $N$ activated experts; $k_1\cdot G(t)$ and $k_3\cdot G(\overline{T_{exp}})$ describe the \textit{incremental} trend in execution time as the number of tokens increases.
% Both AR forward and verification in SD essentially involve once forward of the target model, differing only in that the latter processes $\gamma$ times more tokens than the former.
For the draft model, only factor (1) is involved since it is usually dense, with the modeling form shown in line 9 of Alg.~\ref{alg:main}. 

With the expression of SD speedup determined, we fit the measurement inputs to automatically determine the relaxation parameter values, with the optimization criterion being the minimization of Mean Squared Error (MSE) between the model outputs and the ground truth, as shown in line 13 of Alg.~\ref{alg:main}. By applying these optimized parameters in our model (i.e., the \textit{ComputeSpeedup} function in line 3), we obtain the complete modeling for SD speedup. An illustrative diagram of this process and more fitting details are provided in Appendix~\ref{chap:fitting}.

Since our theoretical analyses capture the primary tradeoffs and provide a solid foundation for the modeling, the fitting is very efficient. The fitting results with 21 measurements are displayed in Fig.~\ref{fig:trend}, which show consistent trends with GPU results under various cases. These results validate the reliability of our modeling, thereby establishing it as an effective tool for analyzing the components of the model's forward pass and quantitatively understanding the tradeoffs between different factors. As shown in Sec.~\ref{exp:sim}, we explain some unexpected results with the help of the model.

\subsection{Practical Values of Theoretical Findings}
\label{chap:application}

While previous sections focuses on theoretical analysis, this section demonstrates how these findings translate to practical speedups. Our theoretical analysis has already revealed that SD speedup for MoE is most effective at \textit{moderate} batch sizes, with its trend initially increasing and then decreasing. We discuss their practical values considering both basic deployment and extended configurations.

\vspace{-8pt}
\paragraph{Basic deployment.} (1) Moderate batch sizes are common in private serving, which are increasingly adopted for data security, with representative applications like enterprise in-house chatbots. (2) When latency requirements are strict, large batch sizes are often not feasible. LLM serving must satisfy multiple service level objectives (SLOs)~\cite{goodput}, including time-to-first-token (TTFT) and time-per-output-token (TPOT). Large batches reduce per-request computational resources, causing latency violations. In such cases, moderate batch sizes are common. (3) Our work actually reveals that SD on MoE relaxes the traditional \textit{latency-throughput trade-off}. Specifically, MoE models exhibit a regime where SD speedup increases (lower latency) alongside larger batch sizes (higher throughput).

From the model's perspective, moderate batch sizes represent an "\textit{efficiency gap}" in MoEs. At this scale, all parameters must be loaded (unlike small batches with selective expert activation), yet GPU FLOPs are not fully utilized (unlike large batches). Our findings provide a novel perspective to address this efficiency challenge without compromising model quality.

\vspace{-5pt}
\paragraph{Extended configurations. } We consider typical system optimizations on MoE like \textit{offloading} and \textit{expert parallelism (EP)}. When MoE models exceed GPU memory capacity, FFN parameters are offloaded to CPU memory~\cite{ktransformers}. This degrades parameter loading bandwidth from GPU memory bandwidth to much lower PCIe bandwidth, making the system more memory-bound. Consequently, additional computation does not significantly increase processing time, creating favorable conditions for SD. Notably, existing optimizations like expert prefetching~\cite{xue2024moe,zhong2024adapmoe} and caching~\cite{he2024expertflow,tang2024hobbit} lose efficiency under moderate batch sizes since nearly all experts are activated.

Our findings are also compatible with EP. In EP, experts are distributed across multiple GPUs, which affects neither $N(t)$ nor $\overline{T_{exp}}$, making our previous analyses remain valid. Since components besides MoE FFN are also parallelized, MoE FFN continues to constitute a significant portion of processing time, allowing memory-boundness effects to remain observable in end-to-end performance. Notably, under extensive EP configurations, the inefficiency of SD for MoE at a small batch size may vanish, considering the additional memory bandwidth offered by large amounts of EP GPUs.

\section{Experiments}
\label{chap:exp}

\paragraph{Models and datasets.} We conducted experiments on two pairs of MoE target models and draft models: Qwen2-57B-A14B-Instruct with Qwen2-0.5B-Instruct~\cite{qwen2}, and Mixtral-8x7B-Instruct-v0.1~\cite{mixtral} with Eagle speculation head~\cite{li2024eagle}. They represent two prevalent SD approaches: Qwen2 employs a standalone small model from the same model family as the draft model, while Mixtral uses a specifically trained speculation head. When we need to examine MoEs with different sparsity, we modify the \texttt{num\_experts\_per\_token} in the model's config.json file. 
For comparison with dense models, we use Opt-30b and Opt-350m~\cite{zhang2022opt} as the target and draft models.
Models are evaluated on HumanEval~\cite{humaneval} and MT-bench~\cite{mtbench} datasets for code generation and conversation tasks, following previous works~\cite{li2024eagle,hass,cai2401medusa}. The tokenized prompt lengths range from 38 to 391 tokens for HumanEval and 5 to 356 tokens for MT-bench.

\paragraph{Frameworks and hardware.} 
We used the existing vllm~\cite{kwon2023vllm} framework for our experiments to verify theoretical predictions. Vllm supports batched speculative decoding, cudagraph optimization, and reports comprehensive data such as $T_D$, $T_T$, $T_{reject}$ and $\sigma$, thus being suitable for our experiments. To prevent unstable performance at the beginning, all data were obtained by averaging the results from the last five of the total ten runs. We conducted experiments on different hardware platforms including 2xGPU-A, 2xGPU-B, 4xGPU-A, 4xGPU-C.
% The number of GPUs/NPUs is determined by memory capacity requirements, as sufficient memory is needed to accommodate the entire FP16 target model with 57B parameters.

\subsection{Speedup Trend of Speculative Decoding for MoE}
\label{chap:trend}

% \begin{figure}
%   \centering
%   \includegraphics[width=\textwidth]{project/pic/real_and_metric.pdf}
%   \caption{SD speedup (left y-axis)  as a function of batch size and corresponding \metric{} values (right y-axis). Across different hardware platforms, SD acceleration first increases and then decreases, verifying our theoretical predictions. The \metric{} shows consistent trends with final speedup.}
%   \label{fig:real_and_metric}
% \end{figure}

\begin{figure}
    \begin{small}
  \centering
  \begin{subfigure}[b]{0.328\textwidth}
      \centering
      \includegraphics[width=\textwidth]{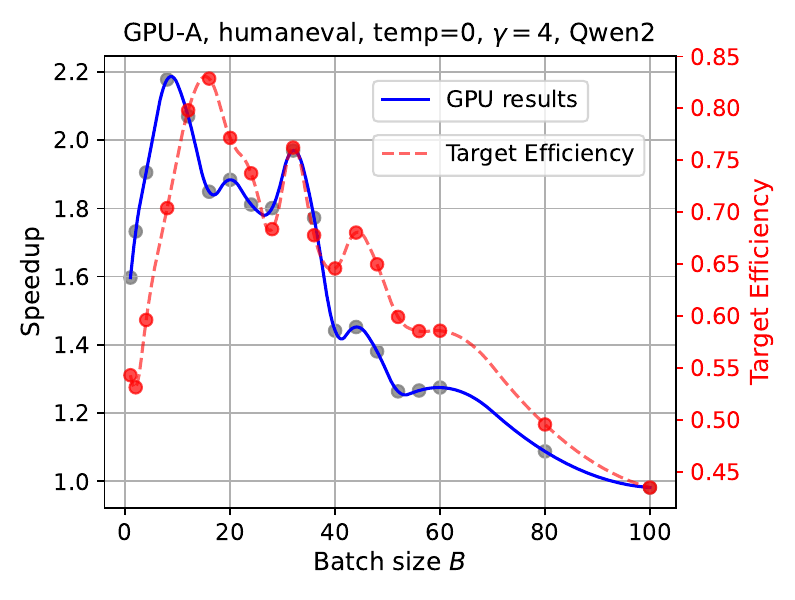}
      \vspace{-15pt}
      \caption{}
      \label{fig:a800}
  \end{subfigure}
  \begin{subfigure}[b]{0.328\textwidth}
      \centering
      \includegraphics[width=\textwidth]{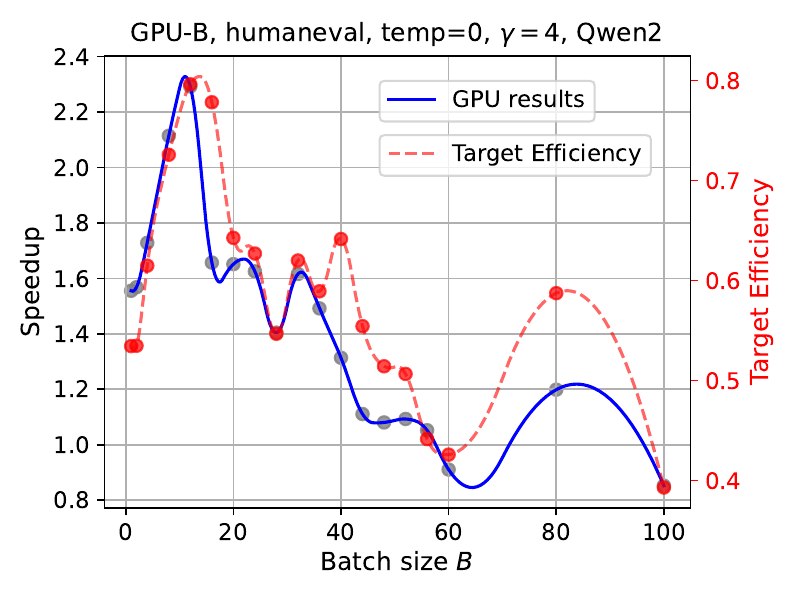}
      \vspace{-15pt}
      \caption{}
      \label{fig:h800}
  \end{subfigure}
  \begin{subfigure}[b]{0.328\textwidth}
      \centering
      \includegraphics[width=\textwidth]{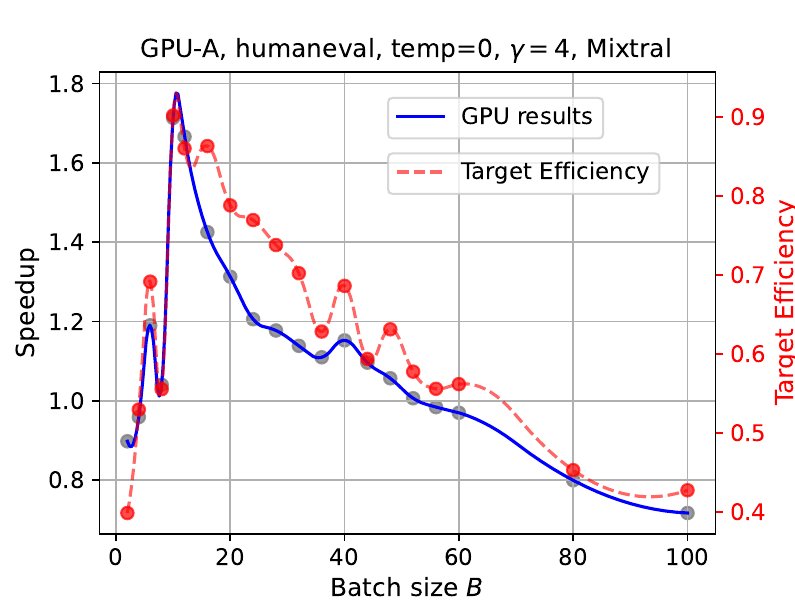}
      \vspace{-15pt}
      \caption{}
      \label{fig:dense_and_moe}
  \end{subfigure}
  \vspace{-3pt}
  \caption{SD speedup (left y-axis)  as a function of batch size and corresponding \metric{} values (right y-axis). Across different hardware platforms and MoE models, SD speedup first increases and then decreases, verifying our theoretical predictions. The \metric{} shows consistent trends with final speedup, validating its effectiveness.
  % While MoE exhibits a trend of initial increase followed by decrease, dense models show consistent decline, indicating that MoE has more idle computing power to leverage at larger batch sizes, which is advantageous for SD acceleration.
  }
  \label{fig:basic}
  \end{small}
  
\end{figure}

% \renewcommand{\arraystretch}{1.2}
% \begin{table}[]
% \caption{The peak speedup ($\mathbf{x}$) of SD across different datasets, temperatures, $\gamma$s and hardware}
% \begin{scriptsize}
% \centering
% \begin{tabular}{ll|llll|llll|llll}
%     \toprule
%     \multirow{2}{*}{{\footnotesize Dataset}} & \multirow{2}{*}{{\footnotesize Temp}} & \multicolumn{4}{c|}{$\gamma=2$}          & \multicolumn{4}{c|}{$\gamma=3$}          & \multicolumn{4}{c}{$\gamma=4$}           \\ \cline{3-14}
%                              &                             & $T_{AR}$     & $T_{SD}$      & $\sigma$   & $\mathbf{x}$ & $T_{AR}$      & $T_{SD}$     & $\sigma$   & $\mathbf{x}$ & $T_{AR}$      & $T_{SD}$     & $\sigma$   & $\mathbf{x}$ \\ \hline
%     HumanEval                & 0.0                         & 18.89   & 11.61   & 0.94    & 1.63    & 15.93   & 8.11    & 0.93    & 1.96    & 15.93   & 7.31    & 0.91    & \textbf{2.18}    \\
%     HumanEval                & 1.0                         & 19.13   & 12.93   & 0.83    & 1.48    & 21.20   & 14.09   & 0.73    & 1.50    & 19.13   & 11.14   & 0.67    & \textbf{1.72}    \\
%     MT-bench                  & 0.0                         & 20.92   & 16.70   & 0.71    & 1.25    & 16.00   & 12.43   & 0.62    & \textbf{1.29}    & 20.92   & 17.53   & 0.55    & 1.19    \\
%     MT-bench                  & 1.0                         & 21.15   & 17.33   & 0.68    & 1.22    & 19.09   & 14.83   & 0.57    & \textbf{1.29}    & 19.09   & 15.93   & 0.48    & 1.20   \\
%     \bottomrule
% \end{tabular}
% \end{scriptsize}
% \label{tab:summary}

\renewcommand{\arraystretch}{1.2}
\begin{table}[t]
\caption{\small{The peak speedup ($\mathbf{x}$) of SD across different datasets, temperatures, $\gamma$s and models on 2xGPU-A}}
\begin{tiny}
    \centering
    \begin{tabular}{c|cc|cccc|cccc|cccc}
        \toprule
        \multirow{2}{*}{Device} & \multirow{2}{*}{{Dataset}}   & \multirow{2}{*}{{Temp}} & \multicolumn{4}{c|}{$\gamma=2$}  & \multicolumn{4}{c|}{$\gamma=3$} & \multicolumn{4}{c}{$\gamma=4$}             \\  \cline{4-15}
        &       &       & $T_{AR}$  & $T_{SD}$  & $\sigma$  & $\mathbf{x}$  & $T_{AR}$  & $T_{SD}$  & $\sigma$  & $\mathbf{x}$ & $T_{AR}$  & $T_{SD}$  & $\sigma$  & $\mathbf{x}$ \\ \hline
        \multirow{4}{*}{Qwen2} & humaneval & 0.0         & 18.89              & 11.61                 & 0.94                  & 1.63                   & 15.93  & 8.11   & 0.93   & 1.96   & 15.93  & 7.31   & 0.91   & \textbf{2.18}   \\
                                & humaneval & 1.0         & 19.13              & 12.93                 & 0.83                  & 1.48                   & 21.20  & 14.09  & 0.73   & 1.50   & 19.13  & 11.14  & 0.67   & \textbf{1.72}   \\
                                & mtbench   & 0.0         & 20.92              & 16.70                 & 0.71                  & 1.25                   & 16.00  & 12.43  & 0.62   & \textbf{1.29}   & 20.92  & 17.53  & 0.55   & 1.19   \\
                                & mtbench   & 1.0         & 21.15              & 17.33                 & 0.68                  & 1.22                   & 19.09  & 14.83  & 0.57   & \textbf{1.29}   & 19.09  & 15.93  & 0.48   & 1.20   \\ \hline
        \multirow{4}{*}{Mixtral} & humaneval & 0.0         & 20.86              & 12.47                 & 0.78                  & 1.67                   & 21.00  & 12.46  & 0.66   & 1.69   & 20.86  & 11.69  & 0.58   & \textbf{1.79}   \\
                                & humaneval & 1.0         & 21.52              & 15.58                 & 0.61                  & \textbf{1.38}                   & 21.39  & 16.03  & 0.46   & 1.33   & 21.48  & 16.23  & 0.39   & 1.32   \\
                                & mtbench   & 0.0         & 21.61              & 16.10                 & 0.61                  & \textbf{1.34}                   & 21.61  & 16.43  & 0.46   & 1.32   & 21.36  & 16.89  & 0.39   & 1.26   \\
                                & mtbench   & 1.0         & 21.33              & 17.70                 & 0.53                  & \textbf{1.21}                   & 21.33  & 17.84  & 0.43   & 1.20   & 21.33  & 18.05  & 0.35   & 1.18   \\  %\hline
        \bottomrule
    \end{tabular}
\end{tiny}
\label{tab:summary_model}
\end{table}
% \end{table}

\renewcommand{\arraystretch}{1.2}
\begin{table}[t]
\caption{\small{The peak speedup ($\mathbf{x}$) of SD across different datasets, temperatures, $\gamma$s and hardware on Qwen2}}
\begin{tiny}
    \centering
    \begin{tabular}{c|cc|cccc|cccc|cccc}
        \toprule
        \multirow{2}{*}{Device} & \multirow{2}{*}{{Dataset}}   & \multirow{2}{*}{{Temp}} & \multicolumn{4}{c|}{$\gamma=2$}  & \multicolumn{4}{c|}{$\gamma=3$} & \multicolumn{4}{c}{$\gamma=4$}             \\  \cline{4-15}
        &       &       & $T_{AR}$  & $T_{SD}$  & $\sigma$  & $\mathbf{x}$  & $T_{AR}$  & $T_{SD}$  & $\sigma$  & $\mathbf{x}$ & $T_{AR}$  & $T_{SD}$  & $\sigma$  & $\mathbf{x}$ \\ \hline
        
        \multirow{4}{*}{2xGPU-B} & humaneval & 0.0         & 15.96              & 9.34                 & 0.95                  & 1.71                   & 15.96  & 7.95   & 0.93   & 2.01   & 15.96  & 6.96   & 0.90   & \textbf{2.29}   \\
                                & humaneval & 1.0         & 17.39              & 12.82                 & 0.82                  & 1.36                   & 13.20  & 8.98  & 0.74   & 1.47   & 13.20  & 7.17  & 0.75   & \textbf{1.84}   \\
                                & mtbench   & 0.0         & 24.42              & 16.74                 & 0.71                  & \textbf{1.46}                   & 24.42  & 16.84  & 0.62   & 1.45   & 24.42  & 17.05  & 0.54   & 1.43   \\
                                & mtbench   & 1.0         & 18.24              & 14.38                 & 0.67                  & \textbf{1.27}                   & 16.25  & 13.28  & 0.56   & 1.22   & 16.25  & 13.76  & 0.48   & 1.18   \\ \hline

        \multirow{4}{*}{4xGPU-A} & humaneval & 0.0         & 11.20              & 6.77                  & 0.95                  & 1.65                   & 11.20  & 5.89   & 0.93   & 1.90   & 11.20  & 5.38   & 0.90   & \textbf{2.08}   \\
                        & humaneval & 1.0         & 11.72              & 8.51                  & 0.81                  & 1.38                   & 12.05  & 8.30   & 0.73   & 1.45   & 11.23  & 7.70   & 0.67   & \textbf{1.46}   \\
                        & mtbench   & 0.0         & 11.26              & 8.92                  & 0.72                  & \textbf{1.26}                   & 11.26  & 9.10   & 0.61   & 1.24   & 11.26  & 9.82   & 0.52   & 1.15   \\
                        & mtbench   & 1.0         & 11.78              & 10.32                 & 0.67                  & 1.14                   & 11.30  & 9.42   & 0.58   & \textbf{1.20}   & 11.78  & 11.25  & 0.47   & 1.05   \\ \hline
        \multirow{4}{*}{4xGPU-C}  & humaneval & 0.0         & 17.84              & 10.00                 & 0.95                  & 1.79                   & 17.84  & 8.33   & 0.93   & 2.14   & 17.84  & 7.94   & 0.90   & \textbf{2.25}   \\
                        & humaneval & 1.0         & 17.89              & 12.27                 & 0.80                  & 1.46                   & 17.89  & 11.07  & 0.74   & 1.62   & 17.89  & 10.91  & 0.65   & \textbf{1.64}   \\
                        & mtbench   & 0.0         & 20.40              & 15.87                 & 0.71                  & 1.29                   & 20.40  & 16.22  & 0.62   & \textbf{1.26}   & 20.40  & 16.33  & 0.54   & 1.25   \\
                        & mtbench   & 1.0         & 20.58              & 16.02                 & 0.68                  & \textbf{1.28}                   & 18.11  & 14.75  & 0.54   & 1.23   & 18.11  & 15.54  & 0.48   & 1.17   \\ 
        \bottomrule
    \end{tabular}
\end{tiny}
\label{tab:summary_hardware}
\end{table}

% \begin{table}[htbp]
%   \centering
%   \caption{SD Speedup Comparison Between Dense Model and MoE Model} % 表格标题（可按需修改）
%   \begin{tiny}
%   \begin{tabular}{c|@{}*{18}{r}} 
%     \toprule % 若未加载booktabs，替换为\hline
%     batch size & 1   & 2   & 4   & 8   & 12  & 16  & 20  & 24  & 28  & 32  & 36  & 40  & 44  & 48  & 52  & 56  & 60  & 80  \\
%     \midrule % 若未加载booktabs，替换为\hline
%     SD speedup for dense model & 3.16& 2.84& 2.57& 2.30& 2.13& 1.83& 1.69& 1.56& 1.44& 1.37& 1.35& 1.23& 1.19& 1.14& 1.04& 1.02& 1.06& 0.96\\
%     SD speedup for MoE model   & 1.60& 1.73& 1.91& 2.18& 2.07& 1.85& 1.88& 1.81& 1.80& 1.97& 1.77& 1.44& 1.45& 1.38& 1.26& 1.27& 1.28& 1.09\\
%     $\frac{\text{SD speedup for MoE model}}{\text{SD speedup for dense model}}$ & 0.51& 0.61& 0.74& 0.95& 0.97& 1.01& 1.12& 1.16& 1.25& 1.44& 1.32& 1.18& 1.22& 1.21& 1.21& 1.25& 1.20& 1.13\\
%     \bottomrule % 若未加载booktabs，替换为\hline
%   \end{tabular}
%   \end{tiny}
%   \label{tab:sd_speedup_comparison} % 表格标签（用于文档内引用）
% \end{table}

Figure~\ref{fig:basic} plots the end-to-end SD speedup (left y-axis) for MoE across various settings, validating our theoretical prediction about acceleration behavior. 
As batch size grows, speedup initially increases due to expert loading saturation, and then decreases due to compute-boundness. 
We denote the maximal speedup across batch sizes as $\mathbf{x}$ and summarize the results in Table~\ref{tab:summary_model}. For both models, SD achieves higher acceleration with longer $\gamma$ for tasks with more predictable patterns (e.g., code generation) or less randomness (e.g., lower temperature), aligning with conclusions from previous research. Figure~\ref{fig:error_bar} in Appendix~\ref{chap:more_trends} further presents SD speedup trends under more settings, including individual runs and their mean to show the statistical significance of our findings.

We further evaluate Qwen2-57B-A14B-Instruct on multiple hardware platforms (Table~\ref{tab:summary_hardware}) to verify the generality of our conclusions. Combined with results of Qwen2 in Table~\ref{tab:summary_model}, two observations can be made:
(1) GPUs with higher ridge points yield larger SD speedups (e.g., 2×GPU-A vs. 2×GPU-B, 4×GPU-A vs. 4×GPU-C), since they provide more arithmetic units for verification.
(2) Scaling from 2×GPU-A to 4×GPU-A reduces absolute runtimes ($T_{AR}$ and $T_{SD}$), but the SD speedups slightly degrade. This is because the large model benefits from inter-GPU parallelization, whereas the small draft model remains single-GPU, making its relative forward cost higher.

% H100的ridge point比A100更高，

\begin{wrapfigure}{Lb}{0.35\textwidth}
  \centering
  \includegraphics[width=0.35\textwidth]{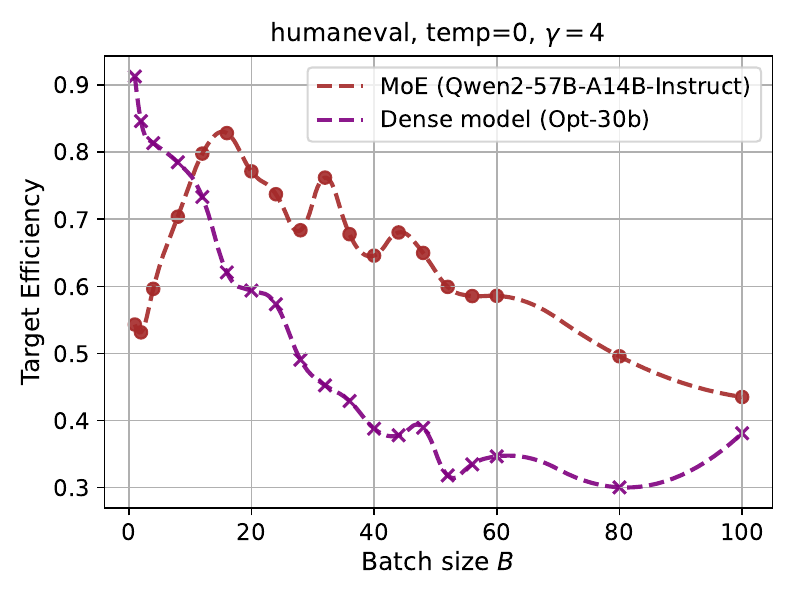}
  \caption{Comparison of \metric{}: MoE vs dense model.}
  \label{fig:dense_and_moe}
\end{wrapfigure}

Figure~\ref{fig:basic} also highlights the effectiveness of our metric \metric{}. It is computed as $\frac{T_T(B,1)}{T_T(B,\gamma)}$ as explained in Sec.~\ref{chap:general_SD_form}, where both $T_T(B,1)$ and $T_T(B,\gamma)$ are obtained from vllm runtime logs. Target efficiency values are annotated on the right y-axis, showing a consistent trend with the end-to-end speedup. In contrast, the acceptance rate across batch sizes merely \textit{fluctuates within a small range}, unable to effectively explain the dramatic changes in speedup.

 % The plot also displays the corresponding \metric{} values on the right y-axis, showing a consistent trend with overall speedup, thus confirming \metric{} as a reliable indicator of acceleration effectiveness.

We further compare the behaviors of MoE and dense models in SD. Since the effectiveness of \metric{} has been established in analyses above, and to avoid interference from algorithmic factors such as acceptance rate, we compare their \metric{}. As shown in Figure~\ref{fig:dense_and_moe}, the \metric{} for MoE first increases and then decreases, while that for the dense model decreases continuously. Consequently, although SD for MoE is less effective with small batches, it exhibits stronger potential across a wider range of larger batch sizes. Regarding end-to-end performance, SD speedups become more pronounced for MoE when the batch size exceeds 16, as supplemented in Figure~\ref{fig:e2e_moe_vs_dense} in Appendix~\ref{chap:end2end}.

% We further compare the behaviors of MoE and dense models in SD. As the effectiveness of \metric{} has been validated in previous analysis and Figure~\ref{fig:basic}, and to exclude the interference of algorithmic factors like acceptance rate, we compare their \metric{}. As shown in Figure~\ref{fig:dense_and_moe}, while the \metric{} for the MoE initially increases and then decreases, the dense model continuously decreases. As a result, although SD for MoE underperforms in small batches, it demonstrates greater potential across a wider range of larger batch sizes. Regarding the end-to-end speedup, SD speedup is more significant for MoE when the batch size is larger than 16, with experiment results shown in Figure~\ref{fig:e2e_moe_vs_dense} in Appendix~\ref{chap:end2end}.

% SD acceleration trends between MoE and dense models in Figure~\ref{fig:dense_and_moe}. To focus solely on the impact of model architecture on speedup while isolating other factors such as draft accuracy, we compare their \metric{} values rather than end-to-end speedup. The results show that \metric{} for MoE initially increases, then decreases with batch size, while \metric{} for dense models continuously decreases. Although SD for MoE underperforms in small batches, it demonstrates greater potential across a wider range of larger batch sizes.

% \begin{figure}
%   \centering
%   \includegraphics[width=\textwidth]{project/pic/dense_and_moe_comparison.pdf}
%   \caption{Comparison of \metric{} for MoE and dense models across batch sizes.}
%   \label{fig:dense_and_moe}
% \end{figure}

\subsection{Impact of MoE Sparsity and Validation of Modeling Method}
\label{exp:sim}

% \begin{figure}
% \begin{footnotesize}
%   \centering
%   \begin{subfigure}[b]{0.49\textwidth}
%       \centering
%       \includegraphics[width=\textwidth]{project/pic/simulate_half_2.pdf}
%       % \caption{The number of }
%       \caption{Draft length ($\gamma$) = 2}
%       \label{fig:simgamma2}
%   \end{subfigure}
%   \begin{subfigure}[b]{0.49\textwidth}
%       \centering
%       \includegraphics[width=\textwidth]{project/pic/simulate_half_4.pdf}
%       % \caption{Qwen1.5-MoE-Chat}
%       \caption{Draft length ($\gamma$) = 4}
%       \label{fig:simgamma4}
%   \end{subfigure}
%   \caption{Comparison between GPU results and Modeling for MoEs with varying sparsity ($\rho$)}
%   \label{fig:trend}
%   \end{footnotesize}
% \end{figure}

\begin{figure}[t]
  \centering
  \includegraphics[width=\textwidth]{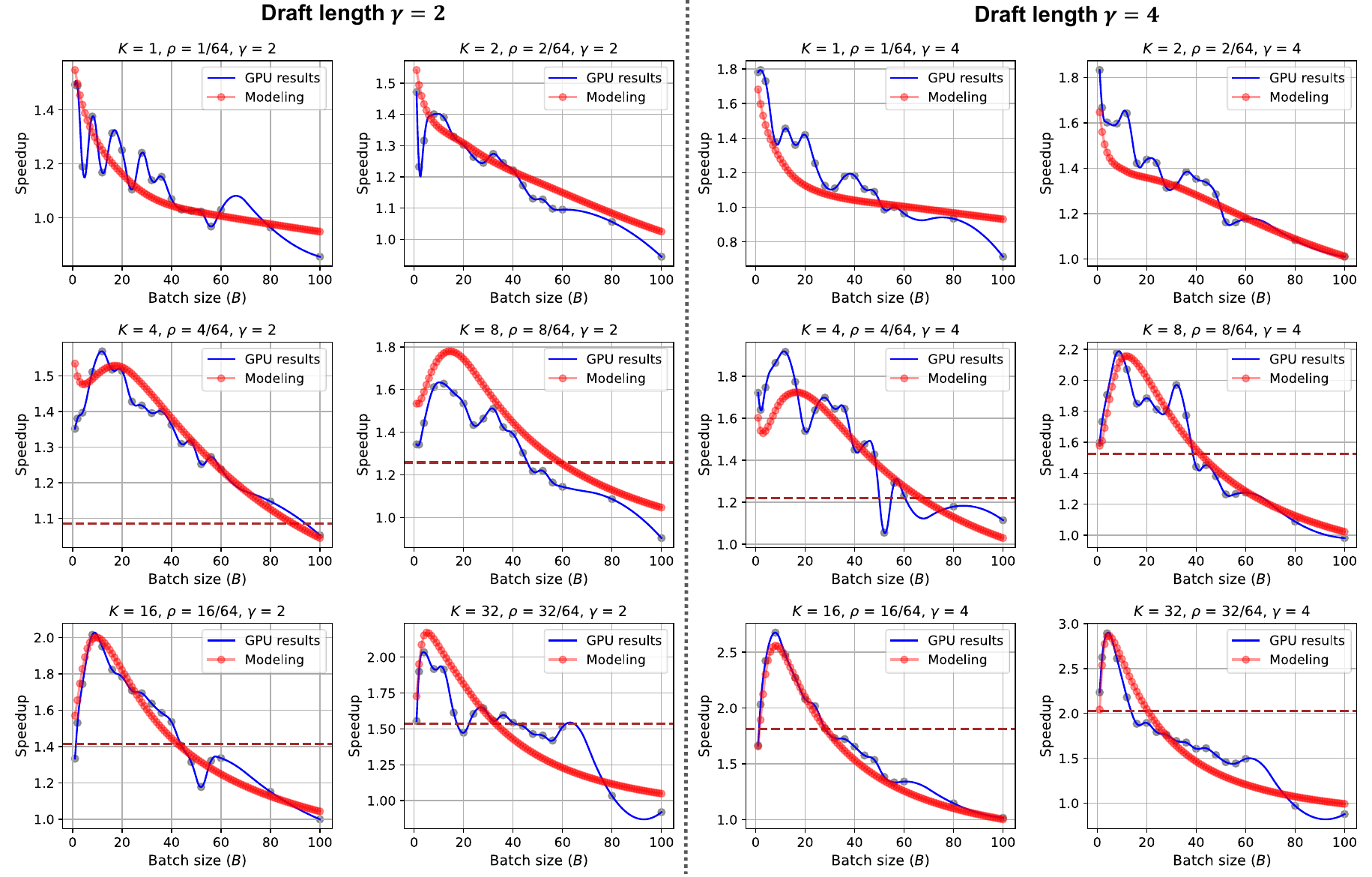}
  \vspace{-10pt}
  \caption{Comparison between GPU results and our modeling for Qwen2-57B-A14B-Instruct with varying sparsity $\rho$ and draft length $\gamma$.}
  \label{fig:trend}
\end{figure}

To evaluate MoE sparsity's impact on SD acceleration, we varied the number of activated experts per token ($K$) of Qwen2-57B-A14B-Instruct. Directly changing $K$ without training affects the target model's performance and speculation accuracy, so we adjust the speedup by multiplying the raw speedup with $\frac{\sigma_{K=8}}{\sigma_{K}}$, whose rationale is exhibited by Eq.~\ref{eq:speedup}. 
Fig.~\ref{fig:trend} shows the adjusted speedup alongside our modeling results for comparison. The parameters used in the modeling are decided using 21 GPU measurements, as explained in Section~\ref{chap:modeling}. The impact of measurement selection for parameter fitting on the modeling's reliability is supplemented in Appendix~\ref{chap:fitting}.
% We obtained a total of 228 GPU measurements as shown in Fig.~\ref{fig:trend}, from which we uniformly selected 21 points (9.2\%) for the model parameter fitting as mentioned in Section~\ref{chap:modeling}. 

There are three key observations. First, the modeling consistently aligns with experiment results across varying sparsity ($\rho$) and draft length ($\gamma$), validating our modeling's reliability. 

Second, while the SD speedup in most MoEs exhibits an initial increase followed by a decrease, very sparse MoEs ($K=1,2$) show continuously decreasing speedup. This appears to conflict with the theoretical analysis, but after examining the components of our modeling, we identified the reason as follows. These very sparse MoEs have a disproportionately low ratio of FFN, thus making the memory-boundness of MoE FFN hard to manifest systematically as indicated by Amdahl's Law. The Qwen2-57B-A14B model is designed based on $K=8$, but by reducing $K$ to 1 or 2, we actually artificially \textit{synthesize} a model where Attention dominates. In practice, however, sparser MoEs typically incorporate more FFN parameters to maintain a balanced ratio between FFN and Attention components, resulting in acceleration patterns more similar to $K=8$ cases. 

Finally, as MoE models become sparser, the system's transition from memory-bound to compute-bound is delayed. This is evidenced by two phenomena in Fig.~\ref{fig:trend}: With smaller $\rho$, (1) the batch size for the maximal speedup ($\mathbf{x}$) becomes larger; (2) the range of batch sizes that maintain speedup above a certain decay threshold (annotated by the brown dashed line in Fig.~\ref{fig:trend} for $\mathbf{x}/\sqrt{2}$) is wider. These validate our theoretical analysis and indicate that SD has broader applicability in sparser MoEs.

\section{Conclusion and Limitation}
\label{chap:conclusion}

In this work, we challenge the conventional wisdom that speculative decoding cannot effectively accelerate MoE models, and point out that with moderate batch sizes, sparser MoEs actually gain greater benefits from SD due to the more memory-bound FFN. We support this conclusion with both theoretical analysis and experimental verification. Considering the complex interplay of multiple factors affecting the final speedup, we develop a reliable modeling for SD, which enables us to comprehend the acceleration process in a transparent and explainable manner. We also introduce \metric{} to help researchers comprehensively understand how SD acceleration is affected by the target model architecture and workload. Our work offers a new perspective for MoE acceleration, particularly effective for private serving with moderate batch sizes or memory-constrained scenarios. Most of our analysis assumes the volume of KV-cache is relatively smaller than that of parameters, while the behavior of SD when KV-cache dominates has already been analyzed by MagicDec~\cite{sadhukhan2024magicdec}. These two works can be combined to offer a more comprehensive view of SD at varying batch sizes.

%我们大部分的分析适用于sequence length不是特别长的时候，with后面这种情况对SD的影响已经被MagicDec分析过。本文定性地指出了\sysname{}在offloading和分布式下的应用前景，将来工作可以对其实现得到定量的性能提升。

\begin{ack}
This work is supported by the National Natural Science Foundation of China (Grant Nos. 92267203) and Deng Feng Fund.
\end{ack}

\bibliography{ref}

\begin{thebibliography}{10}

\bibitem{liu2024deepseek}
Aixin Liu, Bei Feng, Bing Xue, Bingxuan Wang, Bochao Wu, Chengda Lu, Chenggang Zhao, Chengqi Deng, Chenyu Zhang, Chong Ruan, et~al.
\newblock Deepseek-v3 technical report.
\newblock {\em arXiv preprint arXiv:2412.19437}, 2024.

\bibitem{qwen25}
Qwen Team.
\newblock Qwen2.5 technical report.
\newblock {\em arXiv preprint arXiv:2412.15115}, 2024.

\bibitem{dai2024deepseekmoe}
Damai Dai, Chengqi Deng, Chenggang Zhao, RX~Xu, Huazuo Gao, Deli Chen, Jiashi Li, Wangding Zeng, Xingkai Yu, Yu~Wu, et~al.
\newblock Deepseekmoe: Towards ultimate expert specialization in mixture-of-experts language models.
\newblock {\em arXiv preprint arXiv:2401.06066}, 2024.

\bibitem{liu2024deepseekV2}
Aixin Liu, Bei Feng, Bin Wang, Bingxuan Wang, Bo~Liu, Chenggang Zhao, Chengqi Dengr, Chong Ruan, Damai Dai, Daya Guo, et~al.
\newblock Deepseek-v2: A strong, economical, and efficient mixture-of-experts language model.
\newblock {\em arXiv preprint arXiv:2405.04434}, 2024.

\bibitem{shazeer2017outrageously}
Noam Shazeer, Azalia Mirhoseini, Krzysztof Maziarz, Andy Davis, Quoc Le, Geoffrey Hinton, and Jeff Dean.
\newblock Outrageously large neural networks: The sparsely-gated mixture-of-experts layer.
\newblock {\em arXiv preprint arXiv:1701.06538}, 2017.

\bibitem{lepikhin2020gshard}
Dmitry Lepikhin, HyoukJoong Lee, Yuanzhong Xu, Dehao Chen, Orhan Firat, Yanping Huang, Maxim Krikun, Noam Shazeer, and Zhifeng Chen.
\newblock Gshard: Scaling giant models with conditional computation and automatic sharding.
\newblock {\em arXiv preprint arXiv:2006.16668}, 2020.

\bibitem{li2024eagle}
Yuhui Li, Fangyun Wei, Chao Zhang, and Hongyang Zhang.
\newblock Eagle: Speculative sampling requires rethinking feature uncertainty.
\newblock {\em arXiv preprint arXiv:2401.15077}, 2024.

\bibitem{moesd-inefficiency}
Anish Saxena, Po-An Tsai, Hritvik Taneja, Aamer Jaleel, and Moinuddin Qureshi.
\newblock Utility-driven speculative decoding for mixture-of-experts, 2025.

\bibitem{chen2023accelerating}
Charlie Chen, Sebastian Borgeaud, Geoffrey Irving, Jean-Baptiste Lespiau, Laurent Sifre, and John Jumper.
\newblock Accelerating large language model decoding with speculative sampling.
\newblock {\em arXiv preprint arXiv:2302.01318}, 2023.

\bibitem{leviathan2023fast}
Yaniv Leviathan, Matan Kalman, and Yossi Matias.
\newblock Fast inference from transformers via speculative decoding.
\newblock In {\em International Conference on Machine Learning}, pages 19274--19286. PMLR, 2023.

\bibitem{cai2401medusa}
Tianle Cai, Yuhong Li, Zhengyang Geng, Hongwu Peng, Jason~D Lee, Deming Chen, and Tri Dao.
\newblock Medusa: Simple llm inference acceleration framework with multiple decoding heads, 2024.
\newblock {\em URL https://arxiv. org/abs/2401.10774}, 2024.

\bibitem{privatellm1}
Rahul.
\newblock What are private llms? running large language models privately - privategpt and beyond, 2024.
\newblock \url{https://zilliz.com/learn/what-are-private-llms}.

\bibitem{privatellm2}
Esther Julie.
\newblock What is a private llm and why should you build one?, 2024.
\newblock \url{https://www.inoru.com/blog/what-is-a-private-llm-and-why-should-you-build-one/}.

\bibitem{privatellm3}
Hanbo Huang, Yihan Li, Bowen Jiang, Lin Liu, Bo~Jiang, Ruoyu Sun, Zhuotao Liu, and Shiyu Liang.
\newblock Position: On-premises llm deployment demands a middle path: Preserving privacy without sacrificing model confidentiality, 2025.

\bibitem{xie2024moepruner}
Yanyue Xie, Zhi Zhang, Ding Zhou, Cong Xie, Ziang Song, Xin Liu, Yanzhi Wang, Xue Lin, and An~Xu.
\newblock Moe-pruner: Pruning mixture-of-experts large language model using the hints from its router.
\newblock {\em arXiv preprint arXiv:2410.12013}, 2024.

\bibitem{lee2024stun}
Jaeseong Lee, Aurick Qiao, Daniel~F Campos, Zhewei Yao, Yuxiong He, et~al.
\newblock Stun: Structured-then-unstructured pruning for scalable moe pruning.
\newblock {\em arXiv preprint arXiv:2409.06211}, 2024.

\bibitem{frantar2023qmoe}
Elias Frantar and Dan Alistarh.
\newblock Qmoe: Practical sub-1-bit compression of trillion-parameter models.
\newblock {\em arXiv preprint arXiv:2310.16795}, 2023.

\bibitem{imani2024mixture}
HamidReza Imani, Abdolah Amirany, and Tarek El-Ghazawi.
\newblock Mixture of experts with mixture of precisions for tuning quality of service.
\newblock {\em arXiv preprint arXiv:2407.14417}, 2024.

\bibitem{salinas2022knowledge}
Felipe~Cruz Salinas, Kenichi Kumatani, Robert Gmyr, Linquan Liu, and Yu~Shi.
\newblock Knowledge distillation for mixture of experts models in speech recognition.
\newblock Technical report, Technical Report MSR-TR-2022-6, Microsoft Research, May 2022. https://www~…, 2022.

\bibitem{shu2024llava}
Fangxun Shu, Yue Liao, Le~Zhuo, Chenning Xu, Lei Zhang, Guanghao Zhang, Haonan Shi, Long Chen, Tao Zhong, Wanggui He, et~al.
\newblock Llava-mod: Making llava tiny via moe knowledge distillation.
\newblock {\em arXiv preprint arXiv:2408.15881}, 2024.

\bibitem{yang2024moeI}
Cheng Yang, Yang Sui, Jinqi Xiao, Lingyi Huang, Yu~Gong, Yuanlin Duan, Wenqi Jia, Miao Yin, Yu~Cheng, and Bo~Yuan.
\newblock Moe-i$^2$: Compressing mixture of experts models through inter-expert pruning and intra-expert low-rank decomposition.
\newblock {\em arXiv preprint arXiv:2411.01016}, 2024.

\bibitem{li2023merge}
Pingzhi Li, Zhenyu Zhang, Prateek Yadav, Yi-Lin Sung, Yu~Cheng, Mohit Bansal, and Tianlong Chen.
\newblock Merge, then compress: Demystify efficient smoe with hints from its routing policy.
\newblock {\em arXiv preprint arXiv:2310.01334}, 2023.

\bibitem{xue2024moe}
Leyang Xue, Yao Fu, Zhan Lu, Luo Mai, and Mahesh Marina.
\newblock Moe-infinity: Activation-aware expert offloading for efficient moe serving.
\newblock {\em arXiv e-prints}, pages arXiv--2401, 2024.

\bibitem{zhong2024adapmoe}
Shuzhang Zhong, Ling Liang, Yuan Wang, Runsheng Wang, Ru~Huang, and Meng Li.
\newblock Adapmoe: Adaptive sensitivity-based expert gating and management for efficient moe inference.
\newblock In {\em Proceedings of the 43rd IEEE/ACM International Conference on Computer-Aided Design}, pages 1--9, 2024.

\bibitem{he2024expertflow}
Xin He, Shunkang Zhang, Yuxin Wang, Haiyan Yin, Zihao Zeng, Shaohuai Shi, Zhenheng Tang, Xiaowen Chu, Ivor Tsang, and Ong~Yew Soon.
\newblock Expertflow: Optimized expert activation and token allocation for efficient mixture-of-experts inference.
\newblock {\em arXiv preprint arXiv:2410.17954}, 2024.

\bibitem{tang2024hobbit}
Peng Tang, Jiacheng Liu, Xiaofeng Hou, Yifei Pu, Jing Wang, Pheng-Ann Heng, Chao Li, and Minyi Guo.
\newblock Hobbit: A mixed precision expert offloading system for fast moe inference.
\newblock {\em arXiv preprint arXiv:2411.01433}, 2024.

\bibitem{miao2024specinfer}
Xupeng Miao, Gabriele Oliaro, Zhihao Zhang, Xinhao Cheng, Zeyu Wang, Zhengxin Zhang, Rae Ying~Yee Wong, Alan Zhu, Lijie Yang, Xiaoxiang Shi, et~al.
\newblock Specinfer: Accelerating large language model serving with tree-based speculative inference and verification.
\newblock In {\em Proceedings of the 29th ACM International Conference on Architectural Support for Programming Languages and Operating Systems, Volume 3}, pages 932--949, 2024.

\bibitem{he2023rest}
Zhenyu He, Zexuan Zhong, Tianle Cai, Jason~D Lee, and Di~He.
\newblock Rest: Retrieval-based speculative decoding.
\newblock {\em arXiv preprint arXiv:2311.08252}, 2023.

\bibitem{svirschevski2024specexec}
Ruslan Svirschevski, Avner May, Zhuoming Chen, Beidi Chen, Zhihao Jia, and Max Ryabinin.
\newblock Specexec: Massively parallel speculative decoding for interactive llm inference on consumer devices.
\newblock {\em Advances in Neural Information Processing Systems}, 37:16342--16368, 2024.

\bibitem{li2024eagle2}
Yuhui Li, Fangyun Wei, Chao Zhang, and Hongyang Zhang.
\newblock Eagle-2: Faster inference of language models with dynamic draft trees.
\newblock {\em arXiv preprint arXiv:2406.16858}, 2024.

\bibitem{li2025eagle3}
Yuhui Li, Fangyun Wei, Chao Zhang, and Hongyang Zhang.
\newblock Eagle-3: Scaling up inference acceleration of large language models via training-time test.
\newblock {\em arXiv preprint arXiv:2503.01840}, 2025.

\bibitem{batchsd2}
Xiaoxuan Liu, Jongseok Park, Langxiang Hu, Woosuk Kwon, Zhuohan Li, Chen Zhang, Kuntai Du, Xiangxi Mo, Kaichao You, Alvin Cheung, Zhijie Deng, Ion Stoica, and Hao Zhang.
\newblock Turbospec: Closed-loop speculation control system for optimizing llm serving goodput, 2025.

\bibitem{batchsd3}
Qidong Su, Christina Giannoula, and Gennady Pekhimenko.
\newblock The synergy of speculative decoding and batching in serving large language models, 2023.

\bibitem{sadhukhan2024magicdec}
Ranajoy Sadhukhan, Jian Chen, Zhuoming Chen, Vashisth Tiwari, Ruihang Lai, Jinyuan Shi, Ian En-Hsu Yen, Avner May, Tianqi Chen, and Beidi Chen.
\newblock Magicdec: Breaking the latency-throughput tradeoff for long context generation with speculative decoding.
\newblock {\em arXiv preprint arXiv:2408.11049}, 2024.

\bibitem{llmviewer}
Zhihang Yuan, Yuzhang Shang, Yang Zhou, Zhen Dong, Zhe Zhou, Chenhao Xue, Bingzhe Wu, Zhikai Li, Qingyi Gu, Yong~Jae Lee, Yan Yan, Beidi Chen, Guangyu Sun, and Kurt Keutzer.
\newblock Llm inference unveiled: Survey and roofline model insights, 2024.

\bibitem{roofline}
Georg Ofenbeck, Ruedi Steinmann, Victoria Caparros, Daniele~G Spampinato, and Markus P{\"u}schel.
\newblock Applying the roofline model.
\newblock In {\em 2014 IEEE International Symposium on Performance Analysis of Systems and Software (ISPASS)}, pages 76--85. IEEE, 2014.

\bibitem{fedus2022loadbalance1}
William Fedus, Barret Zoph, and Noam Shazeer.
\newblock Switch transformers: Scaling to trillion parameter models with simple and efficient sparsity, 2022.

\bibitem{goodput}
Zhibin Wang, Shipeng Li, Yuhang Zhou, Xue Li, Rong Gu, Nguyen Cam-Tu, Chen Tian, and Sheng Zhong.
\newblock Revisiting slo and goodput metrics in llm serving, 2024.

\bibitem{ktransformers}
kvcache ai.
\newblock Ktransfromers: A flexible framework for experiencing cutting-edge llm inference optimizations, 2025.
\newblock \url{https://github.com/kvcache-ai/ktransformers/tree/main}.

\bibitem{qwen2}
An~Yang, Baosong Yang, Binyuan Hui, Bo~Zheng, Bowen Yu, Chang Zhou, Chengpeng Li, Chengyuan Li, Dayiheng Liu, Fei Huang, Guanting Dong, Haoran Wei, et~al.
\newblock Qwen2 technical report, 2024.

\bibitem{mixtral}
Albert~Q. Jiang, Alexandre Sablayrolles, Antoine Roux, Arthur Mensch, Blanche Savary, Chris Bamford, Devendra~Singh Chaplot, Diego de~las Casas, Emma~Bou Hanna, Florian Bressand, Gianna Lengyel, Guillaume Bour, Guillaume Lample, Lélio~Renard Lavaud, Lucile Saulnier, Marie-Anne Lachaux, Pierre Stock, Sandeep Subramanian, Sophia Yang, Szymon Antoniak, Teven~Le Scao, Théophile Gervet, Thibaut Lavril, Thomas Wang, Timothée Lacroix, and William~El Sayed.
\newblock Mixtral of experts, 2024.

\bibitem{zhang2022opt}
Susan Zhang, Stephen Roller, Naman Goyal, Mikel Artetxe, Moya Chen, Shuohui Chen, Christopher Dewan, Mona Diab, Xian Li, Xi~Victoria Lin, et~al.
\newblock Opt: Open pre-trained transformer language models.
\newblock {\em arXiv preprint arXiv:2205.01068}, 2022.

\bibitem{humaneval}
Mark Chen, Jerry Tworek, Heewoo Jun, Qiming Yuan, Henrique~Ponde de~Oliveira~Pinto, Jared Kaplan, Harri Edwards, Yuri Burda, Nicholas Joseph, Greg Brockman, Alex Ray, et~al.
\newblock Evaluating large language models trained on code, 2021.

\bibitem{mtbench}
Lianmin Zheng, Wei-Lin Chiang, Ying Sheng, Siyuan Zhuang, Zhanghao Wu, Yonghao Zhuang, Zi~Lin, Zhuohan Li, Dacheng Li, Eric~P. Xing, Hao Zhang, Joseph~E. Gonzalez, and Ion Stoica.
\newblock Judging llm-as-a-judge with mt-bench and chatbot arena, 2023.

\bibitem{hass}
Lefan Zhang, Xiaodan Wang, Yanhua Huang, and Ruiwen Xu.
\newblock Learning harmonized representations for speculative sampling, 2025.

\bibitem{kwon2023vllm}
Woosuk Kwon, Zhuohan Li, Siyuan Zhuang, Ying Sheng, Lianmin Zheng, Cody~Hao Yu, Joseph~E. Gonzalez, Hao Zhang, and Ion Stoica.
\newblock Efficient memory management for large language model serving with pagedattention.
\newblock In {\em Proceedings of the ACM SIGOPS 29th Symposium on Operating Systems Principles}, 2023.

\bibitem{nvidia-mm}
NVIDIA~Docs Hub.
\newblock Matrix multiplication background user's guide.
\newblock \url{https://docs.nvidia.com/deeplearning/performance/dl-performance-matrix-multiplication/index.html}.

\bibitem{flashattention3}
Jay Shah, Ganesh Bikshandi, Ying Zhang, Vijay Thakkar, Pradeep Ramani, and Tri Dao.
\newblock Flashattention-3: Fast and accurate attention with asynchrony and low-precision, 2024.

\bibitem{nanoflow}
Kan Zhu, Yufei Gao, Yilong Zhao, Liangyu Zhao, Gefei Zuo, Yile Gu, Dedong Xie, Tian Tang, Qinyu Xu, Zihao Ye, Keisuke Kamahori, Chien-Yu Lin, Ziren Wang, Stephanie Wang, Arvind Krishnamurthy, and Baris Kasikci.
\newblock Nanoflow: Towards optimal large language model serving throughput, 2025.

\end{thebibliography}
\bibliographystyle{unsrt}

\newpage
\appendix

\section{Supplementary Experimental Results}

This section presents additional experimental results referenced in Section~\ref{chap:exp}, which are included here due to space limitations.

\subsection{Trends of SD speedup under more configurations}
\label{chap:more_trends}

\begin{figure}
  \centering
  \includegraphics[width=\textwidth]{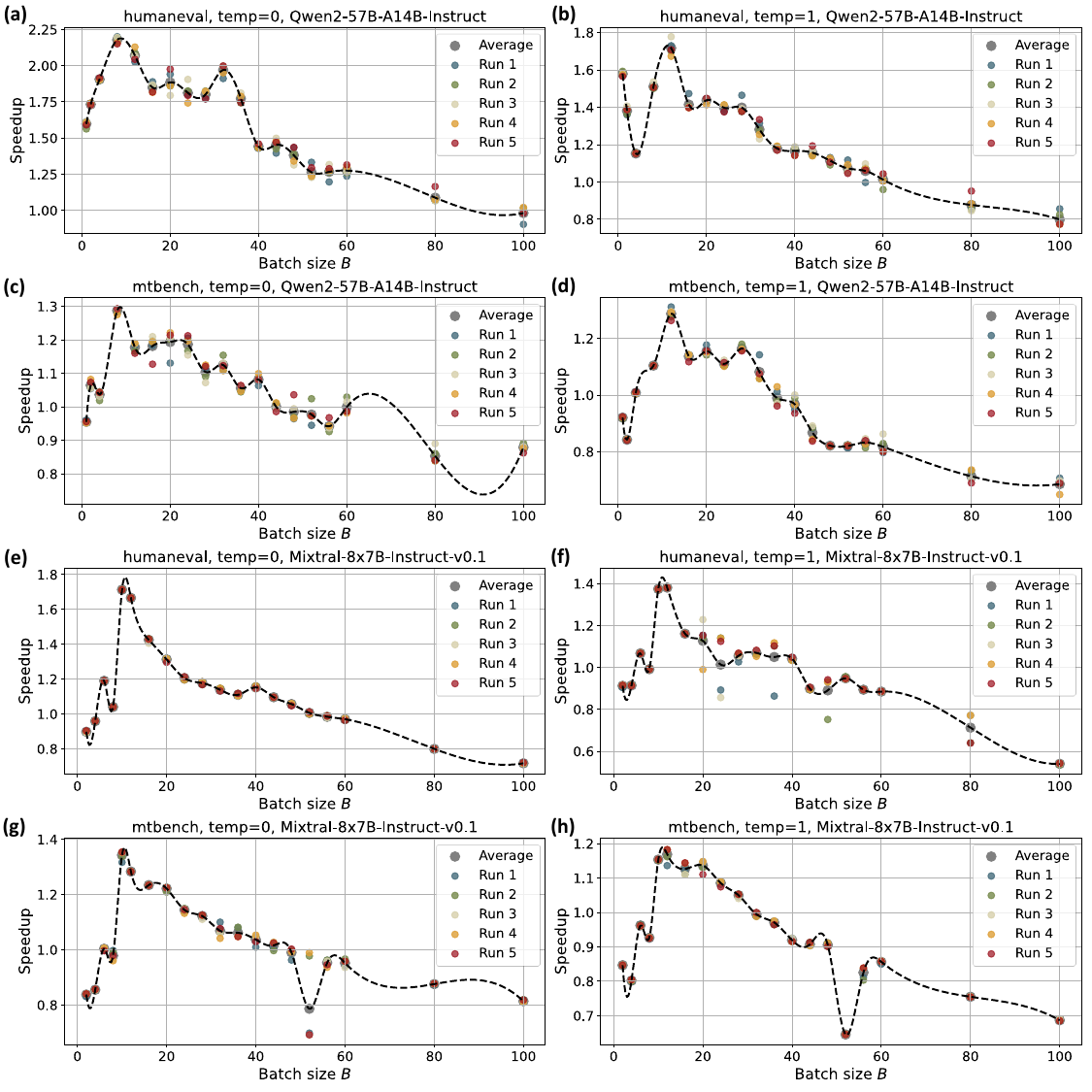}
  \caption{SD speedup trends across more settings with individual runs and averages shown.}
  \label{fig:error_bar}
\end{figure}

Figure~\ref{fig:error_bar} presents additional trends of SD speedup across different datasets, temperatures, and model types, serving as a supplement to Figure~\ref{fig:basic}. The results demonstrate that SD speedup exhibits a consistent first-increase-then-decrease pattern, which aligns well with our theoretical analysis.

To confirm the statistical significance of our findings, we also present the five individual runs that constitute the averages in Figure~\ref{fig:error_bar}. The variance across different runs is minimal, which is expected since the random seed is fixed across all runs to ensure identical workloads.

While the overall trend follows the first-increase-then-decrease pattern, local fluctuations are observable in the curves. For instance, Figure~\ref{fig:error_bar}(c) exhibits a sawtooth-like decreasing trend. This phenomenon can be attributed to GPU \textit{quantization effects}, as documented in NVIDIA's documentation~\cite{nvidia-mm}. When dimensions are not evenly divisible by the GPU's native tile sizes, computational performance degrades. AR decoding is more sensitive to this effect than SD, making the time ratio of AR to SD (namely, SD speedup) fluctuate. Despite these local variations, the overall speedup trend follows our theoretical predictions, confirming the validity of our conclusions.

\subsection{End-to-end speedup comparison between MoE and dense models}
\label{chap:end2end}

\begin{figure}
  \centering
  \includegraphics[width=\textwidth]{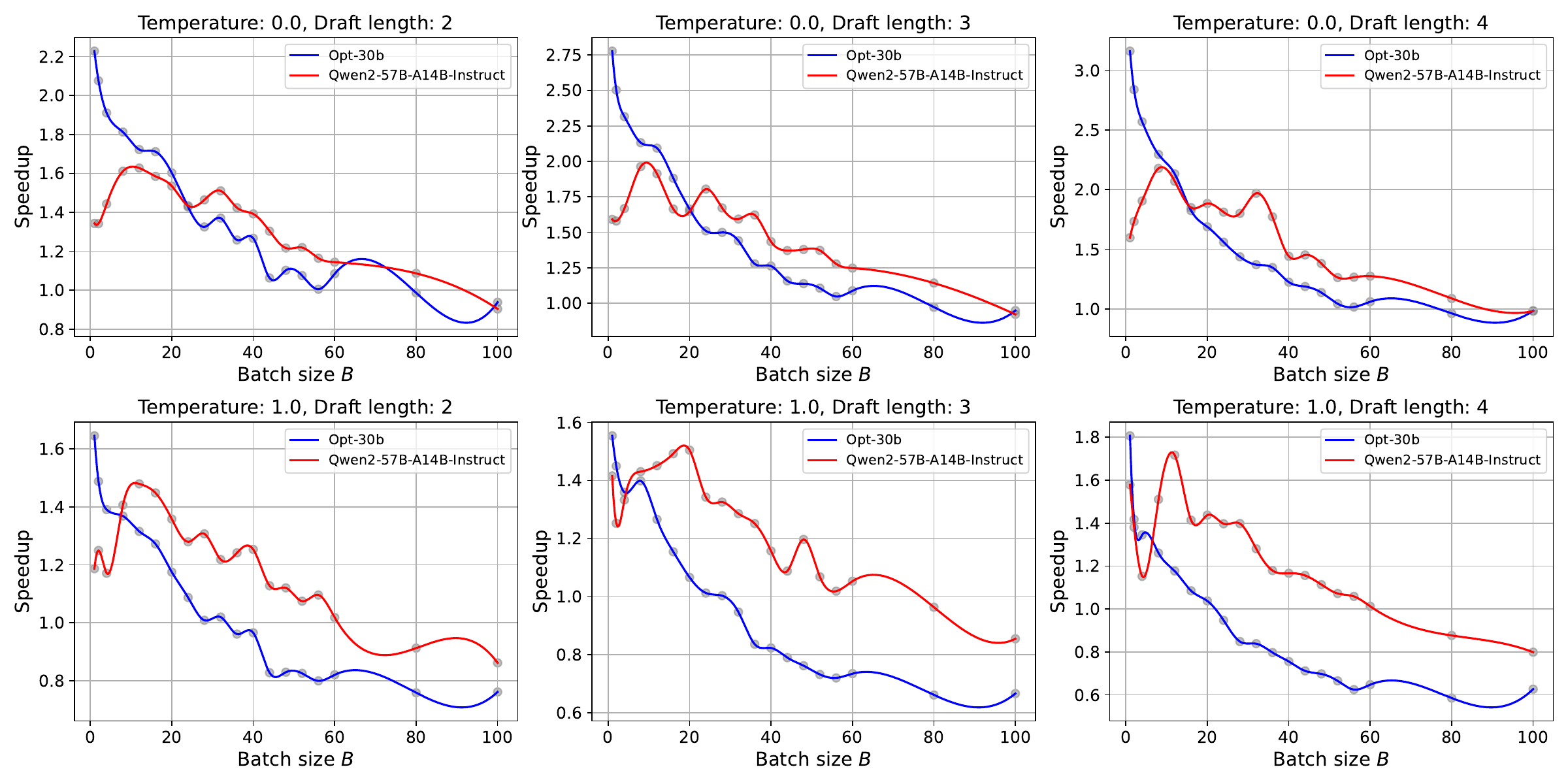}
  \caption{End-to-end speedup comparison of MoE and dense models under various settings.}
  \label{fig:e2e_moe_vs_dense}
\end{figure}

To isolate the effects of acceptance rate variations and enable a clearer focus on system bottlenecks, we have compared MoE and dense models using \metric{} in Section~\ref{chap:trend}. In this section, we further compare their end-to-end speedup across various settings in Figure~\ref{fig:e2e_moe_vs_dense} as a supplement.

Two key observations emerge from the experiment results. First, while SD speedups for MoE models initially increase before declining, SD speedups for dense models continue to decrease. Consequently, SD achieves more substantial end-to-end speedups for MoE models at moderate batch sizes, which aligns with the trend in Figure~\ref{fig:dense_and_moe} in Section~\ref{chap:trend}. Second, the extent to which SD favors MoE over dense models varies across different configurations. For instance, at temperature = 1 (second row), SD demonstrates greater relative benefits for MoE compared to temperature = 0 (first row). This variation stems from diverse acceptance rates under different settings, which can obscure the observation of systemic bottlenecks. In summary, \metric{} serves as a reliable comparison metric while controlling for the confounding effects of algorithmic optimizations.

\section{Proof of $\overline{T_{exp}}(T;\rho)$'s Trend with Varying $\rho$}
\label{app:prove}

% In Section~\ref{chap:MoeAnlysis}, we show through Fig.~\ref{fig:Texp} and mention the conclusion that: Given input token count $t=T>1$, the number of tokens each expert process on average $\overline{T_{exp}}(T;\rho)$ decreases as $\rho$ decreases. We prove this by showing $\frac{\textbf{d}(\overline{T_{exp}}(T;\rho))}{\textbf{d}\rho} > 0$ when $T > 1$.
% \begin{equation}
%     \frac{\textbf{d}(\overline{T_{exp}}(T;\rho))}{\textbf{d}\rho} = \frac{\textbf{d}(\frac{\rho T}{1-(1-\rho)^T})}{\textbf{d}\rho} = \frac{T(-\rho T(1-\rho)^{T-1}-(1-\rho)^T+1)}{(1-(1-\rho)^T)^2}\\
% \end{equation}
% Since $\rho$ stands for MoE sparsity $\in (0,1)$, the original proposition is equivalent to proving:
% \begin{equation}
%     \mathbf{F}(\rho;T) = (1-\rho)^{T-1}(\rho T+1-\rho) < 1
%     \label{eq:1stgradientconvert}
% \end{equation}
% Noticing $F(\rho;T) \to 1$ when $\rho \to 0$. Therefore, if we can prove $\mathbf{F}(\rho;T)$ decreases as $\rho$ increases, then the original proposition is proven. We prove this by computing $\frac{\textbf{d}(\mathbf{F}(\rho;T))}{\textbf{d}\rho}$:
% \begin{equation}
%     \frac{\textbf{d}(\mathbf{F}(\rho;T))}{\textbf{d}\rho} = \frac{\textbf{d}((1-\rho)^{T-1}(\rho T+1-\rho))}{\textbf{d}\rho} = -\rho T(T-1)(1-\rho)^{T-2}\\
% \end{equation}
% When $T>1$, $\frac{\textbf{d}(\mathbf{F}(\rho;T))}{\textbf{d}\rho} < 0$. Therefore, the original proposition that, when $T>1$, $\overline{T_{exp}}(T;\rho)$ decreases as $\rho$ decreases is true.

In Section 3.2, Fig. 1(c) demonstrates that: Given input token count $t=T>1$, the number of tokens each expert processes on average $\overline{T_{\text{exp}}}(T;\rho) = \frac{\rho T}{1-(1-\rho)^T}$ decreases as $\rho$ decreases. We prove this by showing $\frac{\mathrm{d}(\overline{T_{\text{exp}}}(T;\rho))}{\mathrm{d}\rho} > 0$ when $T > 1$.
\begin{equation}
    \frac{\mathrm{d}(\overline{T_{\text{exp}}}(T;\rho))}{\mathrm{d}\rho} = \frac{\mathrm{d}(\frac{\rho T}{1-(1-\rho)^T})}{\mathrm{d}\rho} = \frac{T(-\rho T(1-\rho)^{T-1}-(1-\rho)^T+1)}{(1-(1-\rho)^T)^2}
\end{equation}
Since $\rho$ represents MoE sparsity $\in (0,1)$, the original proposition is equivalent to proving:
\begin{equation}
    \mathbf{F}(\rho;T) = (1-\rho)^{T-1}(\rho T+1-\rho) < 1
    \label{eq:1stgradientconvert}
\end{equation}
Note that $\mathbf{F}(\rho;T) \to 1$ as $\rho \to 0$. Therefore, if we can prove that $\mathbf{F}(\rho;T)$ decreases as $\rho$ increases from 0 to 1, then the original proposition is proven. We demonstrate this by computing $\frac{\mathrm{d}(\mathbf{F}(\rho;T))}{\mathrm{d}\rho}$:
\begin{equation}
    \frac{\mathrm{d}(\mathbf{F}(\rho;T))}{\mathrm{d}\rho} = \frac{\mathrm{d}((1-\rho)^{T-1}(\rho T+1-\rho))}{\mathrm{d}\rho} = -\rho T(T-1)(1-\rho)^{T-2}
\end{equation}
When $T>1$, $\frac{\mathrm{d}(\mathbf{F}(\rho;T))}{\mathrm{d}\rho} < 0$. This confirms that $\mathbf{F}(\rho;T)$ decreases as $\rho$ increases, which proves our original proposition: when $T>1$, $\overline{T_{\text{exp}}}(T;\rho)$ decreases as $\rho$ decreases. 
% This means that, given the input token count, the number of tokens each expert processes on average decreases as the MoE becomes sparser.

\section{More details about the Modeling Method}
\label{chap:fitting}

% As mentioned in Section 3.3, since GPU execution is dynamic in practice, and not all operators are optimized to their theoretical limits, we introduced several parameters for relaxation. The values of these parameters are then automatically determined by fitting to GPU measurements under the least squares criterion. The results shown in Figure 3 in Section 4.2 were obtained by fitting to 21 GPU measurements. In this Section, we further explain:

The main design considerations and expressions of our modeling method have been presented in Section~\ref{chap:modeling}. In this section, we provide additional content on the following topics to give a more comprehensive view of the modeling method:

\begin{itemize}
    \item Description and an illustrative diagram of the modeling process. (Appendix~\ref{chap:overview})
    \item Fitting Details of the modeling shown in Figure~\ref{fig:trend} in Section~\ref{chap:trend}. (Appendix ~\ref{chap:21_items})
    \item How the modeling is affected by alternative measurement selection. (Appendix ~\ref{chap:other_items})
\end{itemize}
The value of our modeling is twofold. On one hand, it achieves alignment with real measurements with only a small number of simple parameters, thus validating the correctness of our theoretical analyses. On the other hand, it provides the decomposition of various factors in the end-to-end results, making the entire SD acceleration process explainable and transparent.

\subsection{Description and Overview of the Modeling Process}
\label{chap:overview}

\begin{figure}
  \centering
  \includegraphics[width=\textwidth]{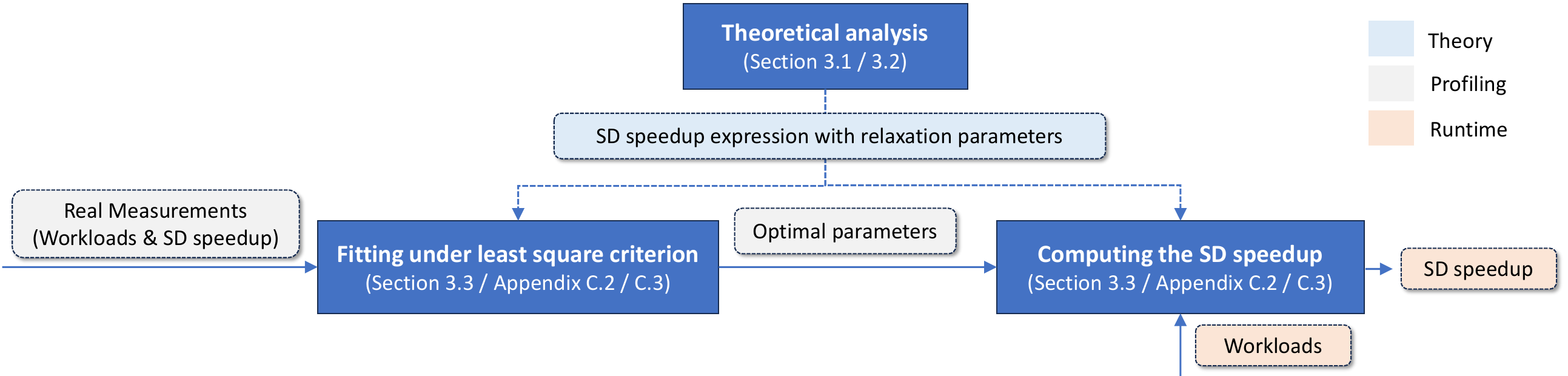}
  \caption{The overall diagram of the modeling method.}
  \label{fig:system_diagram}
\end{figure}

Figure~\ref{fig:system_diagram} presents the overall diagram of our modeling method. Building upon the theoretical analyses in Sections~\ref{chap:general_SD_form} and~\ref{chap:MoeAnlysis}, we derive an expression for SD speedup as a function of workloads. This expression contains several relaxation parameters to be determined for complete modeling. We determine these parameters through empirical profiling. We first collect a small set of real measurements comprising various workloads and their corresponding SD speedups. We then perform parameter fitting using these measurements under the least squares criterion to obtain optimal parameter values. Since our theoretically-derived SD speedup expression already captures the fundamental performance tradeoffs, the fitting process is computationally lightweight and robust, as will be demonstrated in Appendix~\ref{chap:21_items} and~\ref{chap:other_items}. Once the optimal parameters are obtained, the resulting expression can predict SD speedups for arbitrary workloads.

% Figure~\ref{fig:system_diagram} is the overall diagram of our modeling method. Building on the theoretical analyses in Section~\ref{chap:general_SD_form} and~\ref{chap:MoeAnlysis}, we get the expression of SD speedup given the workload. The expression contains several relaxation parameters, which need to be decided for a complete expression. We decide them by profiling. Specifically, we get a small amount of real measurements, containing workloads and their corresponding SD speedups. We then fit them into our expression to get the optimal parameters, under the least squares criterion. Since the SD speedup expression based on theoretical analysis already catches the main tradeoff, the fitting process is quite lightweight and robust, as demonstrated in the following Appendix. After obtaining the optimal parameters, we apply them to the expression, which can then yield SD speedups given workloads. 

We now explain why profiling is necessary and why we cannot derive the complete SD speedup expression purely through theoretical analysis, examining both software and hardware considerations.

\textbf{Software considerations:} Actual execution times can deviate significantly from theoretical predictions for complex operators with diverse implementations. On GPUs, GEMM operations are indeed predictable due to their regular structure and highly optimized implementations. However, prediction becomes challenging for operators such as Attention, which involve customized kernel optimizations (e.g., FlashAttention1/2, eager attention, SDPA attention) and operator fusion strategies (incorporating various nonlinear layers or positional encodings such as RoPE and its variants). To illustrate this complexity, we examine profiling results from Qwen2-57B-A14B (hidden size 3584) and Mixtral-8x7B (hidden size 4096). For FFN, Qwen takes a shorter time than Mixtral (143us vs 226us), aligning with their relative hidden sizes. However, for Attention, Qwen takes a longer time than Mixtral (271us vs 115us), contradicting the theoretical expectation based on hidden size scaling.

% Software aspect: The actual execution time can deviate from the theoretical limits for some complicated operators with diverse implementations. On GPUs, GEMM is indeed predictable because it involves regular and highly optimized operators. However, this becomes difficult for operators like Attention that involve customized kernel optimizations (FlashAttention1/2, eager attention, SDPA attention) and fusion (of different types of nonlinear layers or positional encodings like RoPE and its variants), as pointed out by previous works [1,2,3]. We take profiling results of Qwen2-57B-A14B (hidden size 3584) and Mixtral-8x7B (hidden size 4096) as an example to illustrate this. For FFN, Qwen takes a shorter time than Mixtral (143μs vs 226μs), aligning with their relative hidden sizes. However, for Attention, Qwen takes a longer time than Mixtral (271μs vs 115μs), contradicting the theoretical expectation based on hidden size scaling.

\textbf{Hardware considerations:} GPU microarchitectures vary across different series, which can greatly impact execution times. For instance, attention efficiency depends heavily on hardware-aware programming optimizations, while different GPUs vary in cache configurations and thread-memory interaction patterns across memory hierarchy levels. By taking advantage of new capabilities in modern hardware, FlashAttention-3 successfully increases GPU utilization from 35\% to 75\% on H100 GPUs~\cite{flashattention3}. Moreover, many critical hardware details remain undisclosed by GPU vendors, making theoretical predictions impractical.

% Hardware aspect: the GPU microarchitecture can also have a great impact on the execution time, but it varies among different series of GPUs. For example, the efficiency of flash-attention greatly relies on the cache sizes, while different GPUs varies in cache sizes and how their threads interact with different levels of memory. oreover, many hardware details are not disclosed by GPU vendors, making such predictions even less practical.

Therefore, modeling speedup trends with pure analytical methods requires \textit{case-by-case} analysis for different operator implementations and GPU microarchitectures. In contrast, the hyperparameter approach offers a more generalizable paradigm and is easy to use: all parameters possess clear physical interpretations, only minimal profiling data are required, and the computational overhead is low. Our method achieves a balance between effectiveness and practicality: on one hand, it captures the primary performance drivers (i.e. the number of activated experts and roofline trends); on the other hand, it avoids getting entangled in low-level implementation complexities. This approach is also used by other system optimization frameworks such as NanoFlow~\cite{nanoflow}, which similarly adopt a two-stage strategy of profiling followed by runtime execution.

\subsection{Fitting Details of the Modeling shown in Figure~\ref{fig:trend}}
\label{chap:21_items}

We first explain how we select the 21 measurements. Due to GPU resource and time constraints, we obtained a total of 228 GPU measurements across varying experimental settings, including 6 different numbers of activated experts per token ($K$), 2 draft lengths ($\gamma$), and 19 batch sizes ($B$). These measurements are sorted first by $K$, then by $\gamma$ within each $K$ group, and finally by $B$ within each $(K, \gamma)$ combination, forming the total dataframe (\texttt{df}). We then uniformly sampled measurements from this sorted dataset with a fixed stride, namely \texttt{M = df[begin:end:11]}. This sampling strategy enables our selected measurements to contain different settings, making the modeling more robust.

The SD speedup function (namely, \textit{ComputeSpeedup} defined in line 3 of Algorithm 1) is nonlinear. To optimize its MSE, we employed the \texttt{scipy.optimize.least\_squares} function with the Trust Region Reflective (TRR) algorithm. TRR is an optimization method for bound-constrained nonlinear least squares problems that combines trust region methods with reflection techniques. It constructs quadratic models within trust regions and uses reflection strategies near boundaries to maintain feasibility while ensuring convergence. The fitting process for these 21 data points is efficient, completing in approximately 0.114 seconds.
Our modeling incorporates 10 parameters requiring relaxation, with their search boundaries specified as follows:

\begin{itemize}[leftmargin=1cm]
    \item \textit{bias}: It represents the time required to load the dense parameters of the target model. We denote the model's non-FFN parameter count as $V_{dense}$. Consequently, the theoretical minimum loading time can be calculated as $\text{\textit{bias}}_{min}=\frac{V_{dense}\times bitwidth}{\text{\textit{peak memory bandwidth}}}$. For the upper bound of the relaxation range, we set $\text{\textit{bias}}_{max} = 5\times \text{\textit{bias}}_{min}$.

    \item \textit{k1}: It adjusts the intensity of the roofline effect of dense components. 
    It should be larger than 0 to ensure the execution time increases as the token count increases. 
    We don't set a definite upper limit for $k1$, as its value is affected by other parameters. Given the hardware with fixed arithmetic units, the execution time grows linearly with the token count in the compute-bound regime. As shown in line 6 of Algorithm 1, $k1$ appears as a coefficient in the term $k1\cdot G(t;\lambda, s)$, whose gradient in the compute-bound regime is $k1\cdot ln(s)\cdot s^{\lambda RP}$. As $s$ approaches 1, $k1$ needs to continuously increase to counterbalance $ln(s)$ that approaches 0.

    \item \textit{k2}: It represents the time required to load one expert. Given a target model, we denote the parameter count per expert as $V_{exp}$. Consequently, the theoretical minimum loading time can be calculated as $\text{\textit{k2}}_{min}=\frac{V_{exp}\times bitwidth}{\text{\textit{peak memory bandwidth}}}$. For the upper bound of the relaxation range, we set $\text{\textit{k2}}_{max} = 5\times k_{2min}$.

    \item \textit{k3}: It adjusts the intensity of the roofline effect of sparse components. Similar to \textit{k1}, we set $\text{\textit{k3}}_{min} = 0$ and $\text{\textit{k3}}_{max} = \text{\textit{inf}}$.

    \item \textit{draft\_bias}: It represents the time required to load the dense draft model. We denote the draft model's parameter count as $V_{draft}$. Consequently, the theoretical minimum loading time can be calculated as $\text{\textit{draft\_bias}}_{min}=\frac{V_{draft}\times bitwidth}{\text{\textit{peak memory bandwidth}}}$. For the upper bound of the relaxation range, we set $\text{\textit{draft\_bias}}_{max} = 5\times \text{\textit{draft\_bias}}_{min}$.

    \item \textit{draft\_k}: It adjusts the intensity of the roofline effect of the dense draft model. Similar to \textit{k1}, we set $\text{\textit{draft\_k}}_{min} = 0$ and $\text{\textit{draft\_k}}_{max} = \text{\textit{inf}}$.

    \item \textit{reject\_bias}: It represents the fixed overhead when performing rejection sampling. Vllm reports its elapsed time during SD, and we denote the maximum across measurements as $T_{rej}$. We then set $\text{\textit{reject\_bias}}_{min}=0$ and $\text{\textit{reject\_bias}}_{max}=T_{rej}$.

    \item \textit{reject\_k}: It represents the incremental processing time in rejection sampling as the input token count increases. For simplicity, we set $\text{\textit{reject\_k}}_{min}=0$ and $\text{\textit{reject\_k}}_{max}=T_{rej}$ just like \textit{reject\_bias}.

    \item $\lambda$: It represents the ratio of the empirical ridge point to the theoretical ridge point. Since memory bandwidth is typically less utilized than arithmetic units, we set $\lambda_{min}=0.2$ and $\lambda_{max}=1$.

    \item $s$: It adjusts the growing rate of execution time as input token count increases. Since $s$ serves as the base of $G(t)$, it must exceed 1 to ensure monotonic growth. However, $s$ should not be too large, as it would result in an excessively steep growth rate. In experiments, we set $s_{min}=1$ and $s_{max}=2$.

\end{itemize}

\subsection{Exploration of Alternative Measurement Selection}
\label{chap:other_items}

In this section, we demonstrate the impact of varying the number ($m$) of measurements used for fitting on the modeling results. Given that our model incorporates 10 parameters, a minimum of 10 profiling data points ($m \ge 10$) are required to determine all parameters. We present the modeling fitting with $m$ ranging from 10 to 228. The data selection method follows the stride-based approach described in the previous section, specifically \texttt{M = df[begin:end:$stride$]}. Measurement count $m$ and $stride$ satisfy the following relation: $m = \lceil 228/stride\rceil$. 

We present the MSE values of different $m$s and their corresponding fitting figures in Table~\ref{tab:m}. We also list the distinct batch sizes involved in the selected measurements, which helps explain why some configurations show inferior model fit. Due to integer division constraints, $m$s are not continuous at larger magnitudes. Generally, the modeling fits well with the real measurements, except for $m = 10, 12, 13$. The reasons are as follows. When $m=10$, the number of measurements equals the parameter count, resulting in insufficient data for robust fitting. When $m=12$ and $m=13$, the distribution of the measurement data is biased. With stride-based selection, measurements at $m=12$ and $m=13$ demonstrate notable gaps in batch size coverage (specifically, $m=12$ does not include batch sizes greater than 40, while $m=13$ does not include batch sizes within 1$\sim$24). Their MSE values are larger than that of $m=11$, despite the latter containing fewer data points for fitting. Based on this analysis, we recommend prioritizing uniform data distribution when selecting measurements, as this approach enables the development of more reliable models even with smaller datasets.

\renewcommand{\arraystretch}{1.2}
\begin{table}[]
\caption{}
\centering
\begin{small}
\begin{tabular}{c|c|c|c|l}
    \toprule
    $\bm{m}$ &$\bm{stride}$ & \textbf{MSE}         & \textbf{Figure}   & \textbf{Batch Size Involved}          \\
    \midrule
    10  & 25       & 2.216 & \ref{fig:10} & [1, 12, 16, 20, 36, 40, 44, 60, 80, 100]                                  \\
    11  & 22       & 1.764 & \ref{fig:11} & [1, 4, 8, 16, 20, 28, 32, 40, 44, 56, 100]                                \\
    12  & 20       & 4.288 & \ref{fig:12} & [1, 2, 4, 8, 12, 16, 20, 24, 28, 32, 36, 40]                              \\
    13  & 18       & 2.681 & \ref{fig:13} & [1, 24, 28, 32, 36, 40, 44, 48, 52, 56, 60, 80, 100]                      \\
    14  & 17       & 2.041 & \ref{fig:14} & [1, 2, 8, 16, 24, 32, 40, 44, 48, 52, 56, 60, 80, 100]                    \\
    15  & 16       & 1.668 & \ref{fig:15} & [1, 2, 4, 12, 16, 24, 28, 36, 40, 48, 52, 56, 60, 80, 100]                \\
    16  & 15       & 1.508 & \ref{fig:16} & [1, 2, 4, 8, 16, 20, 24, 32, 36, 40, 48, 52, 56, 60, 80, 100]             \\
    17  & 14       & 1.563 & \ref{fig:17} & [1, 2, 4, 8, 12, 20, 24, 28, 32, 40, 44, 48, 52, 56, 60, 80, 100]         \\
    18  & 13       & 1.525 & \ref{fig:18} & [1, 2, 4, 8, 12, 16, 24, 28, 32, 36, 40, 44, 48, 52, 56, 60, 80, 100]     \\
    19  & 12       & 2.080 & \ref{fig:19} & [1, 2, 4, 8, 12, 16, 20, 24, 28, 32, 36, 40, 44, 48, 52, 56, 60, 80, 100] \\
    21  & 11       & 1.679 & \ref{fig:21} & [1, 2, 4, 8, 12, 16, 20, 24, 28, 32, 36, 40, 44, 48, 52, 56, 60, 80, 100] \\
    23  & 10       & 1.800 & \ref{fig:23} & [1, 2, 4, 8, 12, 16, 20, 24, 28, 32, 36, 40, 44, 48, 52, 56, 60, 80, 100] \\
    26  & 9        & 1.716 & \ref{fig:26} & [1, 2, 4, 8, 12, 16, 20, 24, 28, 32, 36, 40, 44, 48, 52, 56, 60, 80, 100] \\
    29  & 8        & 1.524 & \ref{fig:29} & [1, 2, 4, 8, 12, 16, 20, 24, 28, 32, 36, 40, 44, 48, 52, 56, 60, 80, 100] \\
    33  & 7        & 1.526 & \ref{fig:33} & [1, 2, 4, 8, 12, 16, 20, 24, 28, 32, 36, 40, 44, 48, 52, 56, 60, 80, 100] \\
    38  & 6        & 1.715 & \ref{fig:38} & [1, 2, 4, 8, 12, 16, 20, 24, 28, 32, 36, 40, 44, 48, 52, 56, 60, 80, 100] \\
    46  & 5        & 1.644 & \ref{fig:46} & [1, 2, 4, 8, 12, 16, 20, 24, 28, 32, 36, 40, 44, 48, 52, 56, 60, 80, 100] \\
    57  & 4        & 1.509 & \ref{fig:57} & [1, 2, 4, 8, 12, 16, 20, 24, 28, 32, 36, 40, 44, 48, 52, 56, 60, 80, 100] \\
    76  & 3        & 1.553 & \ref{fig:76} & [1, 2, 4, 8, 12, 16, 20, 24, 28, 32, 36, 40, 44, 48, 52, 56, 60, 80, 100] \\
    114 & 2        & 1.485 & \ref{fig:114} & [1, 2, 4, 8, 12, 16, 20, 24, 28, 32, 36, 40, 44, 48, 52, 56, 60, 80, 100] \\
    228 & 1        & 1.523 & \ref{fig:228} & [1, 2, 4, 8, 12, 16, 20, 24, 28, 32, 36, 40, 44, 48, 52, 56, 60, 80, 100] \\
    \bottomrule
\end{tabular}
\end{small}
\label{tab:m}
\end{table}

\begin{figure}
  \centering
  \includegraphics[width=\textwidth]{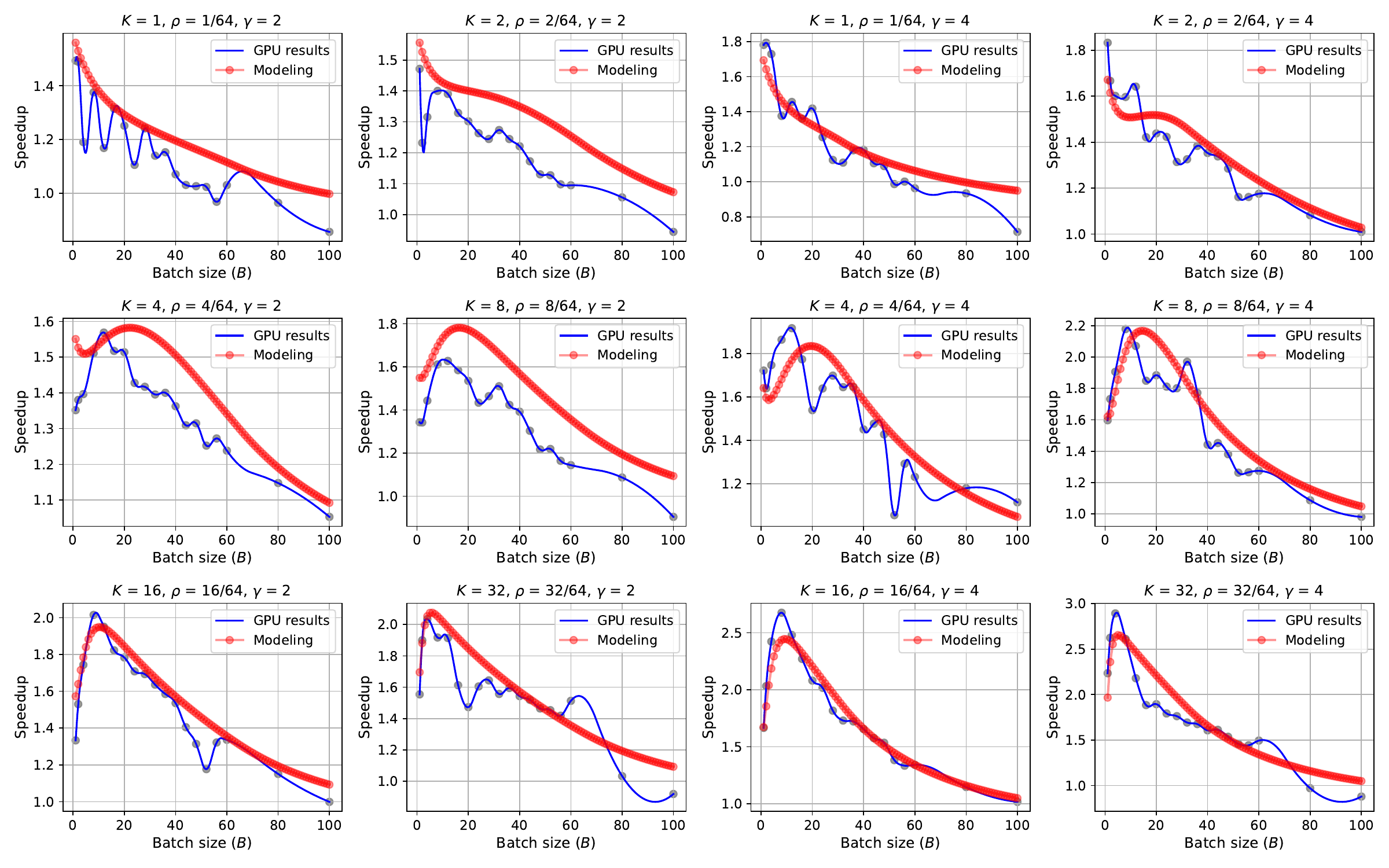}
  \caption{Comparison between GPU results and modeling with 10 measurements.}
  \label{fig:10}
\end{figure}

\begin{figure}
  \centering
  \includegraphics[width=\textwidth]{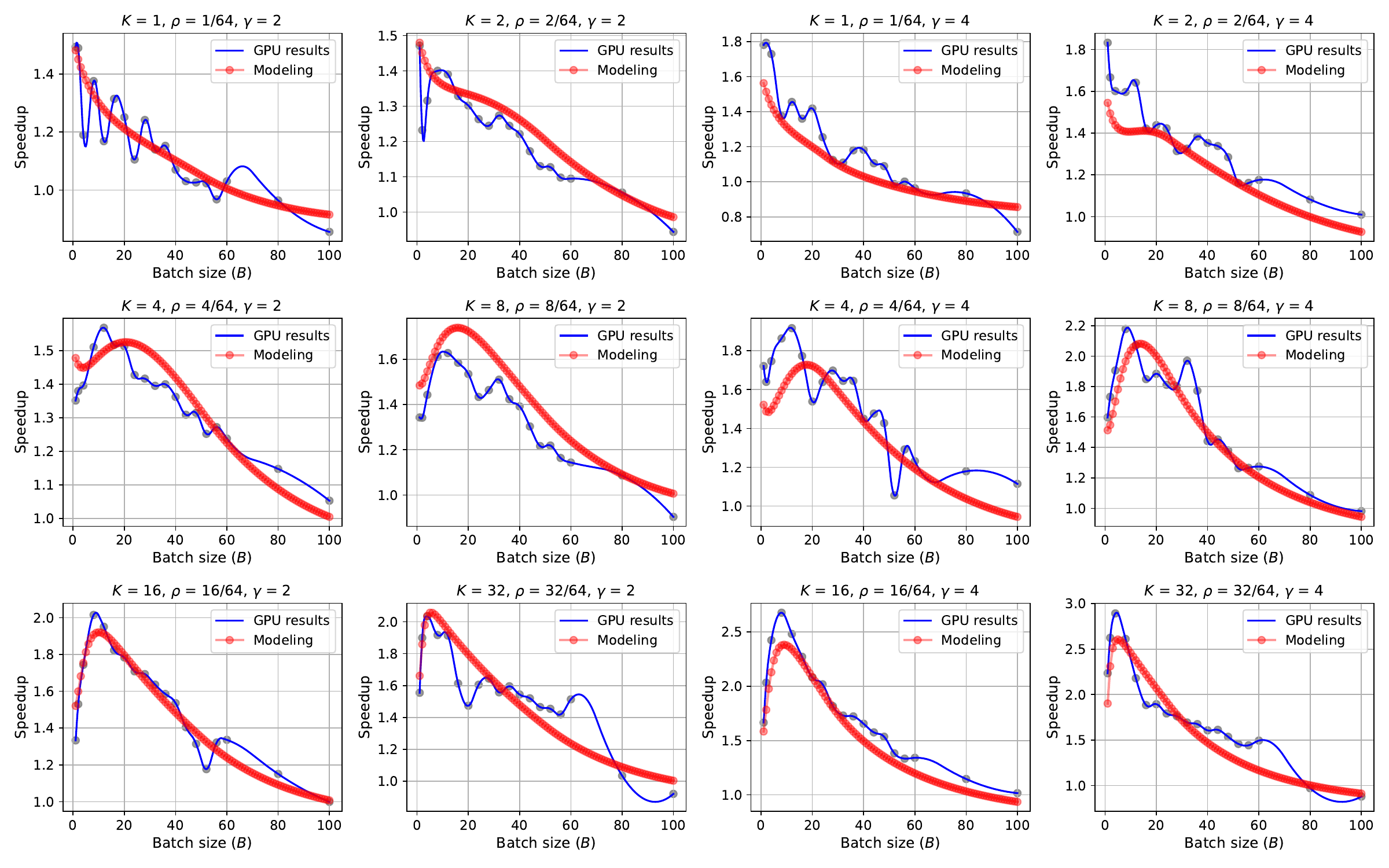}
  \caption{Comparison between GPU results and modeling with 11 measurements.}
  \label{fig:11}
\end{figure}

\begin{figure}
  \centering
  \includegraphics[width=\textwidth]{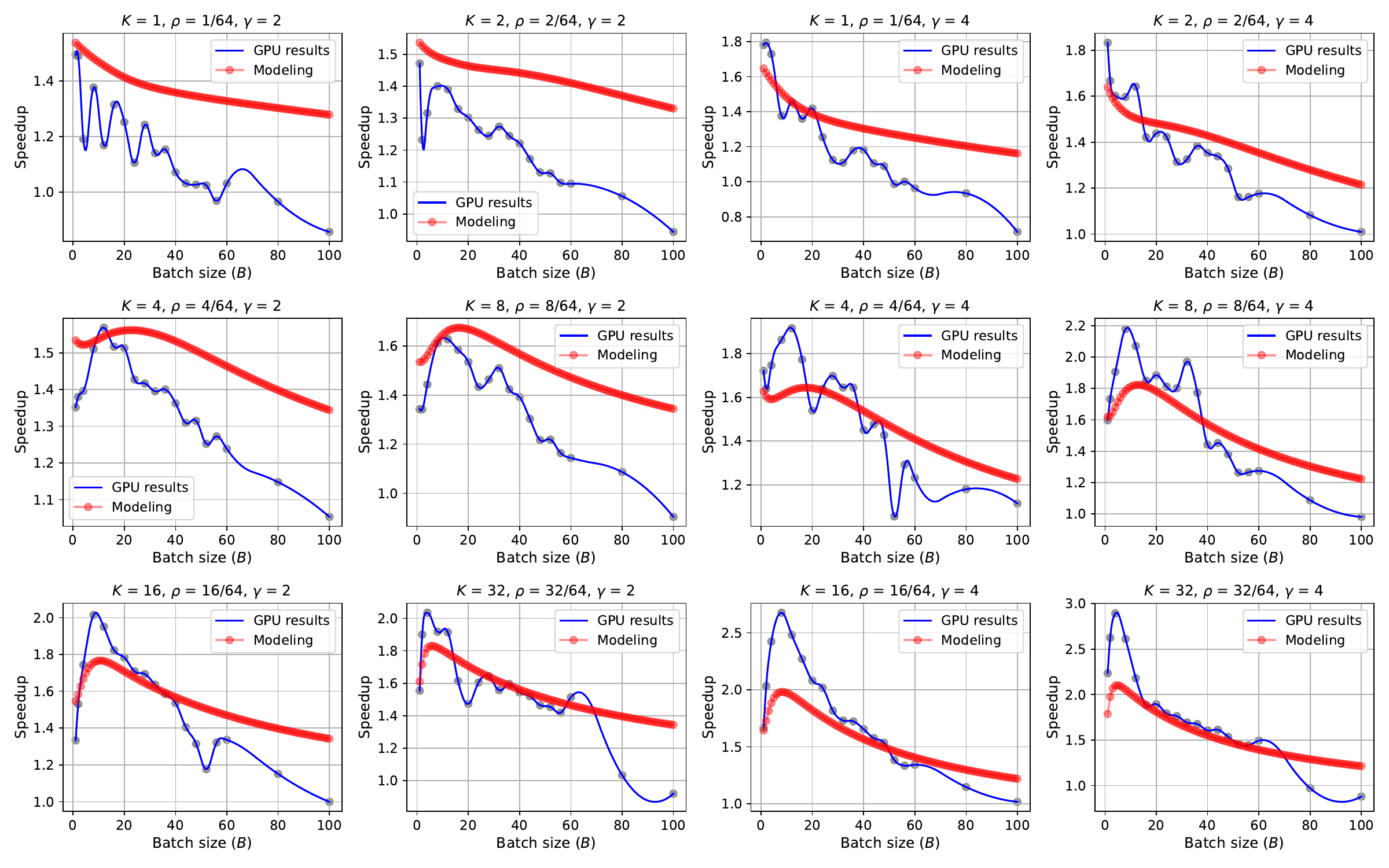}
  \caption{Comparison between GPU results and modeling with 12 measurements.}
  \label{fig:12}
\end{figure}

\begin{figure}
  \centering
  \includegraphics[width=\textwidth]{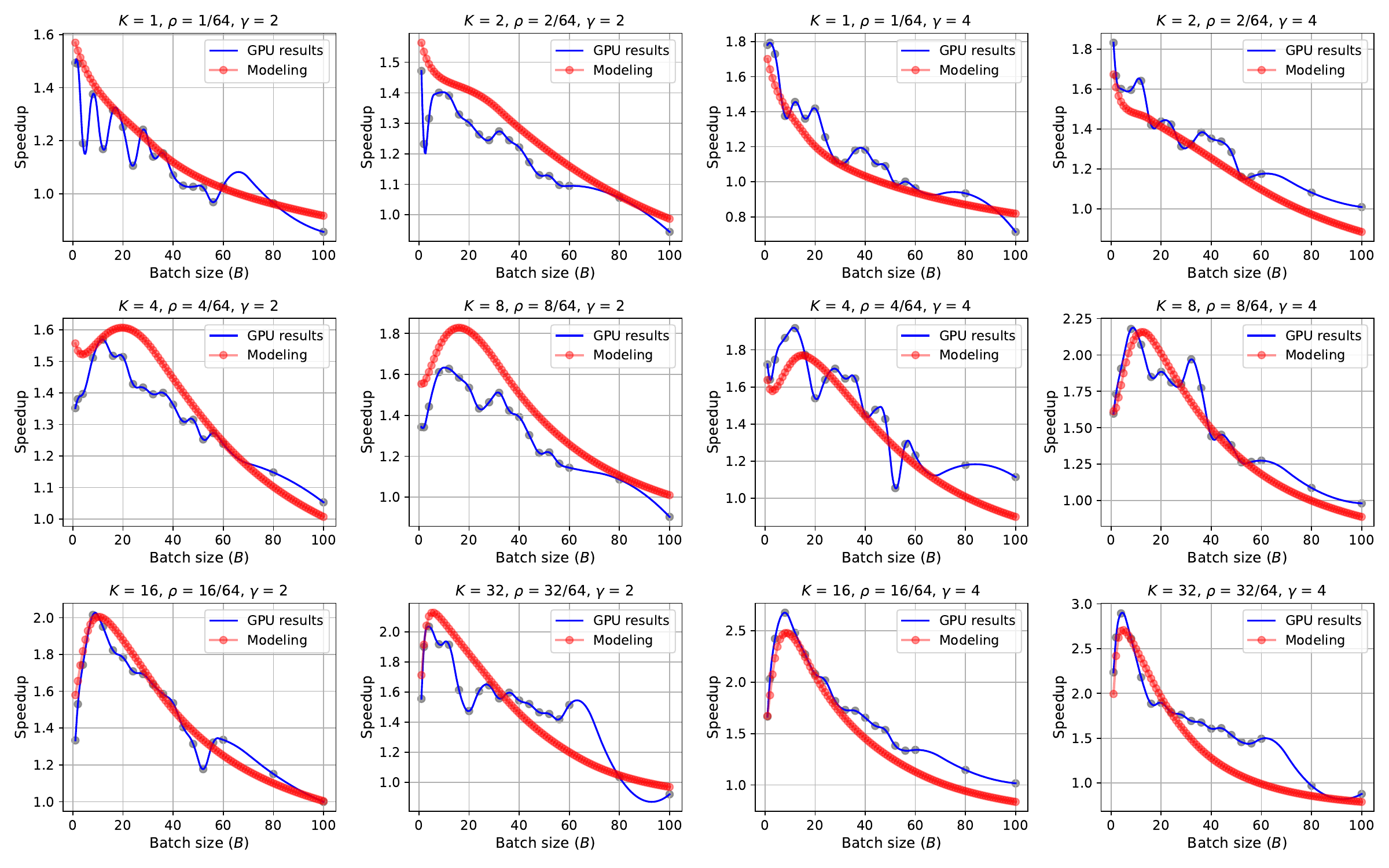}
  \caption{Comparison between GPU results and modeling with 13 measurements.}
  \label{fig:13}
\end{figure}

\begin{figure}
  \centering
  \includegraphics[width=\textwidth]{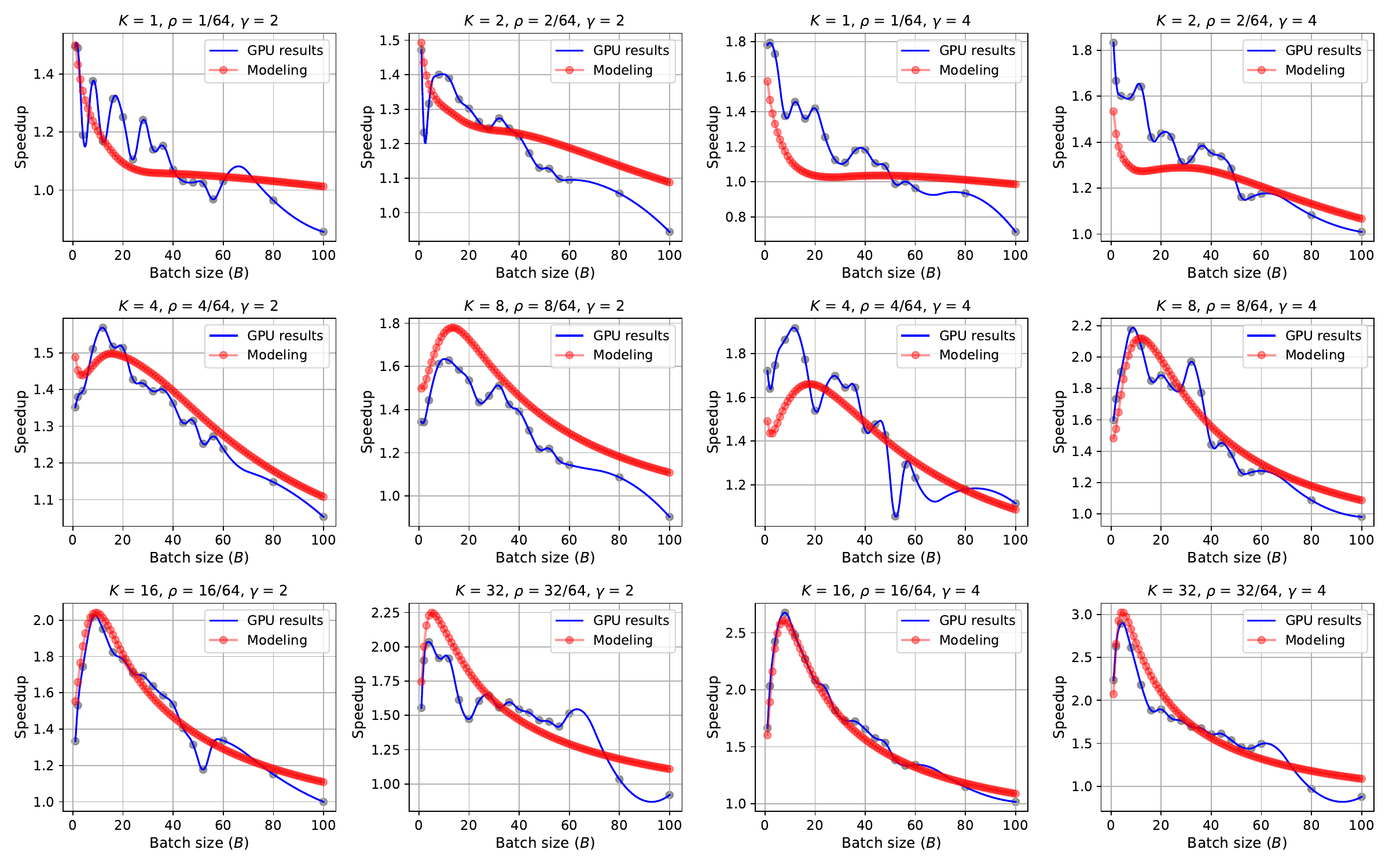}
  \caption{Comparison between GPU results and modeling with 14 measurements.}
  \label{fig:14}
\end{figure}

\begin{figure}
  \centering
  \includegraphics[width=\textwidth]{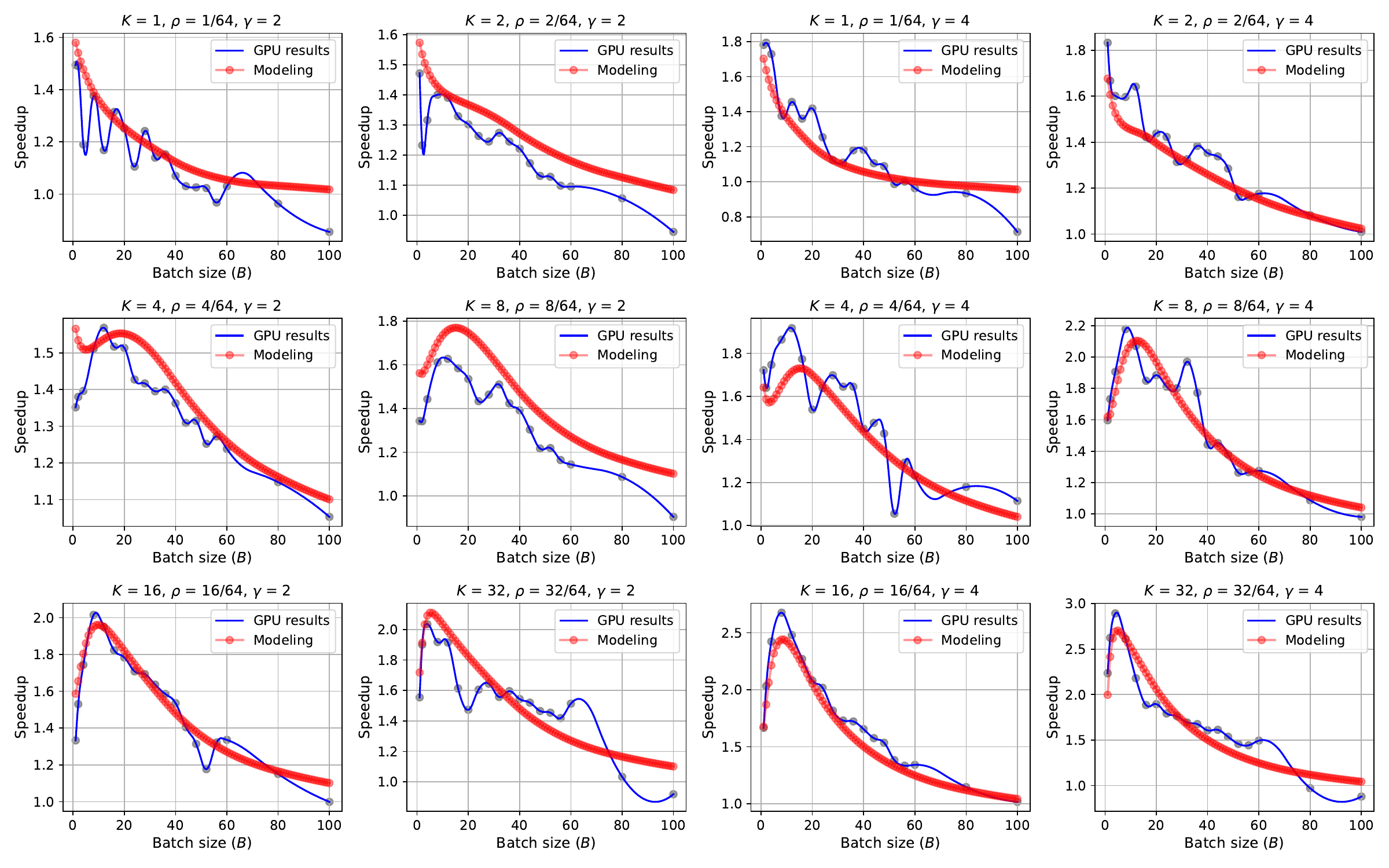}
  \caption{Comparison between GPU results and modeling with 15 measurements.}
  \label{fig:15}
\end{figure}

\begin{figure}
  \centering
  \includegraphics[width=\textwidth]{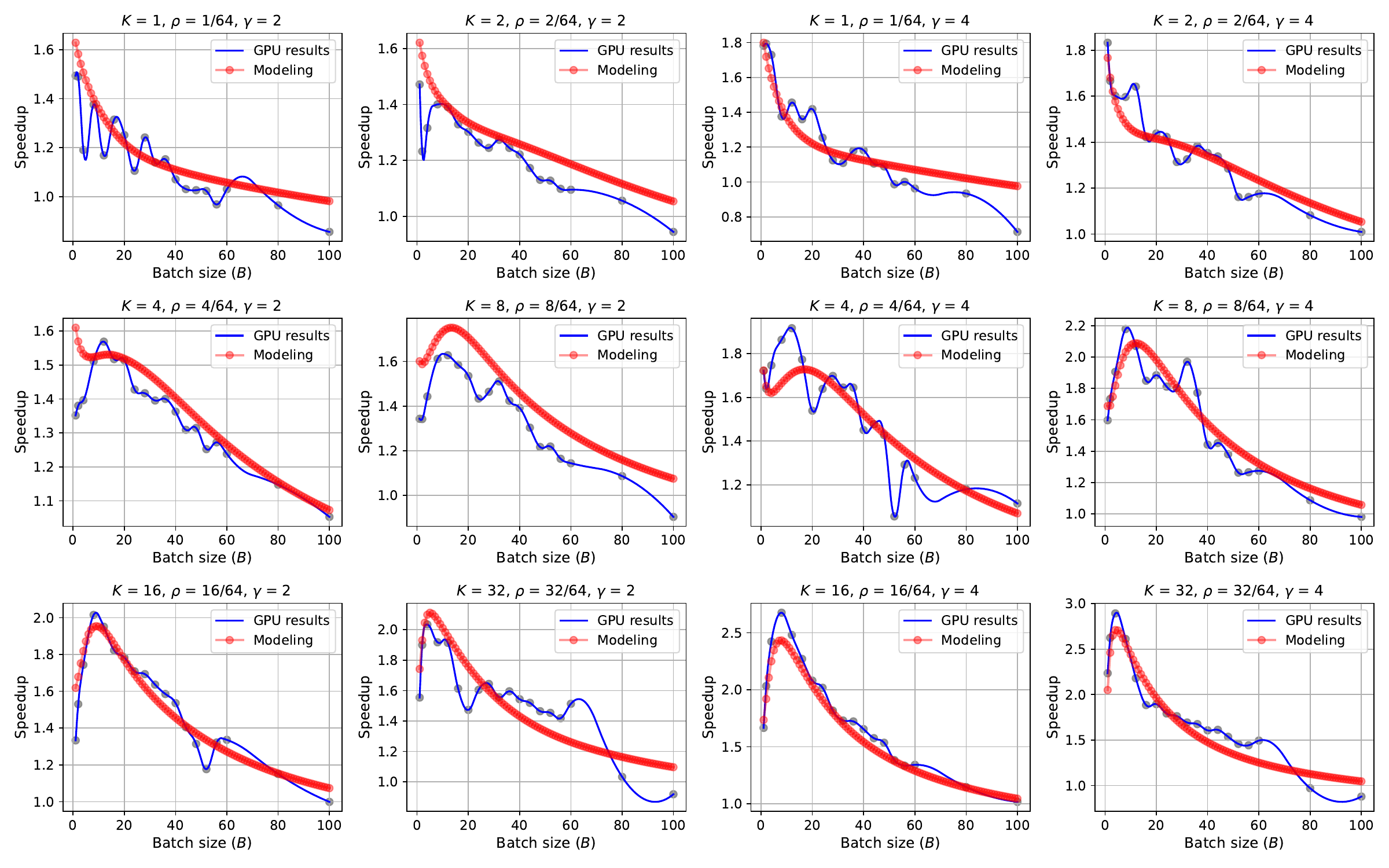}
  \caption{Comparison between GPU results and modeling with 16 measurements.}
  \label{fig:16}
\end{figure}

\begin{figure}
  \centering
  \includegraphics[width=\textwidth]{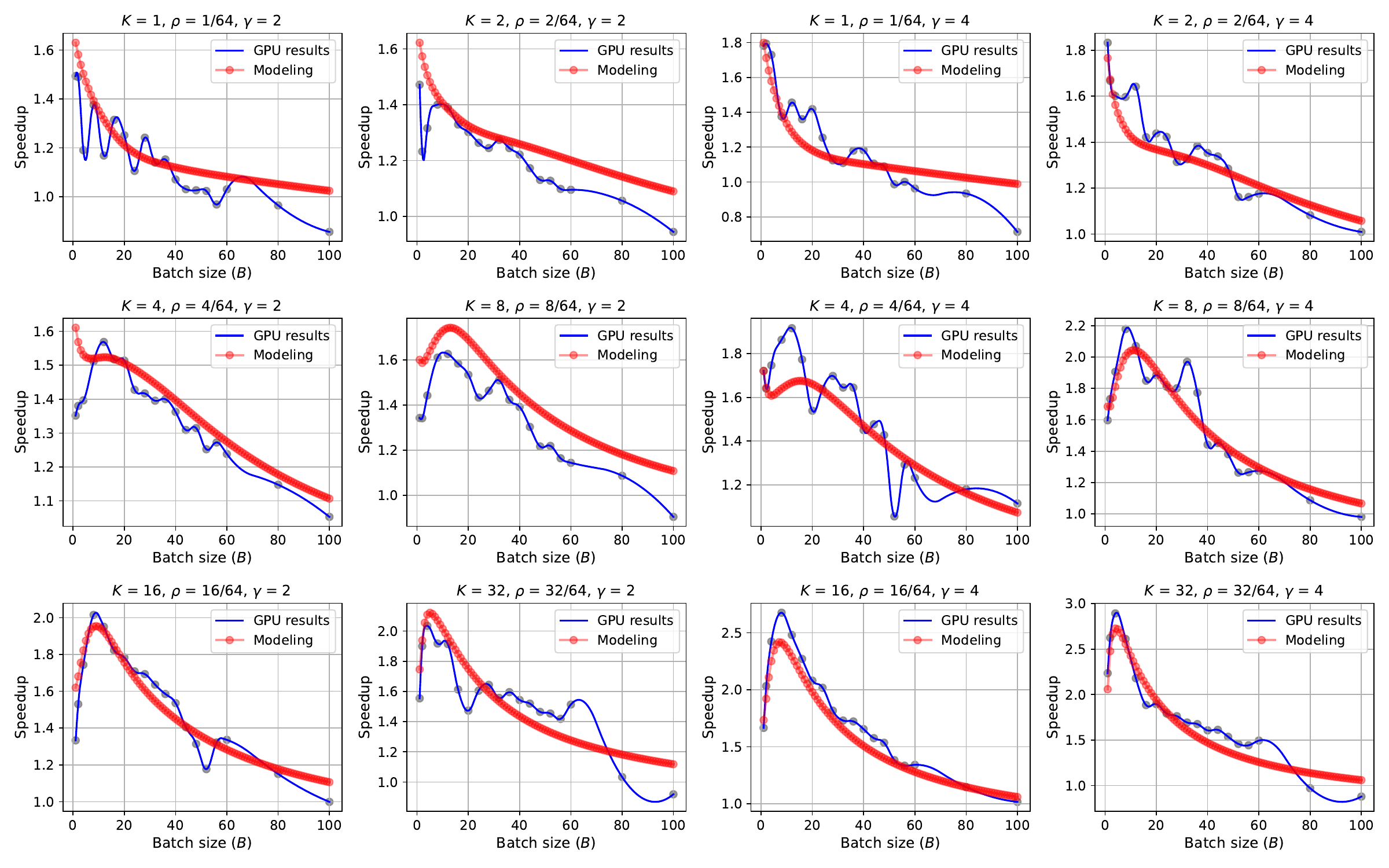}
  \caption{Comparison between GPU results and modeling with 17 measurements.}
  \label{fig:17}
\end{figure}

\begin{figure}
  \centering
  \includegraphics[width=\textwidth]{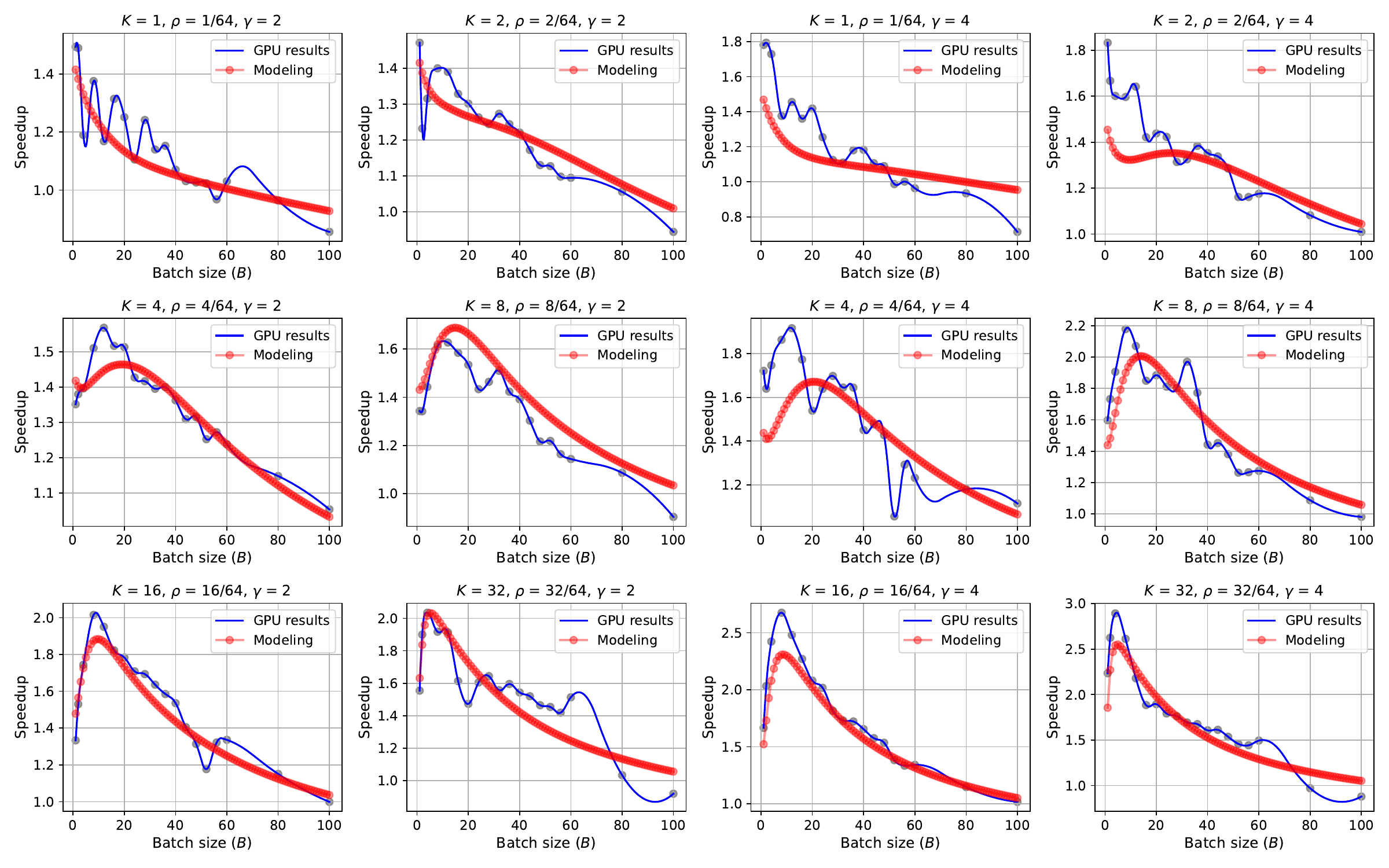}
  \caption{Comparison between GPU results and modeling with 18 measurements.}
  \label{fig:18}
\end{figure}

\begin{figure}
  \centering
  \includegraphics[width=\textwidth]{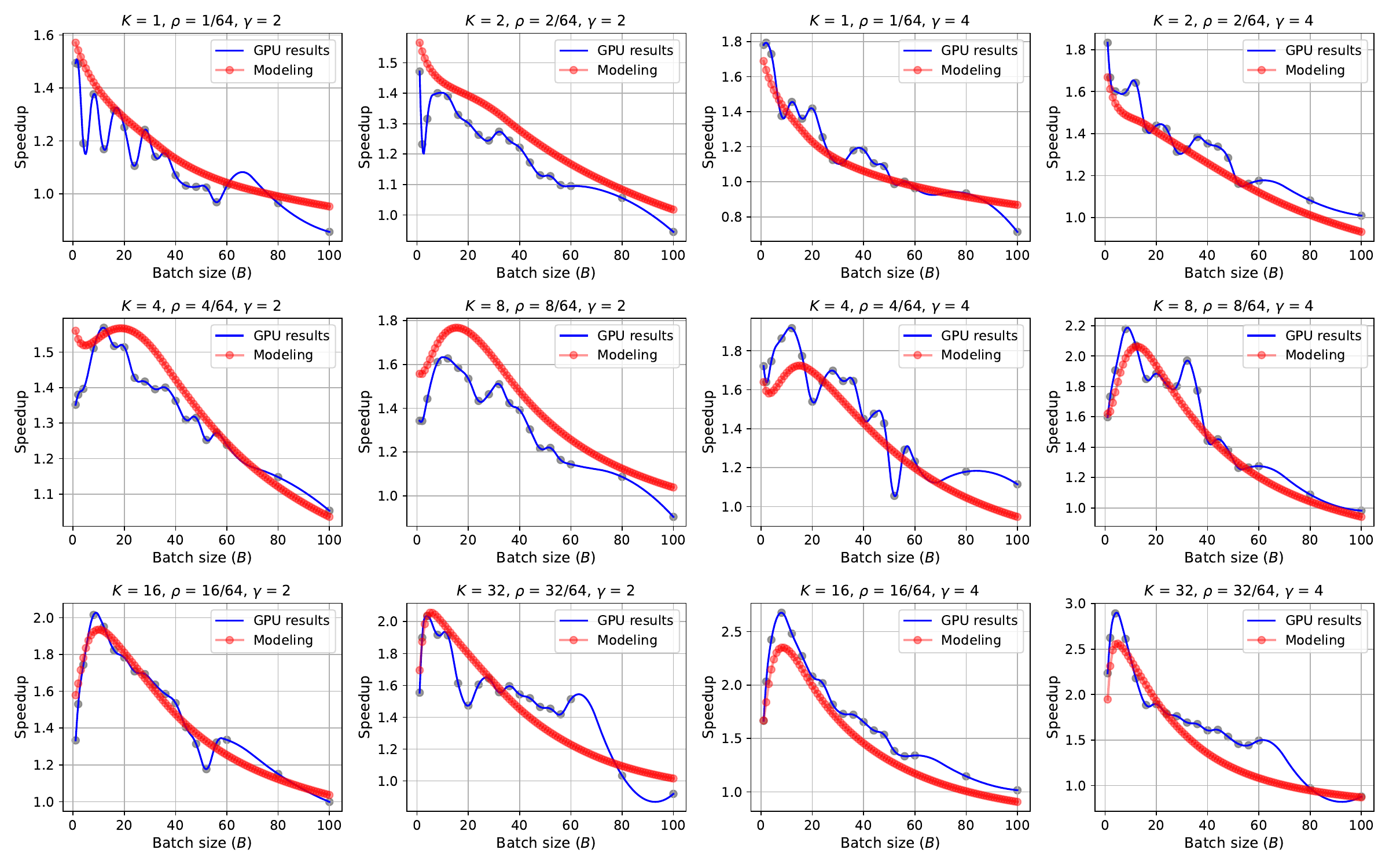}
  \caption{Comparison between GPU results and modeling with 19 measurements.}
  \label{fig:19}
\end{figure}

\begin{figure}
  \centering
  \includegraphics[width=\textwidth]{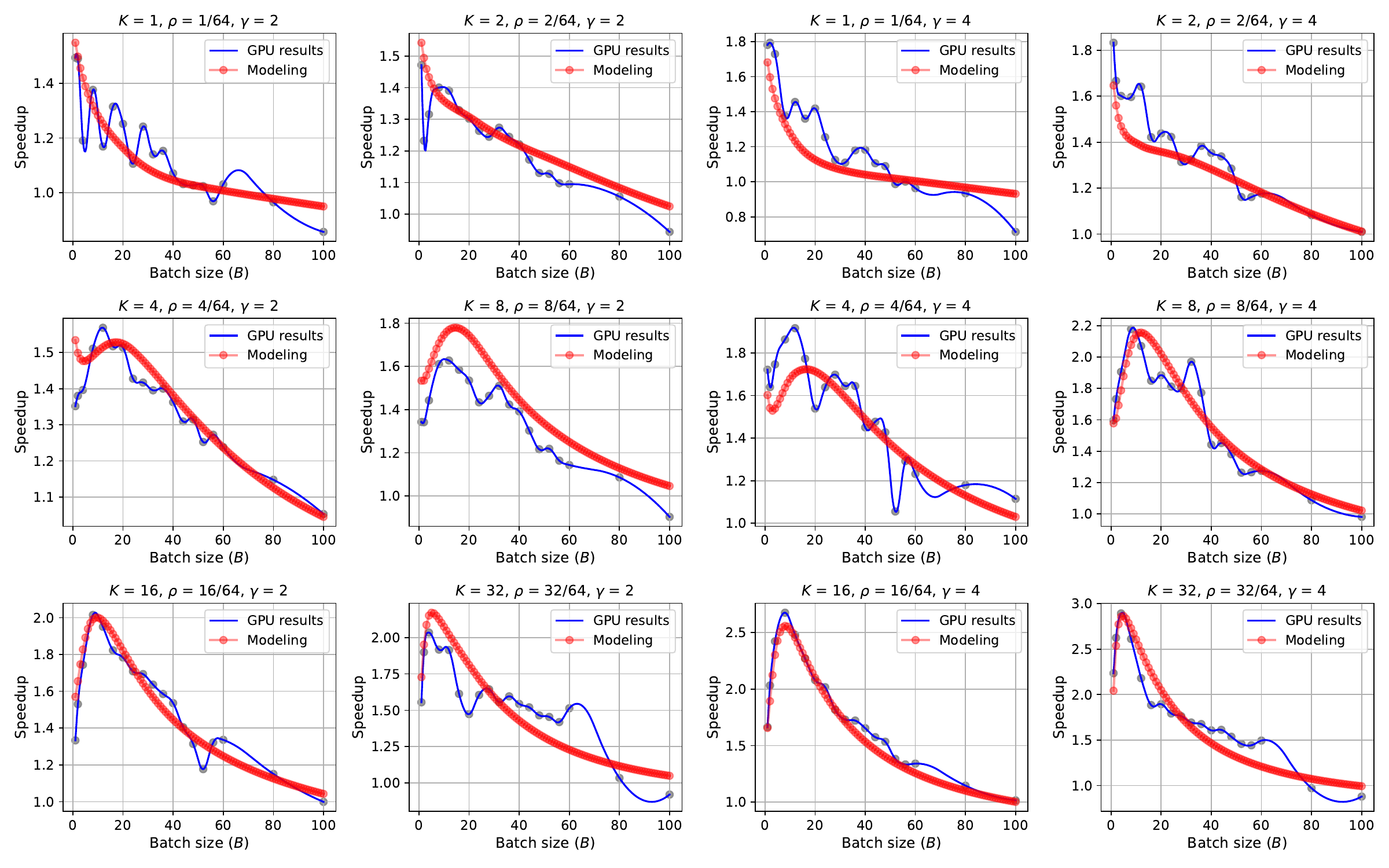}
  \caption{Comparison between GPU results and modeling with 21 measurements.}
  \label{fig:21}
\end{figure}

\begin{figure}
  \centering
  \includegraphics[width=\textwidth]{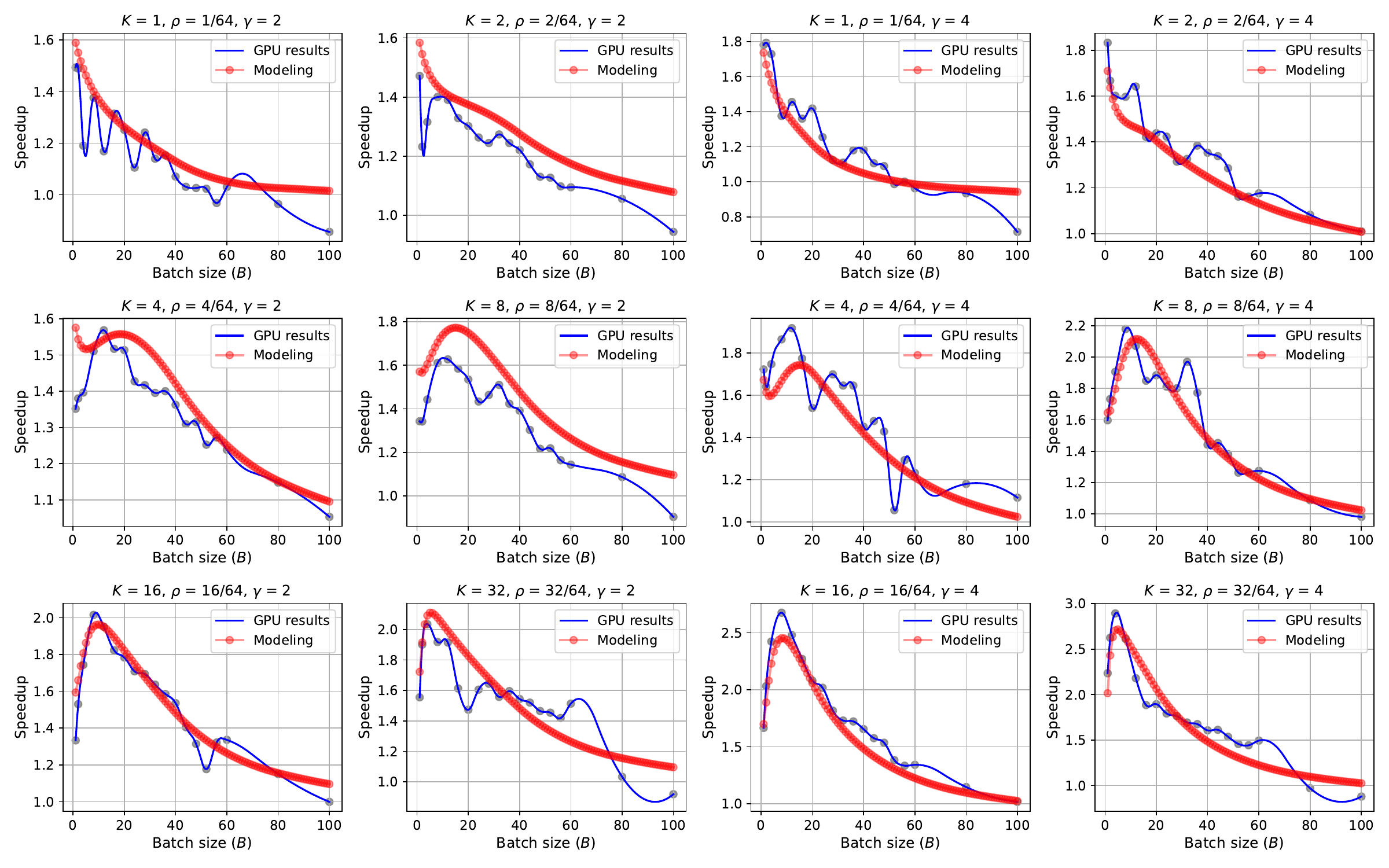}
  \caption{Comparison between GPU results and modeling with 23 measurements.}
  \label{fig:23}
\end{figure}

\begin{figure}
  \centering
  \includegraphics[width=\textwidth]{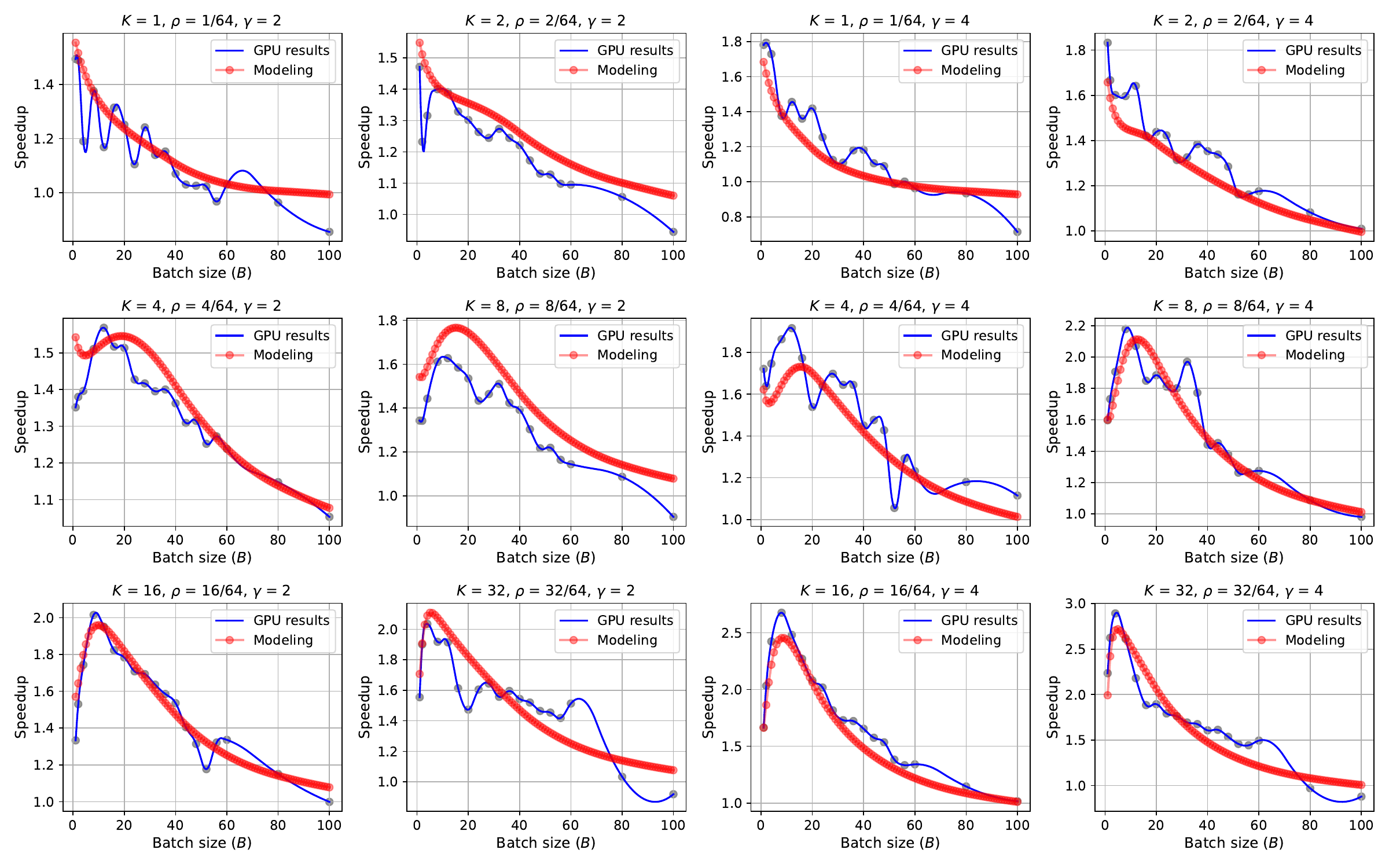}
  \caption{Comparison between GPU results and modeling with 26 measurements.}
  \label{fig:26}
\end{figure}

\begin{figure}
  \centering
  \includegraphics[width=\textwidth]{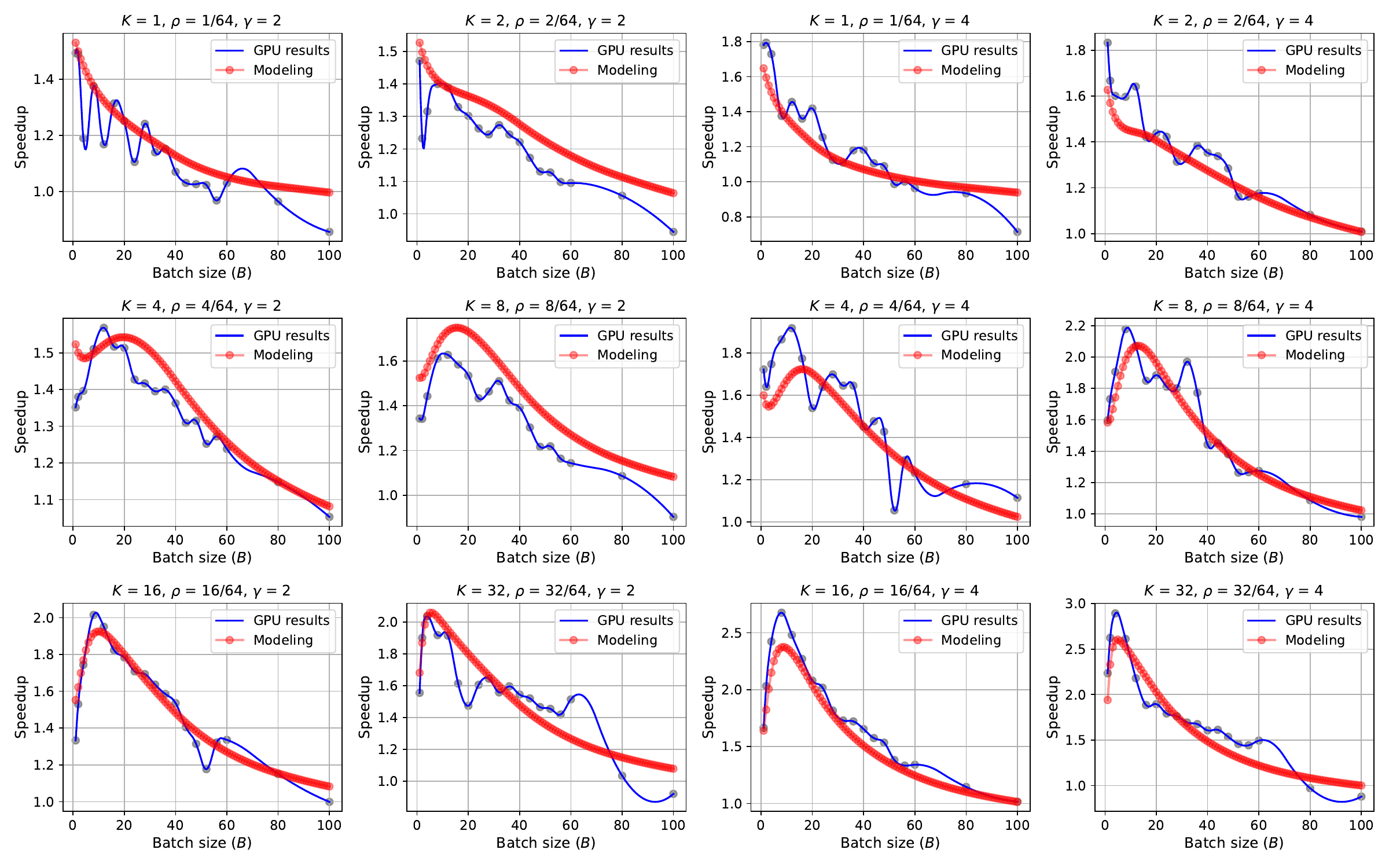}
  \caption{Comparison between GPU results and modeling with 29 measurements.}
  \label{fig:29}
\end{figure}

\begin{figure}
  \centering
  \includegraphics[width=\textwidth]{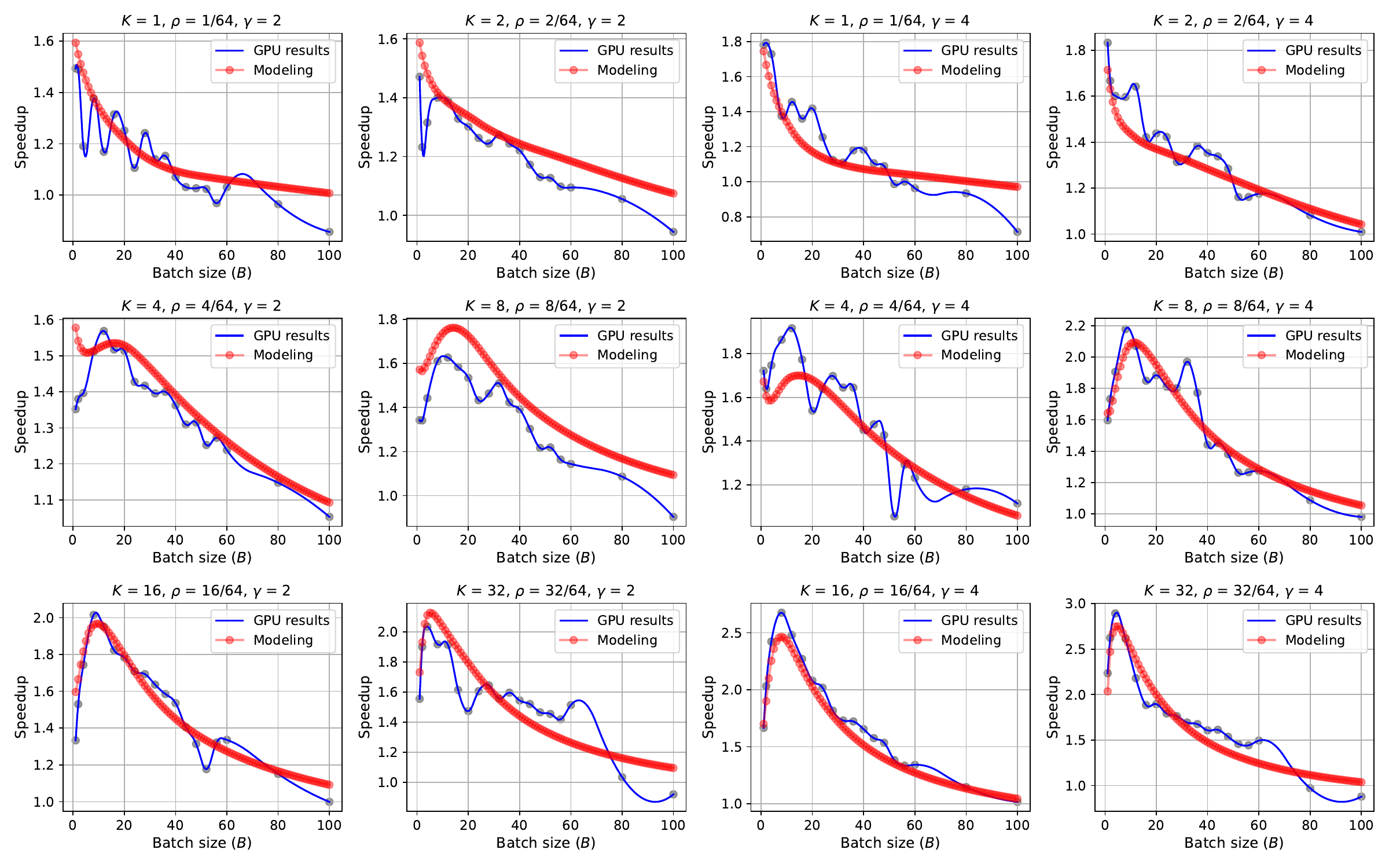}
  \caption{Comparison between GPU results and modeling with 33 measurements.}
  \label{fig:33}
\end{figure}

\begin{figure}
  \centering
  \includegraphics[width=\textwidth]{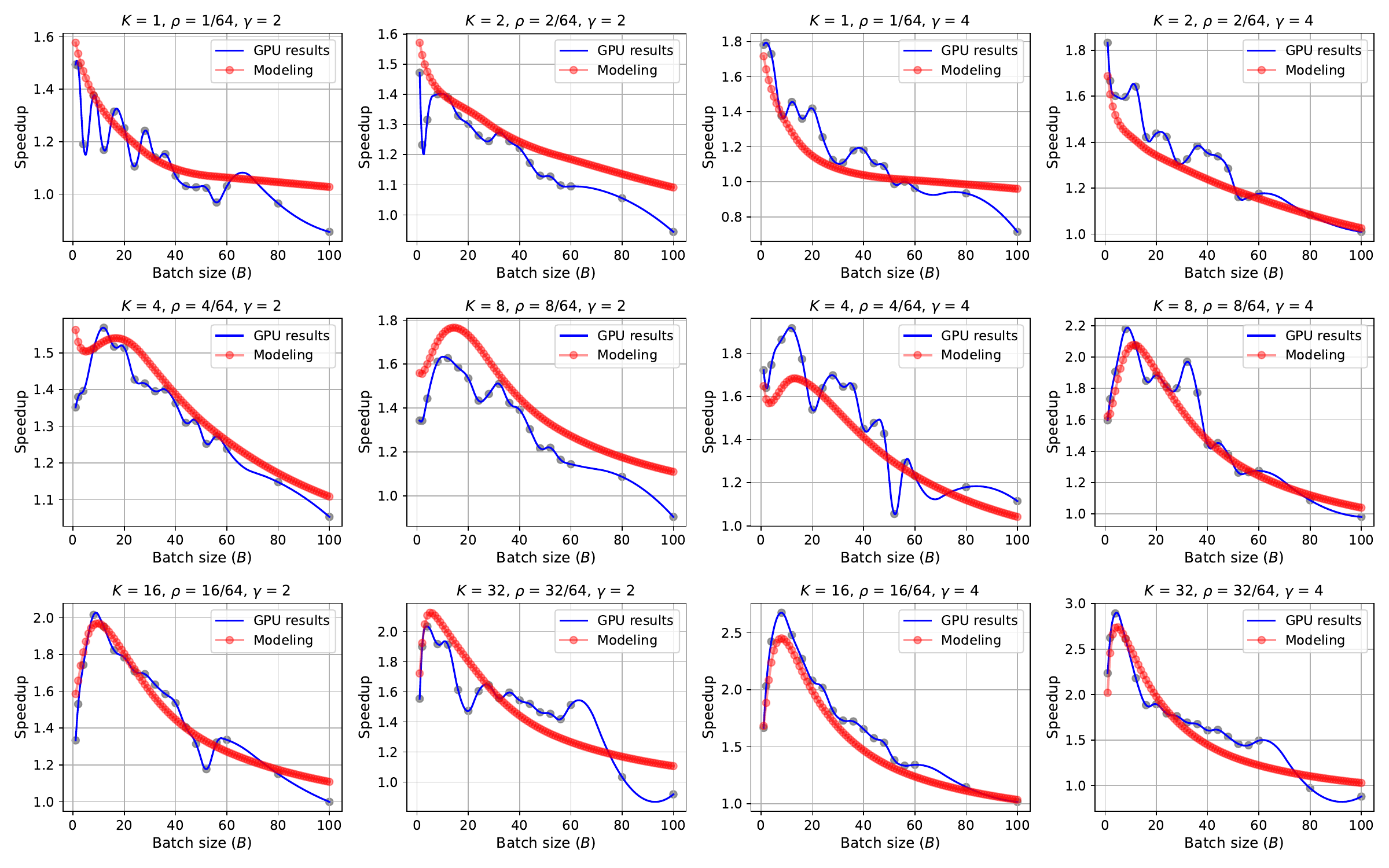}
  \caption{Comparison between GPU results and modeling with 38 measurements.}
  \label{fig:38}
\end{figure}

\begin{figure}
  \centering
  \includegraphics[width=\textwidth]{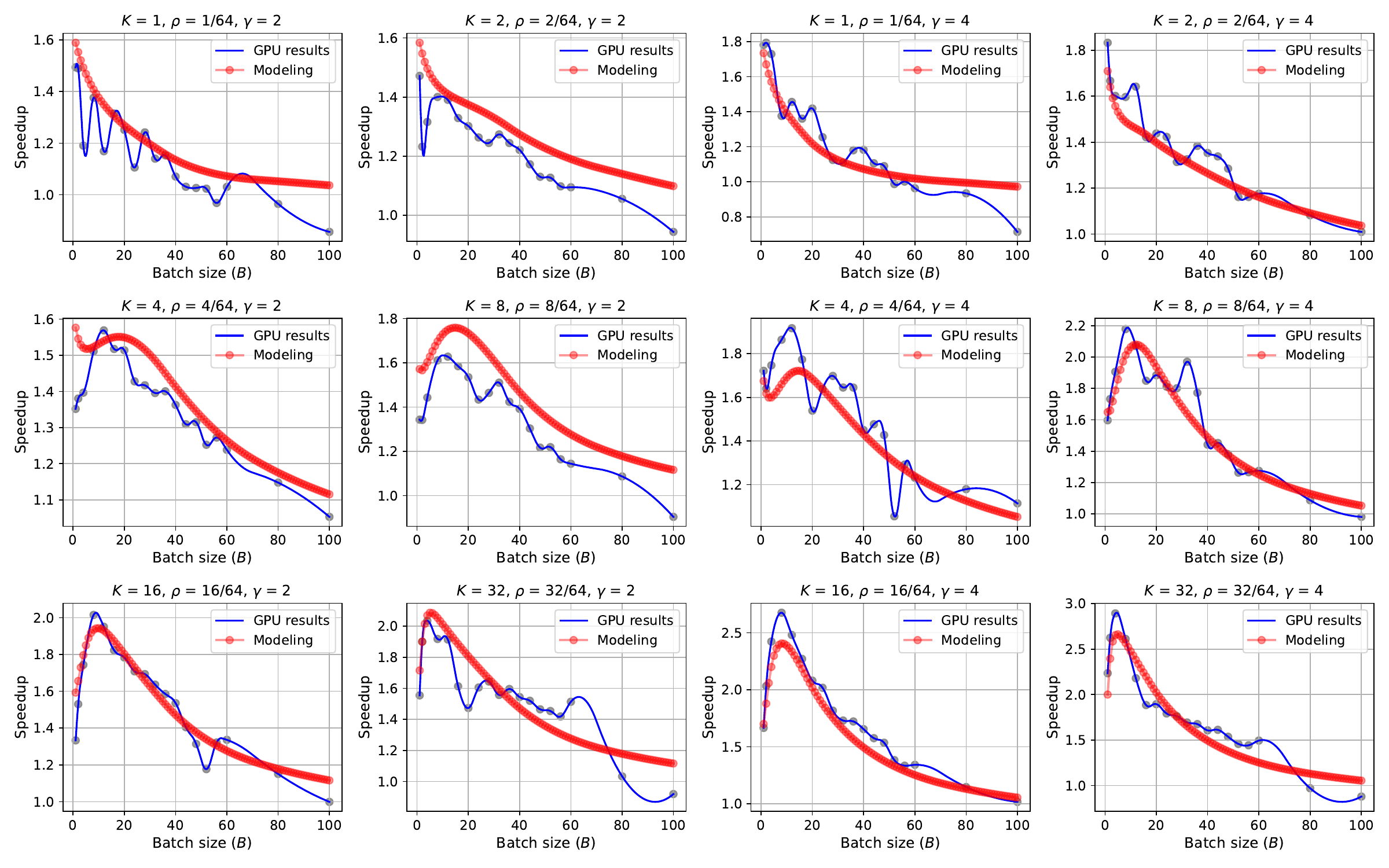}
  \caption{Comparison between GPU results and modeling with 46 measurements.}
  \label{fig:46}
\end{figure}

\begin{figure}
  \centering
  \includegraphics[width=\textwidth]{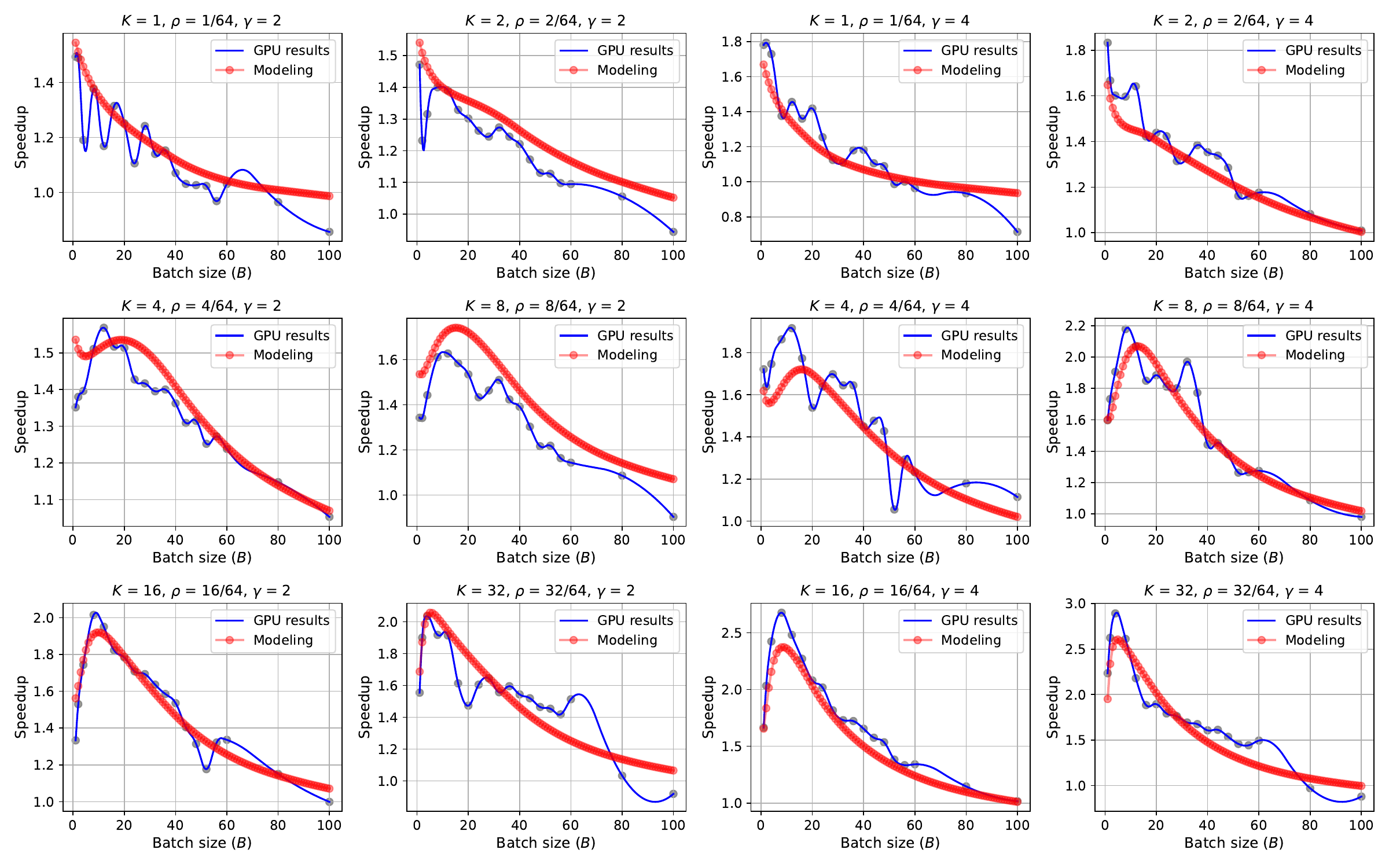}
  \caption{Comparison between GPU results and modeling with 57 measurements.}
  \label{fig:57}
\end{figure}

\begin{figure}
  \centering
  \includegraphics[width=\textwidth]{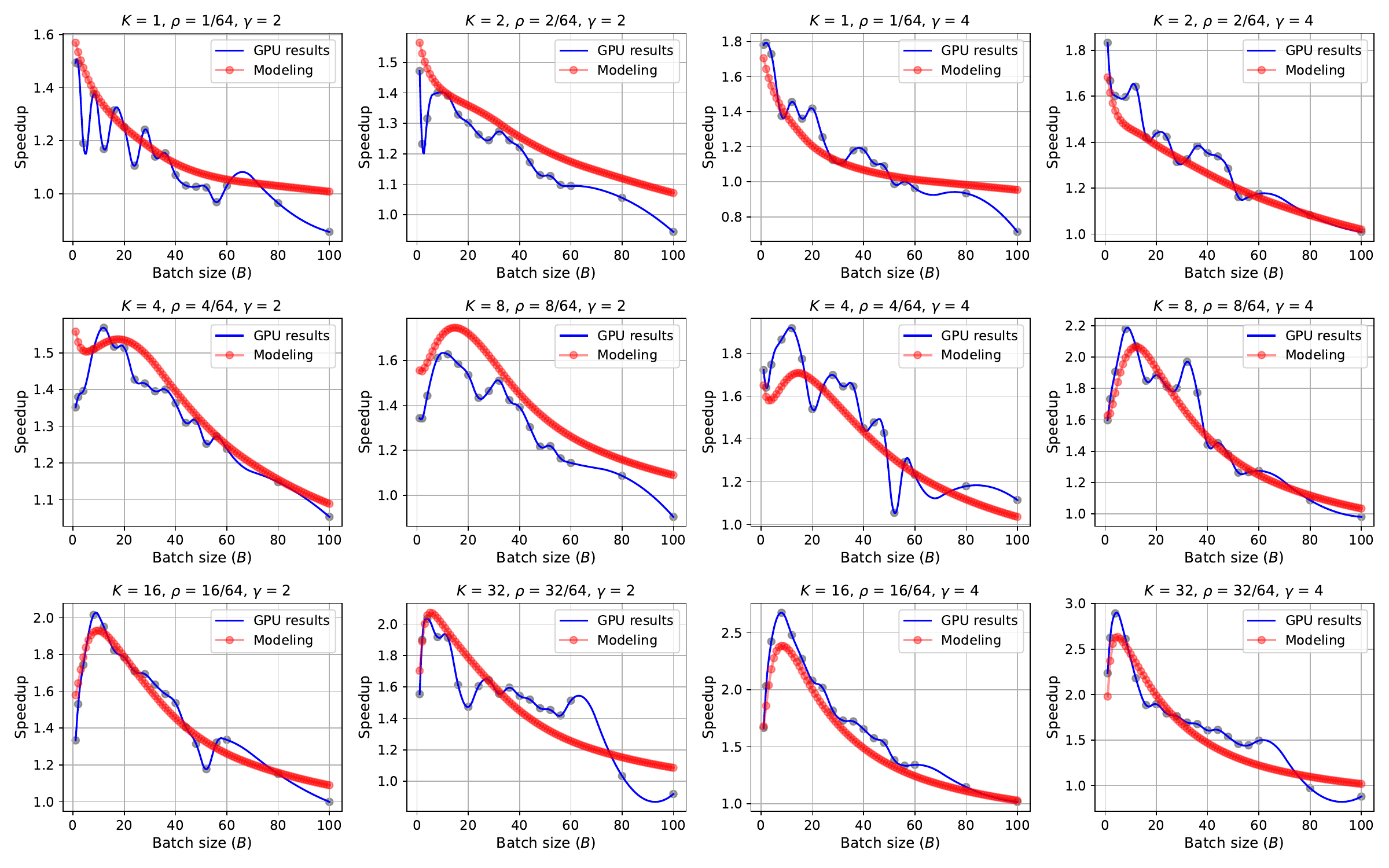}
  \caption{Comparison between GPU results and modeling with 76 measurements.}
  \label{fig:76}
\end{figure}

\begin{figure}
  \centering
  \includegraphics[width=\textwidth]{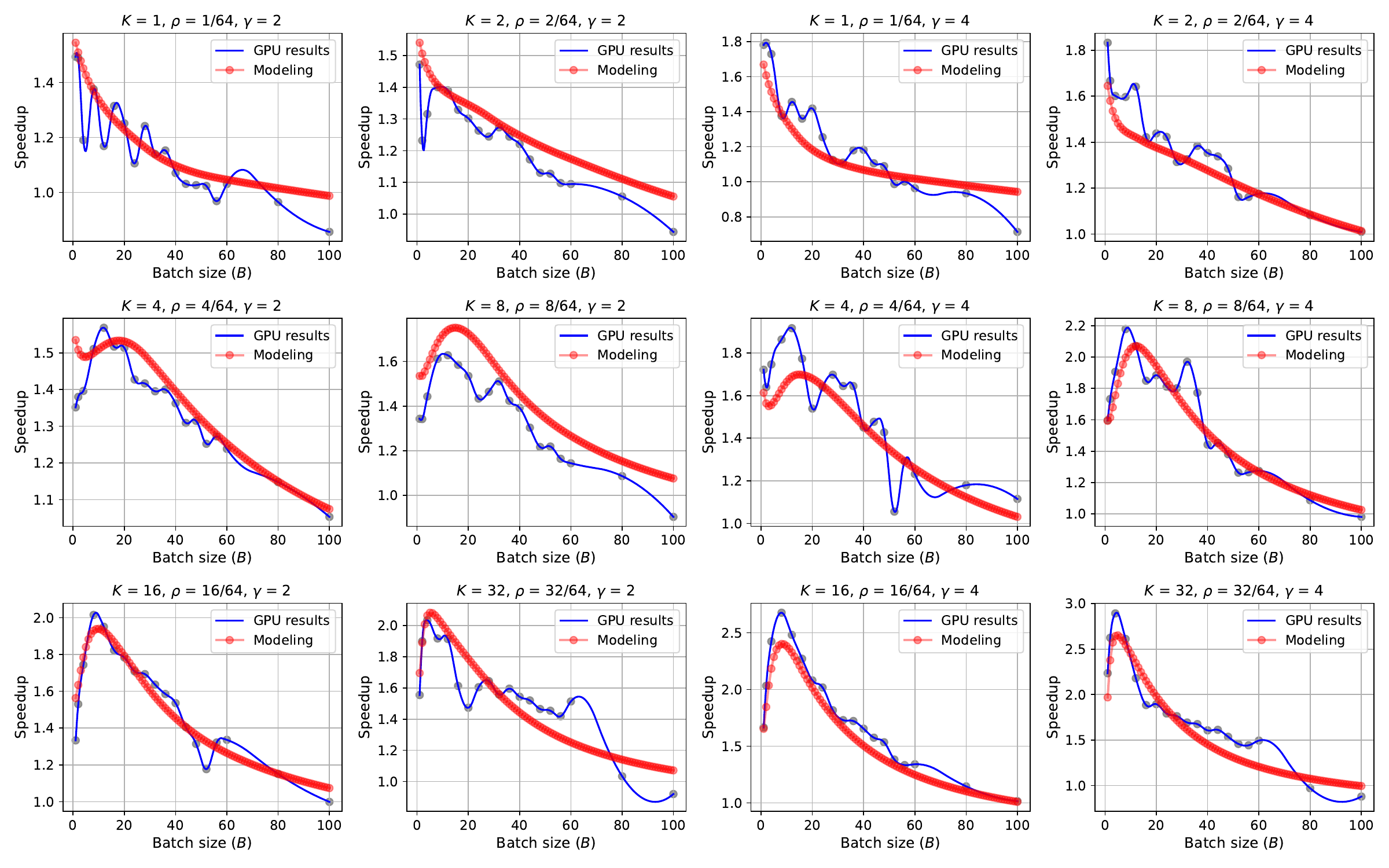}
  \caption{Comparison between GPU results and modeling with 114 measurements.}
  \label{fig:114}
\end{figure}

\begin{figure}
  \centering
  \includegraphics[width=\textwidth]{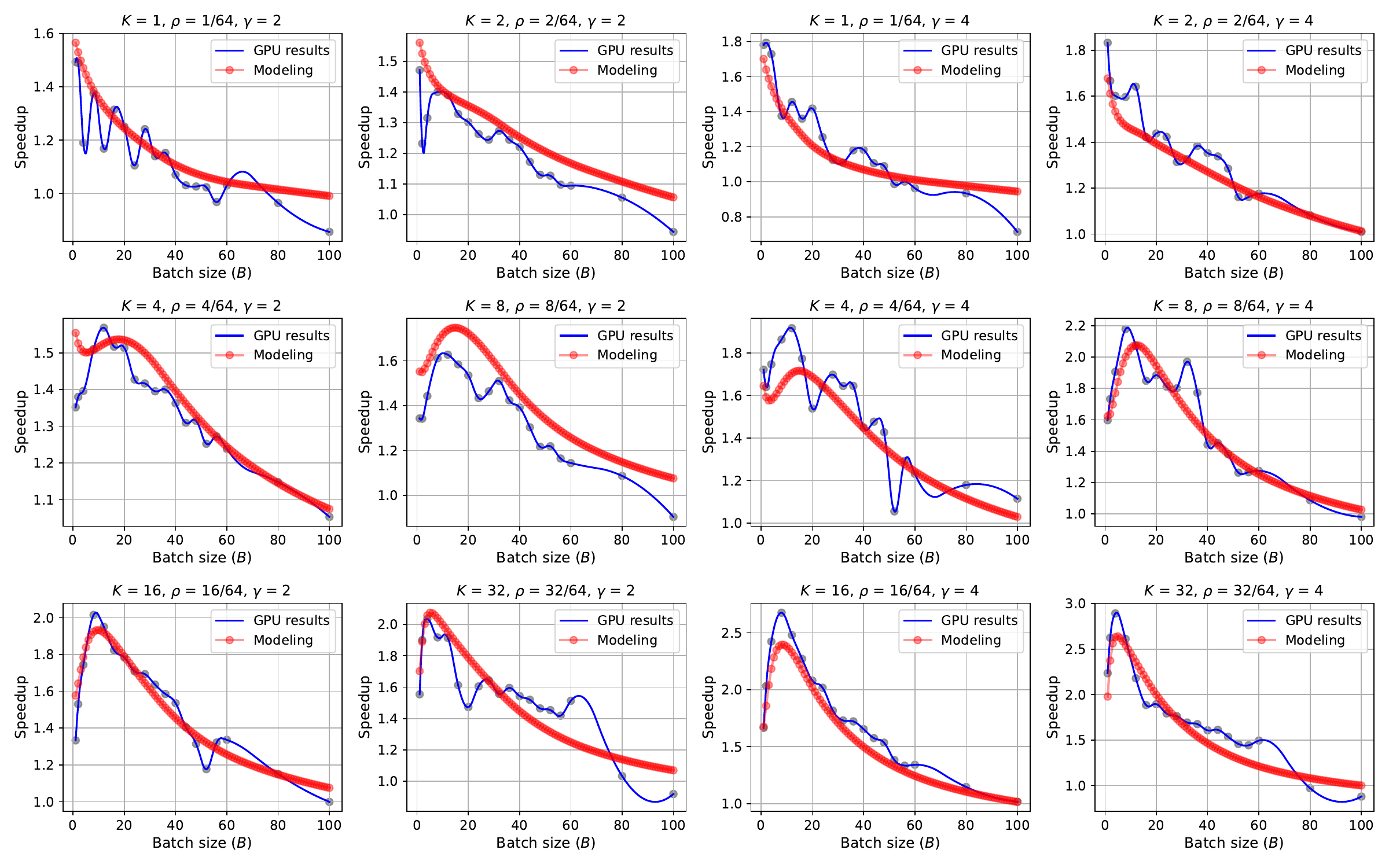}
  \caption{Comparison between GPU results and modeling with 228 measurements.}
  \label{fig:228}
\end{figure}

\end{document}